%% file: 0_main.tex
\documentclass{article}
\input{./math_commands.tex}

\input{./macros}

\usepackage{microtype}
\usepackage{graphicx}
\usepackage{booktabs} 

\usepackage{hyperref}

\usepackage[accepted]{icml2024}

\usepackage{amsmath}
\usepackage{amssymb}
\usepackage{mathtools}
\usepackage{amsthm}
\usepackage{url}
\usepackage{xspace}
\usepackage{xcolor}
\usepackage{enumitem}
\usepackage{subcaption}
\usepackage{wrapfig}
\usepackage{booktabs}
\usepackage{multirow}
\usepackage{tikz,pgfplots}
\usepackage{framed}
\usepackage{color,colortbl}
\usepackage{fancyvrb}
\usepackage{spverbatim}
\usepackage[framemethod=TikZ]{mdframed}
\usepackage{microtype}
\usepackage{titletoc}
\usepackage{float}
\usepackage[capitalize,noabbrev]{cleveref}
\usepackage[linesnumbered,algoruled,boxed,lined,noend,procnumbered,algo2e]{algorithm2e} 
\usepgfplotslibrary{groupplots}

\theoremstyle{plain}

\theoremstyle{definition}

\theoremstyle{remark}

\icmltitlerunning{O3D: Offline Data-driven Discovery and Distillation}

\begin{document}

\twocolumn[
\icmltitle{O3D: Offline Data-driven Discovery and Distillation \\ for Sequential Decision-Making with Large Language Models}



\icmlsetsymbol{equal}{*}

\begin{icmlauthorlist}
\icmlauthor{Yuchen Xiao}{equal,yyy}
\icmlauthor{Yanchao Sun}{equal,yyy}
\icmlauthor{Mengda Xu}{yyy}
\icmlauthor{Udari Madhushani}{yyy}
\icmlauthor{Jared Vann}{yyy}
\icmlauthor{Deepeka Garg}{yyy}
\icmlauthor{Sumitra Ganesh}{yyy}
\end{icmlauthorlist}


\icmlaffiliation{yyy}{JPMorgan AI Research}

\icmlcorrespondingauthor{Yuchen Xiao}{yuchen.xiao@jpmchase.com}

\icmlkeywords{Machine Learning, ICML}

\vskip 0.3in
]



\printAffiliationsAndNotice{\icmlEqualContribution} 

\begin{abstract}
Recent advancements in large language models (LLMs) have exhibited promising performance in solving sequential decision-making problems. 
By imitating few-shot examples provided in the prompts (i.e., in-context learning), an LLM agent can interact with an external environment and complete given tasks without additional training.
However, such few-shot examples are often insufficient to generate high-quality solutions for complex and long-horizon tasks, while the limited context length cannot consume larger-scale demonstrations with long interaction horizons.
To this end, we propose an offline learning framework that utilizes offline data at scale (e.g, logs of human interactions) to improve LLM-powered policies without finetuning. 
The proposed method \emph{O3D (Offline Data-driven Discovery and Distillation)} automatically discovers reusable skills and distills generalizable knowledge across multiple tasks based on offline interaction data, advancing the capability of solving downstream tasks.
Empirical results under two interactive decision-making benchmarks (ALFWorld and WebShop) verify that O3D can notably enhance the decision-making capabilities of LLMs through the offline discovery and distillation process, and consistently outperform baselines across various LLMs. 
\end{abstract}

\input{1_intro.tex}
\input{2_related.tex}

\input{3_method.tex}
\input{4_exp.tex}
\input{5_discuss.tex}


\section*{Impact Statements}
This paper presents work whose goal is to advance the field of Machine Learning. There are many potential societal consequences of our work, none which we feel must be specifically highlighted here.

\bibliography{icml2024_conference}
\bibliographystyle{icml2024}

\newpage
\appendix
\onecolumn

\input{6_appendix.tex}




\end{document}

%% file: math_commands.tex
\usepackage{amsmath,amsfonts,bm}

\def\eqref#1{equation~\ref{#1}}

\def\1{\bm{1}}

\DeclareMathAlphabet{\mathsfit}{\encodingdefault}{\sfdefault}{m}{sl}
\SetMathAlphabet{\mathsfit}{bold}{\encodingdefault}{\sfdefault}{bx}{n}



%% file: macros.tex
\newcommand{\ourmodfull}{{Offline Data-driven Discovery and Distillation}\xspace}
\newcommand{\ourmod}{{\texttt{O3D}}\xspace}
\newcommand{\ourmodcode}{{\texttt{O3D-Code}}\xspace}
\newcommand{\ourmodhuman}{{\texttt{O3D-Human}}\xspace}

\newcommand{\llmpolname}{{Text-based-Policy}\xspace}
\newcommand{\codepolname}{{Code-based-Policy}\xspace}
\newcommand{\contrastname}{{Trajectory Contrasting}\xspace} 
\newcommand{\discoverpmt}{\textcolor{blue}{$\mathsf{DiscoverPrompt}$}\xspace}
\newcommand{\distillpmt}{\textcolor{red}{$\mathsf{DistillPrompt}$}\xspace}

\newcommand{\llmpolicy}{\pi_{\mathrm{llm}}}
\newcommand{\codepolicy}{\pi_{\mathrm{code}}}
\newcommand{\basepolicy}{\pi_{\mathrm{base}}}
\newcommand{\prims}{\mathcal{P}}
\newcommand{\skills}{\mathcal{Z}}
\newcommand{\tips}{\mathcal{T}}
\newcommand{\skill}{z}
\newcommand{\pretpara}{\theta_{\mathrm{pret}}}
\newcommand{\prompara}{\theta_{\mathrm{pmt}}}
\newcommand{\bs}{N}

\newcommand{\data}{\mathcal{D}}

%% file: 1_intro.tex
\section{Introduction}
\label{sec:intro}


Recent years have witnessed remarkable advancements in artificial intelligence (AI), particularly in the development of Large Language Models (LLMs). One of the standout features of LLMs is their in-context learning ability, where the LLM can perform tasks with only a few-shot examples provided in the prompts, making it possible to deploy LLMs to various applications seamlessly. 

Although most existing research focuses on one-step text generation such as question answering, many real-world scenarios desire autonomous agents that can interact with external environments and make sequential decisions to complete given tasks. 
Some recent works successfully showcase the application of LLMs in sequential decision-making~\citep{shunyu2023react,shinn2023reflexion,liu2023reflect,yang2023auto}, by either directly providing few-shot examples of acting and reasoning in the prompt~\citep{shunyu2023react}, or learning via trial and error during online interactions with the environment~\citep{shinn2023reflexion}.

However, existing methods still struggle to solve many complex domains with LLMs due to the intrinsic difficulties that arise from long-horizon interactive tasks. 
On the one hand, it is widely known that the task complexity increases exponentially with the interaction horizon~\citep{sutton2018reinforcement}, such that a large amount of data or demonstrations can be desired for an agent to fully understand the environment dynamics, especially for heterogeneous and stochastic real-world tasks, where cross-task generalizability is important.
On the other hand, the in-context learning ability is constrained by the limited context window of an LLM. Even if many demonstrations exist, it is hard and inefficient to prompt LLMs with a large number of examples. Although finetuning is a solution, it can be much more expensive and less accessible for normal users. 

\begin{figure*}[!t]
    \centering
    \includegraphics[scale=0.62]{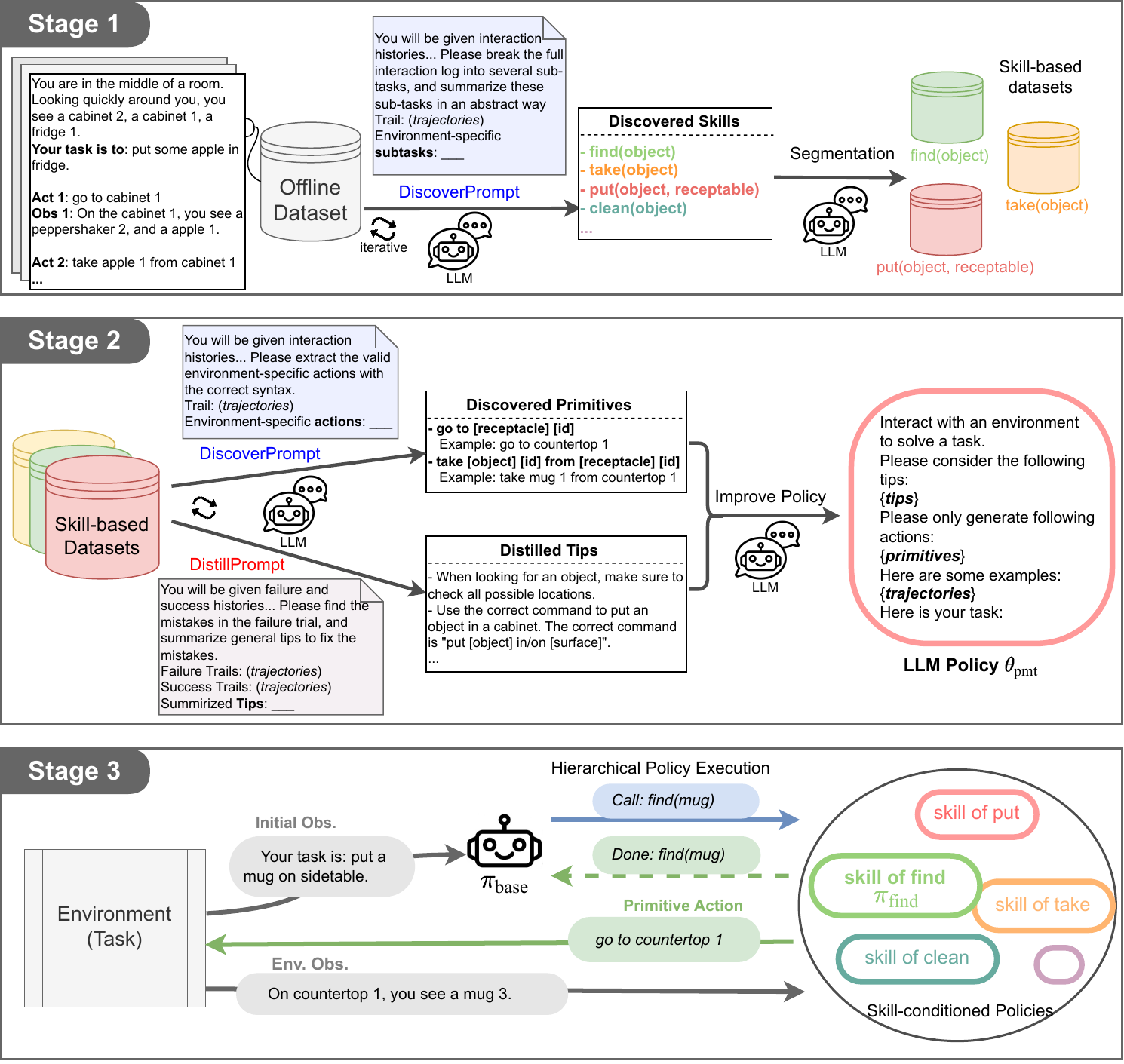}
    \caption{The proposed \ourmod framework. Stage 1: offline skill discovery and data segmentation. Stage 2: offline policy improvement with knowledge distillation. Stage 3: downstream interaction with hierarchical policy execution.}
    \label{fig:method}
\end{figure*}

In response to these challenges, this paper asks and aims to answer the follow question:
\begin{center}
    \emph{Can we develop a data-driven learning framework for sequential decision-making with LLMs, which learns from large-scale offline data without the need of model training?}
\end{center}

In this paper, we define an offline learning framework to enable LLMs to discover and distill useful knowledge from interaction trajectories on multiple tasks.
For LLM-powered agents/policies, we carefully design a generic learning paradigm called \emph{\ourmodfull} (\ourmod), which can iterate over the offline dataset and keep improving the policy.
Importantly, our method does not require a high-quality expert offline dataset, as it can benefit from both positive examples and negative examples of environment interactions, making the framework easier and cheaper to use in practice.
As shown in \Cref{fig:method}, \ourmod is composed of 3 stages: the first stage aims to discover reusable skills by segmenting the offline interaction trajectories; the second stage then conducts skill-conditioned policy improvement by distilling knowledge from offline data; the third stage constructs the interactive policy by calling the learned skills given diverse tasks. 
All stages are based on querying LLMs and iterating over existing datasets, to improve or utilize LLM-powered policies.
As a result, \ourmod can learn better policies from offline dataset at scale without any model finetuning.
Experiments in two commonly used domains (ALFWorld and WebShop) show that our LLM agent augmented by offline knowledge has much better few-shot performance than prior methods on various tasks.

\textbf{Summary of contributions:}
\textbf{(1)} We establish the first offline in-context learning framework for LLM sequential decision-making agents, such that the LLM can learn from offline experience without finetuning. This offline learning paradigm allows more effective usage of past interactions (including both good and bad behaviors) generated by humans or other agents, alleviating the cost of online learning.
\textbf{(2)} Different from prior work which prompts and solves different types of tasks independently, our algorithm leverages offline experience from multiple tasks. By letting LLMs automatically distill shared high-level knowledge, our algorithm achieves few-shot generalization to various types of tasks with a single set of prompts.
\textbf{(3)} \ourmod is a versatile framework that can boost the performance of existing methods. Empirical results show that the knowledge distilled by \ourmod can be combined with existing in-context learning approaches to achieve superior performance as shown in \Cref{sec:exp}.
\ourmod can also be used to facilitate code generation in various tasks, as detailed in \Cref{sec:extension_code}.

%% file: 2_related.tex
\section{Related Work}
\label{sec:related}

{\bf{In-context learning of LLMs for sequential decision-making.}}
{\textit{(1) Generating text.}}
\citet{yang2023auto} and \citet{liu2023agentbench} conduct extensive experiments to showcase the ability of LLMs to make sequential decisions in a range of challenging multi-step reasoning tasks. 
\citet{shunyu2023react} propose ReAct, a method to combine multi-step reasoning and interactive decision-making, which achieves significant improvement in multiple domains.
\citet{shinn2023reflexion} develop agents that can verbally summarize from its past failures and incorporate the reflective text into subsequent trials of the same task, analogous to an online reinforcement learning paradigm. Our method, in contrast, adopts an offline learning method to minimize the cost of online learning and can adapt to various new tasks.
{\textit{(2) Combined with planning.}}
LLMs can also be combined with classical planning approaches, such as Planning Domain Definition Language (PDDL)  \citep{liu2023llm+,silver2023generalized}. 
Differently, our paper focuses on end-to-end LLM policies without additional planning algorithm or knowledge. 
{\textit{(3) Generating code.}} There are a number of recent work showing that embodied agents can be empowered by LLM-written code~\citep{codeaspolicies2022,wang2023voyager,wang2023demo2code}. Although the main focus of this paper is LLM agents that directly interact with environments by language, the proposed method naturally extends to code-based policies. Extensive experiments in \Cref{sec:extension_code} shows that \ourmodcode outperforms existing code-based methods.



{\bf{Training or fine-tuning LLMs for sequential decision-making.}} Another line of related work includes training textual policies with imitation learning \citep{ALFWorld20,ALFRED20} and fine-tuning language models to behave as policies \citep{wang2023voyager, driess2023palm, wang2023describe} in sequential decision-making.
Again, our work is different as it aims at achieving high-quality LLM-based decision-making without fine-tuning.

{\bf{Multi-step reasoning and task decomposition with LLMs.}} 
Multi-step reasoning using LLMs has been widely studied in language domains, including chain-of-thought style reasoning \cite{yao2023tree, fu2022complexity, wei2022chain}, step-by-step feedback based reasoning \cite{lightman2023let, zheng2023progressive}, and self consistency and verification based reasoning \cite{ling2023deductive,wang2022self}.
In contrast, we focus on sequential decision-making tasks in partially observable environments, where each step results in state transitions and only sparse rewards are available.

%% file: 3_method.tex
\section{Methodology}
\label{sec:method}

\subsection{Problem Formulation and Notations}
\label{sec:formulation}

\textbf{LLM-powered policy.}
We first formally define an LLM-powered policy. 
A policy $\pi$ for sequential decision-making is a function that maps the interaction history to a distribution over actions. Let $\pi(a|\tau)$ denote the probability of selection action $a$ given interaction history $\tau$, which is a sequence of all past observations and actions, $\langle o_1,a_1,o_2,a_2,\cdots,o_t\rangle$. 
Then, with a pre-trained LLM which outputs text based on any text input, a policy can be written as
\setlength{\belowdisplayskip}{0pt} \setlength{\belowdisplayshortskip}{0pt}
\setlength{\abovedisplayskip}{2pt} \setlength{\abovedisplayshortskip}{2pt}
\begin{equation}
    \llmpolicy(a|\tau):=LLM(a|\tau; \prompara, \pretpara).
    \label{raw-text-policy}
\end{equation}



The goal is to learn a policy that can maximize the total reward.
During in-context learning, the pre-trained LLM weights $\pretpara$ are fixed, and our \ourmod learns a policy by learning and optimizing the base prompt $\prompara$ from an offline dataset containing multiple interaction trajectories/histories.

\textbf{Skill-conditioned policy.}
A policy can be conditioned on specific skills or subgoals (e.g., find a mug)\citep{xu2023xskill}, which are compositional factors of the original task (e.g., heat some milk). 
Let $\skill$ be a skill, then a skill-conditional policy can be denoted as $\pi^{\skill}$, with $\pi^\skill(a|\tau)$ the probability of selecting action $a$ given history $\tau$ when executing skill $\skill$.

\subsection{Learning from Offline Data}

In many real-world decision-making systems, there exists interaction logs from various users, including experts who can successfully perform the task, as well as non-experts who may fail and make mistakes. Although a final success/failure flag is given for each trajectory, there is no intermediate rewards within the trajectory, making policy learning more difficult. 

Intuitively, seeing the interaction logs from others performing a task can be helpful for one to understand the environment and finish similar tasks. Since LLMs have strong abilities of interpretation and generalization, recent works such as ReAct~\citep{shunyu2023react} have shown that LLMs can solve many interactive decision-making problems when prompted with a few expert demonstrations. This learning paradigm is analogous to behavior cloning~\citep{torabi2018behavioral}, where an agent learns to imitate how experts react to certain scenarios. However, behavior cloning could suffer from the distribution shift between expert demonstrations and the agent's own online interactions, especially when the expert dataset is small and not representative of all scenarios in the domain. Even if we increase the number of demonstrations, the performance is constrained by the limited context window and the high cost of collecting expert demonstrations and annotations (e.g., write reasoning examples). 
In contrast, we introduce an offline learning framework for LLM-powered policies which transfers knowledge across tasks and makes use of large amounts of sub-optimal interaction log to improve downstream decision-making.

\subsection{\ourmod: LLM-Based Offline Policy Improvement}
\label{sec:idea}


Our proposed offline policy learning framework consists of 3 stages, as depicted in \Cref{fig:method}. 
The first stage enables the LLM to discover and abstract reusable skills from offline datasets (potentially from diverse tasks).
Then, the second stage aims to learn a skill-based policy for each discovered skill, through an iterative process of distilling primitives and tips to achieve policy improvement. 
The final stage is to construct the main LLM-powered agent who can reason and call corresponding skills sequentially to solve given tasks. Below we explain each stage in detail.

\textbf{Stage 1: Offline Skill Discovery and Data Segmentation.} 
Many real-world decision-making processes require a number of steps to complete a task, such as controlling a robot to pass several obstacles and navigate to the door, which results in two challenges for LLM-powered agents. First, the limited context length may not be enough to contain the few-shot demonstration and online interaction history. Second, the language model may lose track of its goal and not pay attention to the most important information. 
To mitigate this issue, we propose a hierarchical policy learning framework 
that can iteratively extracts skills from interaction logs with primitive-level executions.
Here the skills are analogous to the options or temporally extended actions~\citep{sutton1999between} in hierarchical reinforcement learning. 
It is well-known that discovering options is difficult in traditional RL, whereas we find that skill discovery with textual logs can be surprisingly well-achieved with the semantic understanding ability of state-of-the-art LLMs. 
Our skill discovery process iterates over the offline trajectories, using a \discoverpmt as shown in Fig.~\ref{fig:method} (Stage 1). The full prompt we use is in \Cref{app:prompts}. 
We ask LLMs
to divide the interaction histories into skill-oriented sub-trajectories, and abstract the skills in function forms. 


\textbf{Stage 2: Offline Policy Improvement with Knowledge Distillation.} 
The main idea of this stage is to distill generalizable knowledge from offline datasets. Such knowledge should be generalizable to tolerate the distribution shift between offline data and online interactions and facilitate downstream tasks.
We aim to distill two types of knowledge from the segmented skill-based trajectories in an iterative process, which leads to improved skill-conditioned policies. 

\textit{$\bullet$ Distilling Primitive Actions.} A common mistake of LLM-powered agents is hallucination, i.e., the agent outputs actions that are not valid in the environment. To ensure effective usage of LLM in decision-making applications, it is important to specify the space of actions in the form of natural language or code. Many prior works~\citep{codeaspolicies2022,liu2023reflect} manually define the available primitive functions, which requires human labor and domain knowledge. Instead, we propose to distill primitive actions or functions from the offline interaction data with LLM, which is easy to scale up and automate the practical operation. Fig.~\ref{fig:method} describes how to distill the primitives with an example, and the full prompt is in \Cref{app:prompts}. 

\textit{$\bullet$ Distilling Policy Improvement Tips with \contrastname.} Inspired by the policy gradient methods in RL, which increases the probability of selecting good actions and lowers the probability of selecting bad ones, we propose to distill knowledge that can enhance good (i.e., achieving high long-term reward) behaviors
and avoid undesired ones in the task distribution.
We propose to distill ``policy improvement tips'' about what actions are preferred under what circumstances. 
However, with the offline data that only provides sequences of interactions and final scores, it is non-trivial for an LLM to figure out the correct credit assignment and useful tips to guide policy improvement. To this end, we propose \textit{\contrastname}, where we sample both successful (high-score) and failed (low-score) trajectories and let the LLM generate tips by contrasting them. As a result, the LLM can identify the key to success and how to avoid failures. 
Fig.~\ref{fig:method} shows the distillation process with a \distillpmt which iteratively updates an LLM-powered policy. More details are in \Cref{sec:imple} and \Cref{app:prompts}.

\begin{figure}
  \begin{center}
    \includegraphics[width=0.5\textwidth]{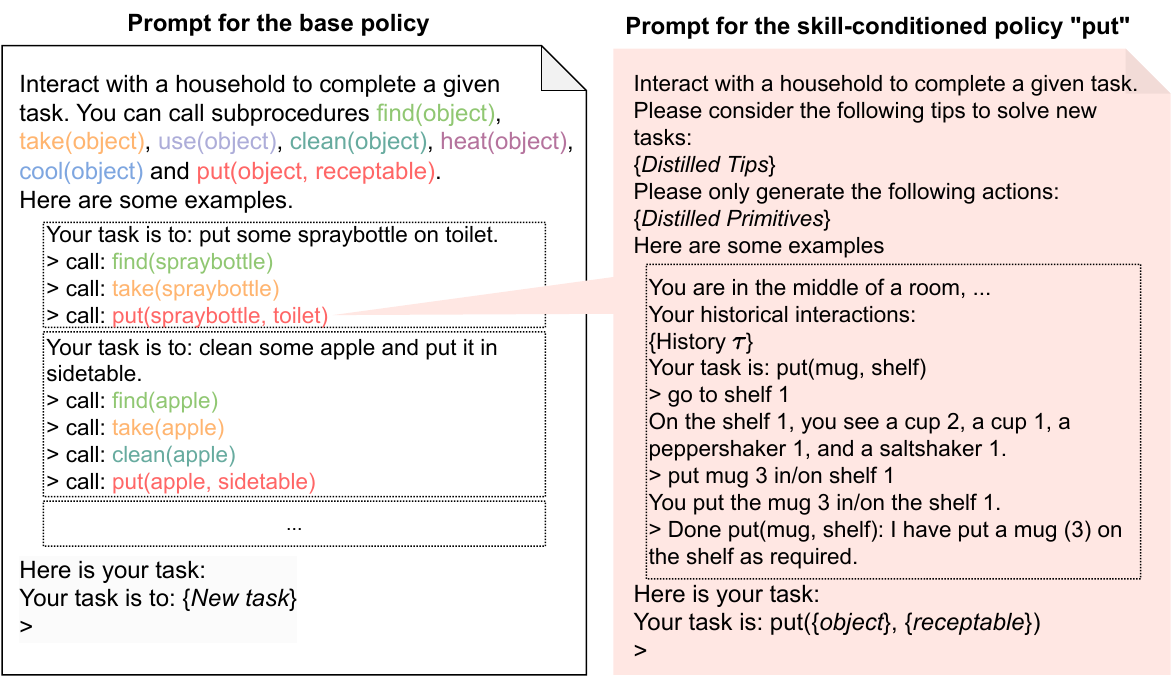}
  \end{center}
  \vspace{-1em}
  \caption{Prompt examples for the base policy and a skill-conditioned policy in hierarchical policy execution.}
  \label{fig:stage3}
  \vspace{-1em}
\end{figure}

\textbf{Stage 3: Downstream Interaction with Hierarchical Policy Execution.} 
So far, Stage 1 discovers a set of skills, while Stage 2 produces and optimizes the corresponding skill-conditioned policies. The final stage is then to compose these skills and interact with the downstream task. 
We prompt a base policy $\basepolicy$ with a few examples (come from LLM's own segmentation of offline trajectories) on calling the proper skills sequentially given a downstream task. 
Therefore, the policy is executed in a hierarchical manner as \Cref{fig:method} shows: the high-level base policy receives the environment and calls the low-level skill-conditioned policies; the called skill-conditioned policy then interacts with the environment and returns a terminate signal when its job is done; the high-level base policy receives the update from the low-level policy and determines whether to execute the next skill.
Fig.~\ref{fig:stage3} shows an example of how to construct the base policy by prompting and how a skill-conditioned policy is prompted when being called. 
\Cref{sec:logs} provides a concrete example of the execution process.

\vspace{-0.5em}
\subsection{Implementation Details of \ourmod}
\label{sec:imple}

The main algorithm is presented in \Cref{alg:main}.
We provide all used prompts and additional implementation details in \Cref{app:prompts}.
In Stage 1 (Line 1-5), we traverse the offline trajectories and ask LLM to summarize potential sub-procedures as skills. Then, based on the discovered skill set, we ask an LLM to segment each trajectory into sub-trajectories, each of which corresponds to a specific skill (Line 5).
In Stage 2 (Line 6-13), the goal is to establish a set of skill-conditioned policies. For each skill, we sample a mini-batch of its (sub-)trajectories, then discover primitive actions and distill policy improvement tips. 
Stage 3 (Line 14-15) is the downstream inference process, where the agent observes the task and sequentially calls the learned skill-conditioned policies to interact with the environment, with few-shot examples of how to call the skills.

\input{./algos.tex}

We also provide an extension of the above method, \ourmodcode, for code generation in sequential decision-making. The details and experiments are in \Cref{sec:extension_code}, where \ourmodcode significantly outperforms existing methods by learning from offline trajectories.

%% file: algos.tex
\begin{figure}[t]
\vspace{-1em}
    \centering
    \begin{algorithm}[H]
\newcommand\mycommfont[1]{\footnotesize\ttfamily\textcolor{gray}{#1}}
\SetCommentSty{mycommfont}
\SetKwFunction{FInit}{InitSkillPolicy}
\SetKwFunction{FImp}{ImprovePolicy}
\SetKwFunction{FCst}{ConstructPolicy}
\SetKwProg{Fn}{Function}{:}{}
    \caption{\ourmodfull}
    \label{alg:main}
    \textbf{Input:} Pre-trained LLM, offline dataset $\data$, batch sizes $\bs_1, \bs_2$, max iteration steps $T_1, T_2$ 
    
    \textbf{Output:} LLM-powered policy $\pi$ 
    
    \tcp{Stage 1: skill discovery}
    Initialize sets of skills $\skills=\emptyset$ 
    
    \For{$t=1,\dots,T_1$}{
        Sample $N_1$ trajectories $d \sim \data$ 
        
        Attempt to discover more skills \\ $\skills \leftarrow LLM(\skills, d; \textcolor{blue}{\mathsf{DiscoverPrompt}})$
    }
    Segment trajectories in $\data$ based on skillset $\skills$, obtain $\data^{\skill_k}$ for each $\skill_k \in \skills$ 
            
    \tcp{Stage 2: knowledge distillation}
    \For{$\skill_k \in \skills$}{
        Initialize primitive set $\prims^{\skill_k}$ and the knowledge set $\tips^{\skill_k}$
        
        Initialize $\pi^{\skill_k} \leftarrow$ Initiate $\prompara$ with $d \sim \data^{\skill}$ and $\prims^{\skill}$ 
        
        \For{$t=1,\dots,T_2$}{
            Sample $N_2$ trajectories $d^{\skill_k} \sim \data^{\skill_k}$
            
            Attempt to discover more primitives $\prims^{\skill_k}\leftarrow LLM(\prims^{\skill_k}, d^{\skill_k}; \textcolor{blue}{\mathsf{DiscoverPrompt}})$ 
            
            Distill policy improvement tips 
            $\tips^{\skill_k} \leftarrow LLM(\tips^{\skill_k}, d^{\skill_k}; \textcolor{red}{\mathsf{DistillPrompt}})$
            
            Ask LLM to incorporate $\tips^{\skill}$ into the prompt of policy $\pi^{\skill}$ 
        }
    }
    \tcp{Stage 3: hierarchical policy execution}
    Sample examples $d \sim \data$ and segment them based on skills
    
    Provide the examples as demonstrations for $\llmpolicy$ to call skill-conditioned policies $\{\pi^\skill\}_{\skill\in\skills}$ given downstream tasks
\end{algorithm}
\vspace{-2em}
\end{figure}

%% file: 4_exp.tex
\section{Experiments}
\label{sec:exp}

\subsection{Experimental Setup}

\textbf{Problem Domains.} We consider two sequential decision-making benchmarks, ALFWorld~\citep{ALFWorld20} and WebShop~\citep{yao2022webshop}.
ALFWorld is a unique environment that mimics household scenarios and allows a decision-making agent to interact with the environment through a text-based interface
We use the same test set as introduced in ALFWorld paper, including 134 tasks in total across six distinct task types. 
Following~\cite{shinn2023reflexion}, we make the problem even more challenging by limiting the horizon of each episode to be 30 (original is 50) steps and terminating the episode if the agent takes the same action twice. 
WebShop provides a real-world online shopping environment with 1.18M products, where an agent must explore the website, check relevant product candidates, and purchase the one that matches a user instruction (e.g., ``I am looking for a queen sized bed that is black, and price lower than 140.00 dollars").
Our evaluation considers the first $500$ out of $12,087$ instructions as test set (following the official implementation~\citep{yao2022webshop}).

\input{tables_comp_react.tex}


\textbf{Metrics.} In ALFWorld, we assess method performance by measuring the success rate (SR) under each task type and a total success rate over 134 tasks. 
Besides the success rate, WebShop sets a product matching score as an extra metric. 

\textbf{Models and Offline Data.} To investigate the robustness of \ourmod across a range of LLMs, we consider two different GPT models in our experiments. 
These include GPT-4-0613 and GPT-3.5-0613,
each providing distinct $\pretpara$ defined in \Cref{raw-text-policy}.
Our offline dataset comprises official human demonstrations in both domains as the success data, and a set of failure data generated by running ReAct on the training task set, as introduced in the original ALFWorld and WebShop implementations (more details are referred to Appendix~\ref{app:offline data}).

\textbf{Baselines and Experiment Organization.}
We evaluate and analyze the proposed \ourmod from multiple angles by comparison with several state-of-the-art baseline methods and their variants. 
We first compare the downstream performance of \ourmod with a well-known baseline ReAct~\citep{shunyu2023react}. 
The results demonstrate the capability of \ourmod to discover and distill knowledge from offline dataset and improve downstream task-solving.
Then, we compare \ourmod with an online learning algorithm Reflexion~\citep{shinn2023reflexion} which interacts with each downstream task for multiple trials and improves the policy by verbally reflecting on its failure history. We also evince the versatility of \ourmod by showing the advantage of combining \ourmod with existing methods (Reflexion$+$\ourmod). 
Furthermore, we conduct comprehensive ablation study for each component of \ourmod including primitives, skills, policy improvement tips, and trajectory contrasting. We also show the comparison with ReAct-long, which inserts as many offline trajectories to the prompt context as possible, to further validate the efficiency of our distillation mechanism.

\subsection{Main Results}

\textbf{Effectiveness of \ourmod in discovering skills and primitives, and distilling policy improvement tips.}
In our experiments, \ourmod can extract high-quality skills from raw successful trajectories in offline data, resulting in seven types of skills under ALFWorld domain, including: \emph{find(object)}, \emph{take(object)}, \emph{put(object, receptacle)}, \emph{cool(object)}, \emph{heat(object)}, \emph{use(object)} and \emph{clean(object)}; and four types of skills under WebShop domain, including \emph{search item}, \emph{select item}, \emph{select item's attributes} and \emph{purchase item}. 
Each skill consists of a set of primitives for execution, accompanied by a group of tips to facilitate skill completion. 
Fig~\ref{skill_example} presents several skill examples from the ALFWorld domain (full results are available in Appendix~\ref{app:skill and knowledge}).
Through learning from offline data, \ourmod is able to capture correct primitives that can be composed to achieve each corresponding skill. 
Importantly, we observe that tips distilled by \ourmod are functional to various degrees, such as suggesting general tips, encouraging exploration, realizing action preconditions and highlighting syntax (as shown in Fig.~\ref{skill_example}). 
To further assess the quality of LLM-distilled knowledge, we designed an oracle baseline, \ourmodhuman, where a domain expert writes down tips and primitives in the prompt according to ReAct's failures.
The comparative results in Appendix~\ref{comp_llm_vs_human} demonstrate that \ourmod is promising to extract human-level knowledge from offline data such as achieving the performance near \ourmodhuman, and even surpassing \ourmodhuman in certain cases.
\begin{figure}[h!]
    \centering
    \includegraphics[scale=0.5]{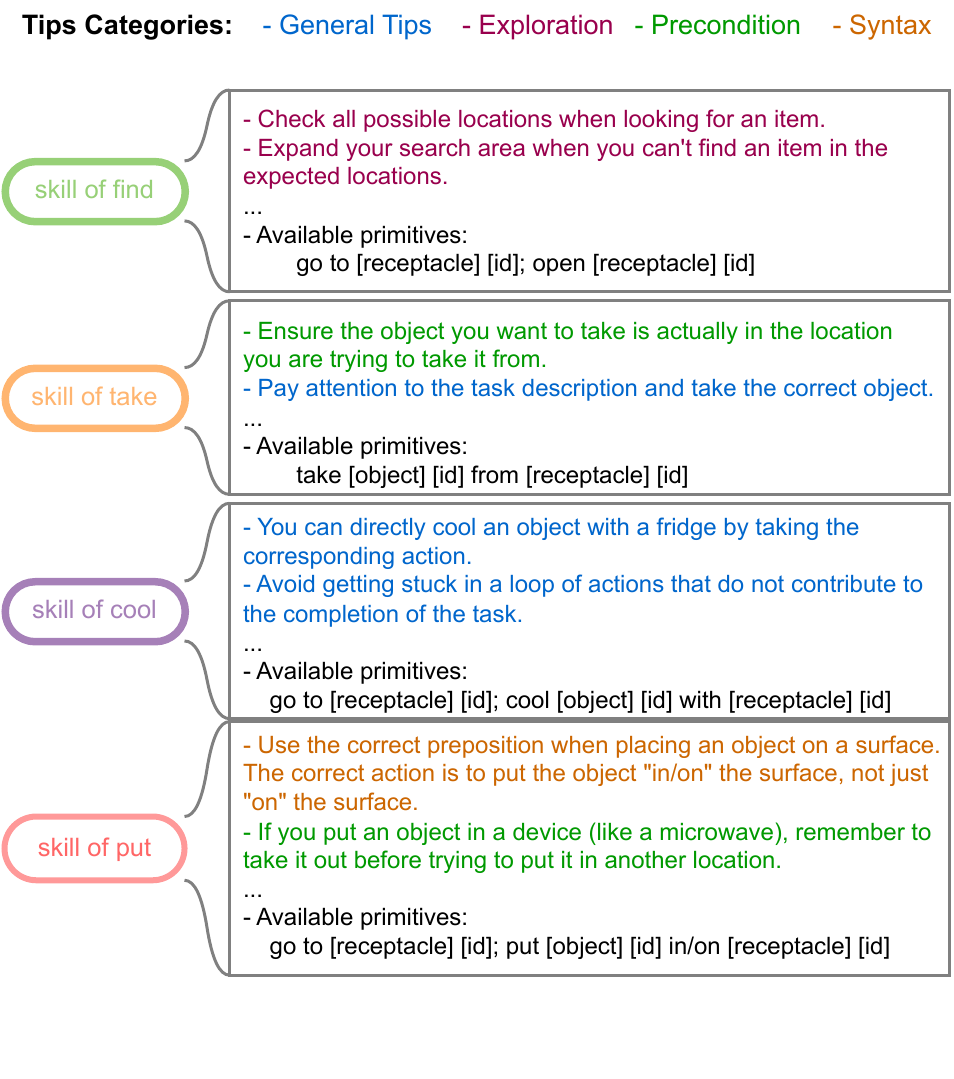}
    \vspace{-2.5em}
    \captionsetup{justification=centering}
    \caption{Examples of discovered skills with primitives and distilled knowledge under ALFWorld.}
    \vspace{-1em}
    \label{skill_example}
\end{figure}

\input{tables_2.tex}

\input{tables_3}

\textbf{\ourmod consistently outperforms baselines across various LLMs under ALFWorld} (Table.~\ref{main_results_alfworld}) and \textbf{WebShop} (Table.~\ref{main_results_webshop}). 
\ourmod achieves higher success rates than ReAct by $19\%$ and $34\%$ in ALFWorld, and $15\%$ and $8\%$ in WebShop, with using GPT-4-0613 and GPT-3.5-0613 respectively.
Furthermore, as shown in Table~\ref{main_results_alfworld}, the success rate achieved by \ourmod in each task category consistently exceeds that of ReAct with the two GPT models. 
This further confirms that the tips distilled by O3D from offline data are generalizable and beneficial across diverse task types. 
For example, the tip ``pay attention to the task description and take the correct object" helps the LLM agent avoid taking a similar object (a pot) rather than the requested one (a pan); and the tip ``The correct action is to put the object 'in/on' the surface, not just 'on' the surface" prevents the LLM agent from making syntactical errors with primitives, which are the two common mistakes made by ReAct.

We also observe that, the overall performance of GPT-3.5-0613 is significantly lower than GPT-4-0613 in the ALFWorld domain. 
Similar poor performance on ALFWorld is also reported by the AgentBench paper~\cite{liu2023agentbench}.
This discrepancy may be attributed to the fact that GPT-3.5-0613 is fine-tuned with data and objectives that are diverged from the ALFWorld domain. \Cref{sec:logs} provide some interaction logs on these tasks and analyze the potential causes of the failure. 
Despite the misalignment of GPT-3.5-0613 with the ALFWorld domain, our approach still achieves $34\%$ improvement over ReAct.

\textbf{Advantages of offline learning over online iterative reflection.} We compare the performance of \ourmod with Reflexion~\citep{shinn2023reflexion} that involves 10 and 3 iterations of online reflection in ALFWorld and WebShop respectively (see Table~\ref{reflexion_alfworld} and Table~\ref{reflexion_webshop}). 
Reflexion allows an LLM agent to reflect on its online interaction history via multiple iterations for performance improvement. 
However, such online interactions in a real-world environment can be expensive or risky. 
In contrast, \ourmod learns from offline data, which may include interaction logs from various users or agents. 
Remarkably, \ourmod outperforms Reflexion in its first deployment trial without additional online reflection.
More importantly, Reflexion learns task-specific textual tips, which only works for the current task it is solving. 
In contrast, \ourmod identifies reusable skills and generalizable textual tips that can facilitate a wide range of downstream tasks.
Also, Reflexion relies on online failure experiences, however, \ourmod offers leverages both successful and failed experiences for improvements, which is more effective in generating proper tips to address mistakes.

\textbf{\ourmod can be combined with existing methods to enhance their performance.}
The comparison between Reflexion and Reflexion+\ourmod validates that, our offline learning framework can improve existing methods by prepending the discovered primitives and distilled tips from offline data into their prompts. 
The results in Table~\ref{reflexion_alfworld} and Table~\ref{reflexion_webshop} show that \ourmod significantly boosts Reflexion’s performance under both ALFWorld and Webshop domains.

\textbf{The extension of \ourmod to code-generation, \ourmodcode, demonstrates advancements comparing with two established baselines}, Demo2Code~\cite{wang2023demo2code} and AdaPlanner~\cite{sun2023adaplanner}, in both domains (see results in Appendix~\ref{app:code_exp}).
On average, \ourmodcode outperforms Demo2Code by $37\%$ and AdaPlanner by $65\%$ in ALFWorld; 
and, in WebShop, \ourmodcode achieves an $18\%$ higher success rate and an improvement of 26 points in the matching score against Demo2Code. 
The major advantage of \ourmodcode lies in its bottom-up code-generation, utilizing diverse offline data to iteratively refine skill functions for robust resue in a higher-level policy composition. 
In contrast, Demo2Code and Adaplanner's top-down, online code generation way often leads to unstable and inconsistent function writings each time without leveraging past experiences or generated functions across tasks, which is a significant limitation of Adaplanner's task-specific code reflection mechanism.
\begin{figure*}[t!]
    \centering
    \includegraphics[scale=0.54]{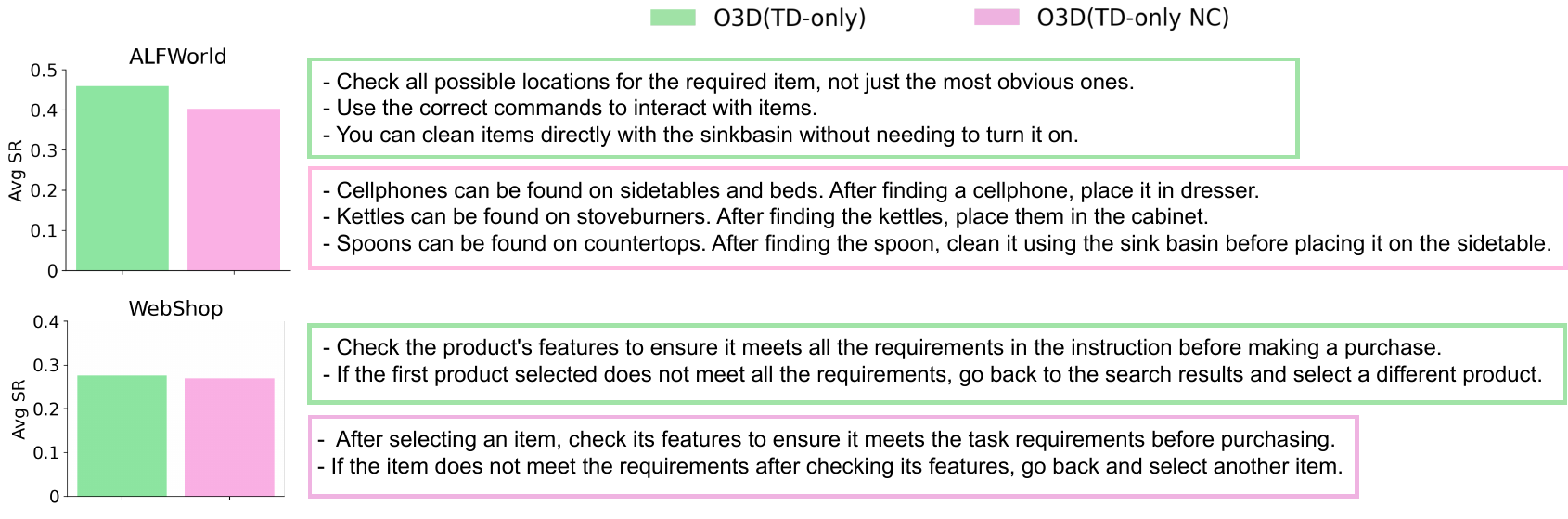}
    \captionsetup{justification=centering}
    \vspace{-2mm}
    \caption{Comparison on averaged success rate over GPT models between using tips distilled via contrastive and non-contrastive (NC) methods, with examples in green and pink boxes respectively.}
    \vspace{-2mm}
    \label{contrast}
\end{figure*}
\begin{figure}[t!]
    \centering
    \includegraphics[scale=0.43]{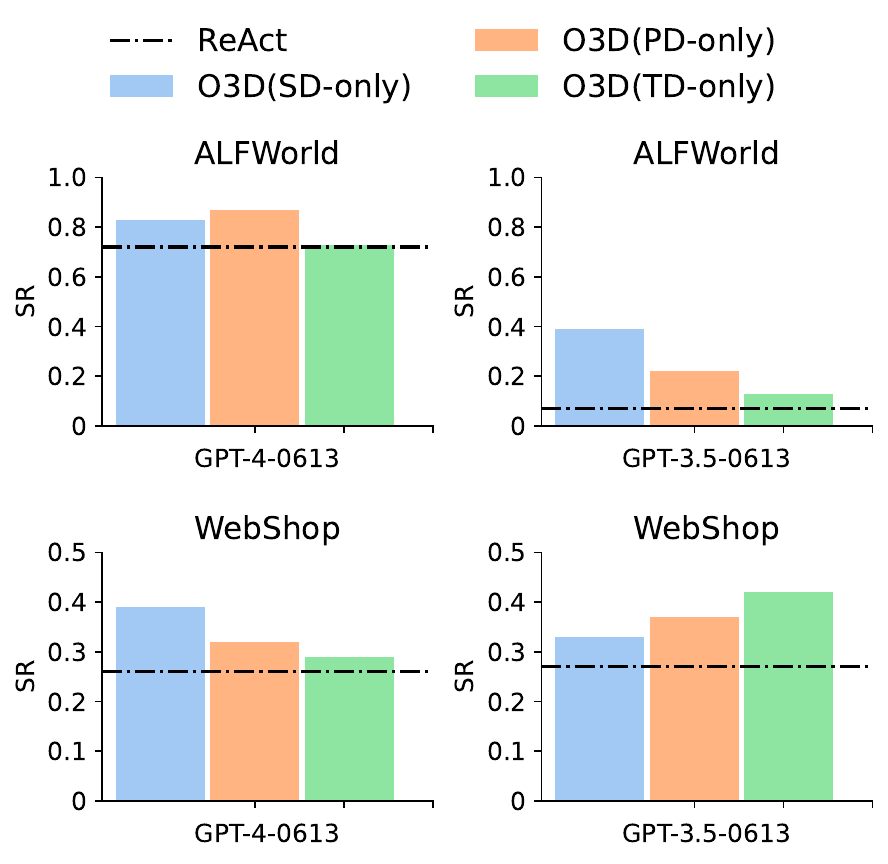}
    \vspace{-1em}
    \captionsetup{justification=centering}
    \caption{Comparison on success rate (SR) with three variants of \ourmod against the baseline method.}
    \label{ablation_study}
    \vspace{-2em}
\end{figure}

\subsection{Ablations}
\textbf{\ourmod is much more effective than naively injecting demonstrations into prompts.} To demonstrate this, we add the same set of offline trajectories into the original ReAct's prompt as many as we can, with GPT-3.5-0613-16k, which is referred to as ReAct-long. 
As results shown in Table~\ref{react_long_alfworld} and Table~\ref{react_long_webshop}, ReAct-long demonstrated improved performance over the original ReAct because of the richer context from the offline trajectories.
Yet, \ourmod's outstanding performance in ALFWorld highlights its superior effectiveness in skill discovery and knowledge distillation from offline data.
In Webshop, \ourmod reaches a higher success rate but a lower score than ReAct-long, because it focuses on distilling tips to purchase the item exactly matching the instruction via success-failure trajectories contrasting, rather than increasing the raw score.
By having the full trajectories with scores, however, ReAct-long has a more fine-grained understanding of the relation between tasks and scores.
A potential improvement for O3D is to contrast trajectories based on a range of scores, rather than just success or failure.
Besides, ReAct-long's long prompt would incur much higher API calling costs for a large number of downstream tasks.

\textbf{Primitives, skills and policy improvement tips independently advance baseline performance.} \ourmod has three major processes: skill discovery (\texttt{SD}), primitives discovery (\texttt{PD}) and policy improvement tip distillation (\texttt{TD}). 
To investigate each component's contribution to performance in downstream task solving, we conducted an ablation study considering three variants of \ourmod, each with only one component. 
Fig.~\ref{ablation_study} shows that, in both domains, the three variants of \ourmod either outperform the baseline or achieve the same performance as the baseline across tested LLM models.
In ALFWorld, \ourmod(\texttt{PD-only}) plays the dominant role in performance improvement with GPT-4-0613. 
This is because the major mistakes made by the baseline are in terms of outputting primitive actions with syntax errors or hallucinating unavailable actions.
\ourmod(\texttt{SD-only}) boosts the performance the most with GPT-3.5-0613, because the tasks in ALFWorld are too complex for ReAct with GPT-3.5-0613, and \ourmod(\texttt{SD-only}) solves the tasks in hierarchy by performing skill selections that greatly reduces the complexity. 
In WebShop, the three components consistently benefit the baseline performance across the two GPT models, with their individual contributions also being model-dependent. 
Since the offline data was collected using GPT-3.5-0613, we observe that the highest overall improvement of the three components occurs in this model.

\textbf{Trajectory contrasting produces domain-dependent advantages over non-contrasting approach.}
Fig.~\ref{contrast} shows that contrasting successful and failed trials in offline data is relatively beneficial in certain domains, compared to the non-contrastive (NC) way based on only successful data.
In ALFWorld, where failures often stem from disregarding domain-specific dynamics and rules, the contrastive method effectively produces generic tips (green box in Fig.~\ref{contrast}) to correct mistakes seen in failure cases.
Conversely, the non-contrastive method merely summarizes the facts (pink box in Fig.~\ref{contrast}) from successful trials, offering limited utility. 
In WebShop, however,
there is often no clear mistake, such as different searching keywords may result in the same products and multiple items fitting the task requirements. Therefore, the advantage of contrastive method is slight, and both methods output analogous tips as shown in Fig.~\ref{contrast}. 

%% file: tables_comp_react.tex
\begin{table*}[ht]
\centering
\caption{Main results on ALFWorld and WebShop in comparison to ReAct.}
\vspace{-0.5em}
\label{tab:main-table}
\begin{subtable}{.64\linewidth}
\centering
\caption{Results in ALFWorld}
\label{main_results_alfworld}
\vspace{-0.5em}
\resizebox{\textwidth}{!}{
\begin{tabular}{c|c|cccccc|c}
\toprule
Model  & Method & Pick & Clean & Heat & Cool & Look & Pick2 & All \\
\midrule
\multirow{ 2}{*}{\shortstack{GPT-4\\(0613)}} & ReAct    & 67 & 74 & 74 & 67 & 100 & 47 & 72\\
                            & \ourmod   & \bf{92} & \bf{100} & \bf{96} & \bf{95} & \bf{100} & \bf{53} & \bf{91}\\
\cmidrule{1-9}
\multirow{ 2}{*}{\shortstack{GPT-3.5\\(0613)}} & ReAct  & 13  & 10  & 0  &  0 & 17 & 0 & 7\\
                            & \ourmod       & \bf{71} & \bf{35} & \bf{4}  & \bf{67} & \bf{44} & \bf{24} & \bf{41}\\
\bottomrule
\end{tabular}
}
\end{subtable}
\hspace{1mm}
\vline
\begin{subtable}{.34\linewidth}
\centering
\caption{Results in WebShop}
\label{main_results_webshop}
\vspace{-0.5em}
\resizebox{0.92\textwidth}{!}{
\begin{tabular}{c|c|c|c}
\toprule
Model  & Method & SR & Score \\
\midrule
\multirow{2}{*}{\shortstack{GPT-4\\(0613)}} & ReAct & 26 & 39 \\
                            & \ourmod& \bf{41} & \bf{58}\\
\cmidrule{1-4}
\multirow{2}{*}{\shortstack{GPT-3.5\\(0613)}} & ReAct & 27 & 60\\
                            & \ourmod                & \bf{35} & \bf{61}\\
\bottomrule
\end{tabular}
}
\end{subtable}%

\vspace{-1mm}
\end{table*}

%% file: tables_2.tex
\definecolor{LightCyan}{rgb}{0.88,1,1}

\begin{table*}[ht]
\centering
\caption{Comparison with online-learning method Reflexion and a combination between Reflexion and \ourmod.}
\vspace{-0.5em}
\label{tab:main-table}
\hspace{-1mm}
\begin{subtable}{.64\linewidth}
\centering
\caption {Results in ALFWorld}
\vspace{-0.5em}
\label{reflexion_alfworld}
\resizebox{\textwidth}{!}{
\begin{tabular}{c|c|cccccc|c}
\toprule
Model  & Method & Pick & Clean & Heat & Cool & Look & Pick2 & All \\
\midrule
\cmidrule{1-9}
\multirow{ 3}{*}{\shortstack{GPT-3.5\\(0613)}} & Reflexion  & 33  & 26  & 26  & 24 & 50 & 18 & 29\\
                            & \ourmod       & \bf{71} & 35 & 4  & 67 & 44 & 24 & 41\\
                            & Reflexion$+$\ourmod       & 54 & \bf{35} & \bf{61}  & \bf{67} & \bf{56} & \bf{29} & \bf{50}\\
\bottomrule
\end{tabular}
}
\end{subtable}
\hspace{0.5mm}
\vline
\hspace{0.5mm}
\begin{subtable}{.34\linewidth}
\centering
\caption {Results in WebShop}
\vspace{-0.5em}
\label{reflexion_webshop}
\resizebox{\textwidth}{!}{
\begin{tabular}{c|c|c|c}
\toprule
Model  & Method & SR & Score \\
\midrule
\cmidrule{1-4}
\multirow{ 3}{*}{\shortstack{GPT-3.5\\(0613)}} & Reflexion & 35 & 55 \\
                            & \ourmod & 35 & 61\\
                            & Reflexion$+$\ourmod & \bf{46} & \bf{62}\\
\bottomrule
\end{tabular}
}
\end{subtable}
\end{table*}

%% file: tables_3.tex
\begin{table*}[ht]
\centering
\caption{Comparison with ReAct-long which naively injects demonstrations into prompts.}
\vspace{-0.5em}
\label{tab:main-table}
\hspace{-1mm}
\resizebox{1.0\textwidth}{!}{
\begin{subtable}{.64\linewidth}
\centering
\caption {Results in ALFWorld}
\vspace{-0.5em}
\label{react_long_alfworld}
\resizebox{\textwidth}{!}{
\begin{tabular}{c|c|cccccc|c}
\toprule
Model  & Method & Pick & Clean & Heat & Cool & Look & Pick2 & All \\
\midrule
\cmidrule{1-9}
\multirow{ 3}{*}{\shortstack{GPT-3.5\\(0613)}} & ReAct  & 13  & 10  & 0  &  0 & 17 & 0 & 7 \\
                            & ReAct-long  & 71  & 6  & 0  & 5 & 17 & 24 & 21 \\
                            & \ourmod       & \bf{71} & \bf{35} & \bf{4}  & \bf{67} & \bf{44} & \bf{24} & \bf{41}\\
\bottomrule
\end{tabular}
}
\end{subtable}
\hspace{0.5mm}
\vline
\hspace{0.5mm}
\begin{subtable}{.33\linewidth}
\centering
\caption {Results in WebShop}
\vspace{-0.5em}
\label{react_long_webshop}
\resizebox{\textwidth}{!}{
\begin{tabular}{c|c|c|c}
\toprule
Model  & Method & SR & Score \\
\midrule
\cmidrule{1-4}
\multirow{ 3}{*}{\shortstack{GPT-3.5\\(0613)}} & ReAct & 27 & 60\\
                            & ReAct-long & 30 & \bf{64} \\
                            & \ourmod & \bf{35} & 61\\
\bottomrule
\end{tabular}
}
\end{subtable}
}
\end{table*}

%% file: 5_discuss.tex
\vspace{-0.5em}
\section{Conclusion}
\vspace{-0.5em}
This paper introduces an offline in-context learning framework, \ourmod, for sequential decision-making, where LLM agents can learn from previous experiences in a scalable offline manner to improve performance without model fine-tuning.
\ourmod stands out by allowing LLMs to distill shared high-level knowledge from offline interaction logs, which is injected into a single set of prompts to be reused in solving diverse downstream tasks. 
Empirically, \ourmod outperforms baseline methods in two challenging benchmark domains, and it is compatible with existing LLM-based methods for sequential decision-making.
Our work offers a possibility for future LLM-based algorithm development in terms of efficiently leveraging offline data at scale for real-world sequential decision-making applications.
\label{sec:discuss}

%% file: 6_appendix.tex

\startcontents[appendices]
\printcontents[appendices]{}{0}{\section*{Appendix}}

\section{Details of Implementation and Extra Results}

\subsection{Offline Data Collection and Usage}
\label{app:offline data}
The offline dataset used in our experiment consists of both success and failure interaction trajectories. The success data in ALFWorld are gathered from a 'traj.json' file in the training dataset provided on their official website. We download human demonstrations from the official website of WebShop, and we select out the demonstrations with 100$\%$ product matching score from the training dataset as the success data. 
We generate the failure data by running ReAct with GPT-3.5-0301 on ALFWorld and running ReAct with GPT-3.5-0613 on the WebShop, which are both conducted on the training task set. 
Eventually, we have totally 80 pairs of success trajectories and failure trajectories on the same tasks in ALFWorld and 90 pairs of success trajectories and corresponding failure trajectories in WebShop.

In Table~\ref{hyper_param}, we detail the usage of the offline data for discovering primitives and skills and distilling policy improvement tips.
We use GPT-4-0613 for all discovery and distillation processes. In our experiments, we note that the batch size may impact the quality of results. The tips distillation in \ourmod is skill-oriented, where we need contrastive data pairs for each skill. Among the first 30 offline data pairs, certain skills have zero failure data (e.g., cool) as the mistakes happened in other skills. We, therefore, generate more offline data (80) for the tips distillation part of \ourmod.

\begin{table}[h!]
\captionsetup{justification=centering}
\caption {Details of offline data usage in ALFworld}
\label{table:majorresults}
\centering
\begin{tabular}{c|c|c|c|c}
\toprule
  Method & Component & Data Type & Batch Size & \# of Trajs \\
\midrule
\multirow{ 3}{*}{\ourmod}&Skill Discovery & success & 1 & 6\\
\cmidrule{2-5}
&Primitives Discovery & success & 1 & 12\\
\cmidrule{2-5}
& Tips Distillation & success $\&$ failure & 1 & 80 $\times$ 2\\
\cmidrule{1-5}
\ourmod\texttt{(SD-only)} & Skill Discovery & success & 1 & 6 \\
\cmidrule{1-5}
\ourmod\texttt{(PD-only)} & Primitives Discovery & success & 1 & 12\\
\cmidrule{1-5}
\ourmod\texttt{(TD-only)} & Tips Distillation & success $\&$ failure & 2 & 30 $\times$ 2\\
\cmidrule{1-5}
\ourmod\texttt{(TD-only NC)} & Tips Distillation & success & 2 & 30\\
\bottomrule
\end{tabular}
\label{hyper_param}
\end{table}

\begin{table}[h!]
\captionsetup{justification=centering}
\caption {Details of offline data usage for in WebShop}
\label{table:majorresults}
\centering
\begin{tabular}{c|c|c|c|c}
\toprule
  Method & Component & Data Type & Batch Size & \# of Trajs \\
\midrule
\multirow{ 3}{*}{\ourmod}& Skill Discovery & success & 2 & 30\\
\cmidrule{2-5}
&Primitives Discovery & success & 1 & 90\\
\cmidrule{2-5}
& Tips Distillation & success $\&$ failure & 1 & 30 $\times$ 2\\
\cmidrule{1-5}
\ourmod\texttt{(SD-only)} & Skill Discovery & success & 2 & 30\\
\cmidrule{1-5} 
\ourmod\texttt{(PD-only)} & Primitives Discovery & success & 1 & 90\\
\cmidrule{1-5}
\ourmod\texttt{(TD-only)} & Tips Distillation & success $\&$ failure & 1 & 30 $\times$ 2\\
\cmidrule{1-5}
\ourmod\texttt{(TD-only NC)} & Tips Distillation & success & 1 & 30\\
\bottomrule
\end{tabular}
\label{hyper_param}
\end{table}

\clearpage
\section{Comparison between LLM-Generated Knowledge and Human-Generated Knowledge}
\label{comp_llm_vs_human}

\input{tables_comp_human.tex}
\begin{table*}[h!]
\caption{Tips of the ``\emph{put(object. receptacle)}" skill in ALFWorld}
\centering
\begin{tabular}{c|p{12cm}}
\toprule
Method & Tips \\
\cmidrule{1-2}
 LLM & 1. Use the correct preposition when placing an object on a surface. The correct action is to put the object "in/on" the surface, not just "on" the surface.\newline 
 2. Place the object in the first available and suitable location instead of unnecessarily checking other locations.\newline 
 3. If you put an object in a device (like a microwave), remember to take it out before trying to put it in another location.\\
\cmidrule{1-2}
 Human & 1. You can directly put an object in/on an occupied receptacle. \newline
 2. Please strictly follow the syntax of primitives.\\
\bottomrule
\end{tabular}
\label{com_tip_alfworld}
\end{table*}
\begin{table*}[h!]
\caption{Knowledge of the ``\emph{select item's attributes}" skill in WebShop}
\centering
\begin{tabular}{c|p{10cm}|p{2cm}}
\toprule
Method & Tips & Primitives\\
\cmidrule{1-3}
 LLM & 1. Ensure to select all the necessary attributes as per the instruction. \newline 2. Be aware that color and size options may not always be straightforward and may include additional information or codes. \newline 3. Be careful with the case and spacing of the words when selecting item attributes. & click[Attribute] click[Color] click[Style] click[Size] click[Flavor] \\
\cmidrule{1-3}
 Human & 1. Remember to select all required attributes using click[Attributes] before click[Buy Now]. \newline 2. You should pay attention to the current observation to check clickable buttons. & click[Attribute]\\
\bottomrule
\end{tabular}
\label{com_tip_webshop}
\end{table*}

\textbf{\ourmod is promising to extract human-level knowledge from offline data such as achieving the performance near \ourmodhuman, and even surpassing \ourmodhuman in certain cases.}
\ourmodhuman is designed as an oracle, where a domain expert writes down tips and primitives in the prompt according to ReAct's failures.
We noticed that human often gives more general tips/primitives, while LLMs can distill both general and granulated ones because of the contrastive way over success-failure pairs. 
Given the same sets of LLM-generated knowledge and human-generated knowledge to three different models, as Table 6 shown, which type of knowledge is better depends on the model's capability, such as \ourmod outperforming \ourmodhuman when using GPT-4 in ALFWorld and GPT-3.5-0613 in WebShop, but being relatively worse than \ourmodhuman on other models. 
Please see the concrete examples in Table.~\ref{com_tip_alfworld} and Table.~\ref{com_tip_webshop}, where the LLM-generated knowledge is more comprehensive than those generated by humans.

In Table.~\ref{com_tip_alfworld}, the LLM-generated tips are more comprehensive such as the concrete in-line examples of obeying action syntax; the suggestion for avoiding redundant behaviors; and the hint on the action precondition. We think this is why \ourmod always outperformed \ourmodhuman on the ``Pick" task across three GPT models.

In Table.~\ref{com_tip_webshop}, the LLM-genreated tips offer detailed guidance on aspects to pay attention during attribute selection, along with more granular primitives being discovered to assist the agent in focusing on various types of attributes.

Overall, the comparison between \ourmod and \ourmodhuman verifies that LLMs can distill tips and discover primitives as good as human-generated ones.
Please see Appendix~\ref{app:human knowledge} and Appendix~\ref{app:knowledge} for the full tips and primitives of human-generated and LLM-generated.

\section{Additional Results}

In this section, we present additional experimental results in ALFWorld and WebShop domains.

\subsection{ALFWorld}

\begin{figure}[h!]
    \centering
    \includegraphics[scale=0.38]{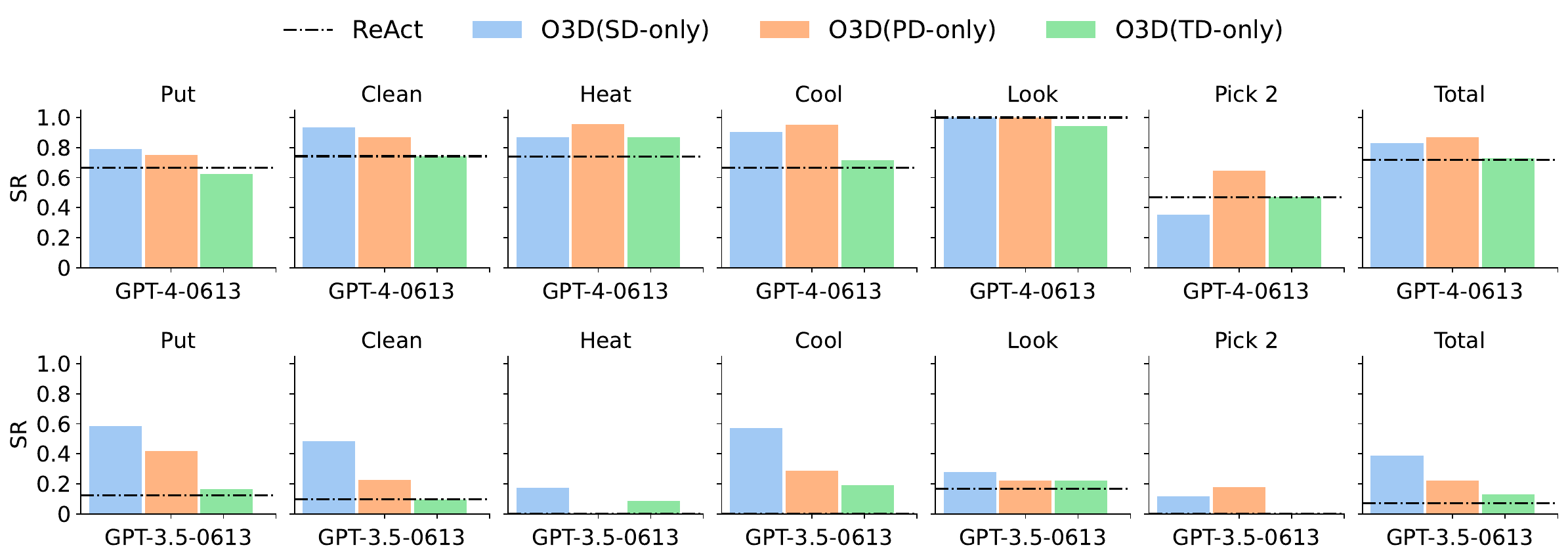}
    \captionsetup{justification=centering}
    \caption{Comparison on success rate (SR) with three variants of \ourmod against the baseline method.}
    \label{app:full_ablation}
\end{figure}

In our ablation study, we also further visualize the performance of the three variants of \ourmod under each type of task with three different models, as shown in Fig~\ref{app:full_ablation}. From the perspective of each task type, the advantage of the three variants of \ourmod versus ReAct is a bit model-dependent, but overall, they outperform ReAct as shown in the last column of Fig.~\ref{app:full_ablation}.


\subsection{WebShop}

\begin{figure}[h!]
    \centering
    \includegraphics[scale=0.4]{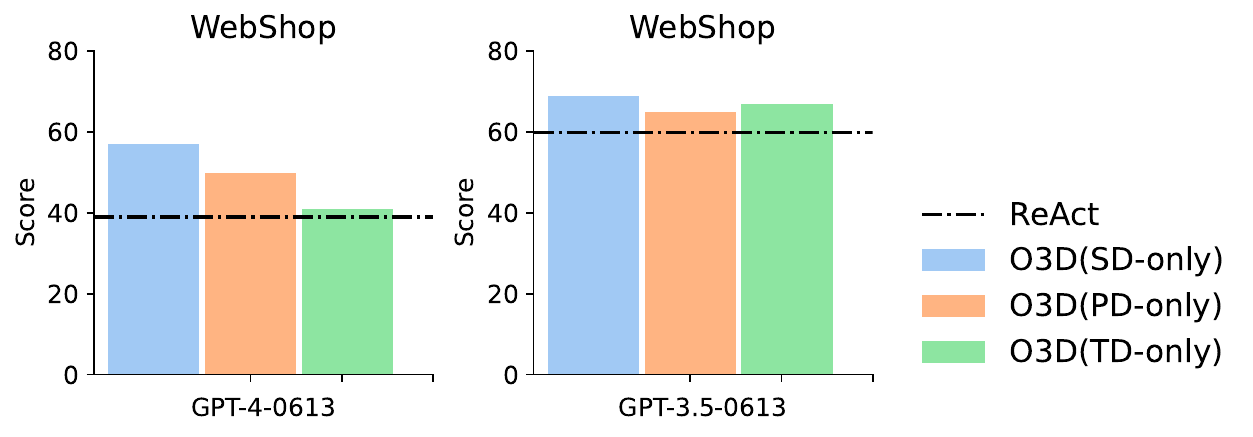}
    \caption{Comparison on the product matching score with three variants of \ourmod against the baseline method.}
    \label{app:webshop_score}
\end{figure}

We show the performance of the three variants of \ourmod on the product matching score against the baseline in Fig~\ref{app:webshop_score}. With GP4-0613 and GPT-3.5-0613, the three variants of \ourmod all outperform ReAct, which validates the contribution of the offline data-driven discovery and distillation to policy improvement. 

\clearpage
\section{Extension of \ourmod to Code-Generation}
\label{sec:extension_code}

\subsection{Methodology}

\input{a1_code}

\subsection{Experiments}
\label{app:code_exp}

In this section, we present additional experimental results of \ourmodcode in ALFWorld and WebShop domains.

\subsubsection{ALFWorld}

We show the comparison of \ourmodcode against two state-of-the-art code-as-policy baselines, Demo2Code~\cite{wang2023demo2code} and AdaPlanner~\cite{sun2023adaplanner} in Table.~\ref{app:o3d-code-alfworld}.

\begin{table}[h!]
\captionsetup{justification=centering}
\caption {Results of \codepolname in ALFWorld}
\label{table:majorresults}
\centering
\begin{tabular}{c|c|cccccc|c}
\toprule
Model  & Method & Pick & Clean & Heat & Cool & Look & Pick2 & SR \\
\midrule
\multirow{ 2}{*}{\shortstack{GPT-4-0613}} & Demo2Code    & 96       & 58      & 13      & 43      & 0       & 65      & 48\\
                                             & \ourmodcode  & \bf{100} & \bf{84} & \bf{87} & \bf{90} & \bf{89} & \bf{88} & \bf{90}\\
\midrule
\multirow{ 3}{*}{\shortstack{GPT-3.5-0613}} & Demo2Code & 96  & 26   & 48   & 29  & 0  & \bf{82} & 46\\
                                & AdaPlanner & 4 & 13 & 0 & 0 & 0 & 0 & 13 \\
                                & \ourmodcode               & \bf{100}  & \bf{71} & \bf{91} & \bf{86} & \bf{89} & 18 & \bf{78}\\
\bottomrule
\end{tabular}
\label{app:o3d-code-alfworld}
\end{table}

\ourmodcode achieves an impressive leading performance against Demo2Code by $42\%$ with GPT-4-0613, $32\%$ with GPT-3.5-0613, and outperforms AdaPlanner by $65\%$ with GPT-3.5-0613.
The principal advantage of our approach lies in its unique method of generating code: it adopts a bottom-up style, effectively constructing policies on top of robust skill functions. Through iterative skill refinement, the model cultivates robust skills by leveraging extensive and diverse data from demonstrations. Skill functions can then be efficiently utilized and reused to compose higher-level policies. In contrast, Demo2Code follows a top-down approach, requiring the generation of code for the same set of skills each time it receives a new task. Due to the context length constraint inherent in LLMs, only a limited number of demonstrations are used to guide skill code generation in Demo2Code, resulting in unstable and inconsistent skills. 
Although Adaplanner incorporates an online reflection process for code correction, typically over 6 iterations as its official setup, it still falls short in terms of success rate compared with \ourmodcode. Notably, its performance is significantly worse than \ourmodcode when utilizing GPT-3.5-0613.
This comparison further confirms the superiority of \ourmodcode's bottom-up way for code-generation and the iterative skill refinement over offline data.

\begin{figure}[h!]
    \centering
    \includegraphics[scale=0.32]{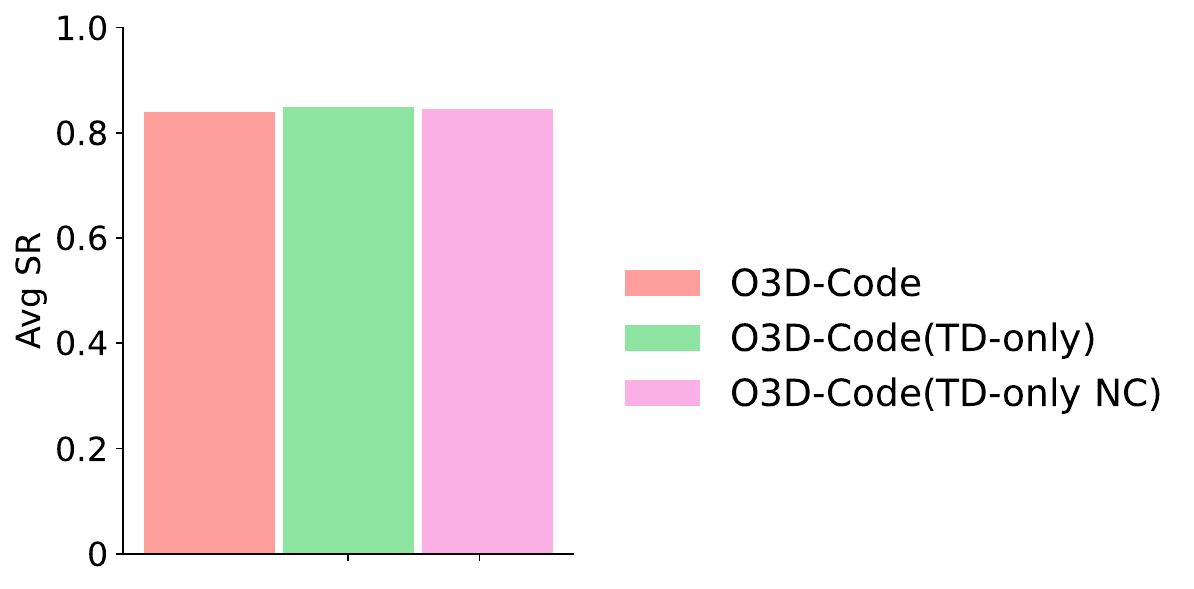}
    \caption{Results of \ourmodcode with the tips distilled via contrastive and non-contrastive ways against the original \ourmodcode in ALFWorld. We use the averaged success rate over the three GPT models as the performance of each method.}
    \label{app:code_contrast_alfworld}
\end{figure}

We also conduct experiments to investigate whether the policy improvement tips distilled for text-based policy can improve the performance of code-based policy or not. As the results shown in Fig.~\ref{app:code_contrast_alfworld}, in ALFWorld, the tips distilled with the contrastive approach and the non-contrastive approach both have a minor contribution to the improvement of code-based policy.

\subsubsection{WebShop}

We show the comparison of \ourmodcode against Demo2Code~\cite{wang2023demo2code} in Table.~\ref{app:o3d-code-webshop}. (Note that AdaPlanner does not provide a solution for WebShop, thus we did not compare with it.)

\begin{table}[h!]
\captionsetup{justification=centering}
\caption {Results of \codepolname in WebShop}
\label{table:majorresults}
\centering
\begin{tabular}{c|c|c|c}
\toprule
Model  & Method & SR & Score \\
\cmidrule{1-4}
\multirow{ 2}{*}{\shortstack{GPT-4-0613}} & Demo2Code & 1 & 5\\
                            & \ourmodcode & \bf{19} & \bf{31}\\
\bottomrule
\end{tabular}
\label{app:o3d-code-webshop}
\end{table}

As the results shown in Table.~\ref{app:o3d-code-webshop}, Webshop poses a substantial challenge for code-based policies owing to its need for comprehensive natural language understanding within the environment feedback. We address this challenge by enabling LLMs to construct skills by employing a limited set of LLM-based functions that encapsulate the underlying LLM capabilities (see Appendix~\ref{app:code:webshop}). To ensure a fair comparison, we also offer the same set of functions to Demo2Code. While our approach substantially enhances the performance of code-based policies in comparison to the baseline, it's important to note that skill generation remains considerably constrained by the intricate and diverse text-based environment feedback in Webshop. 

\begin{figure}[h!]
    \centering
    \includegraphics[scale=0.32]{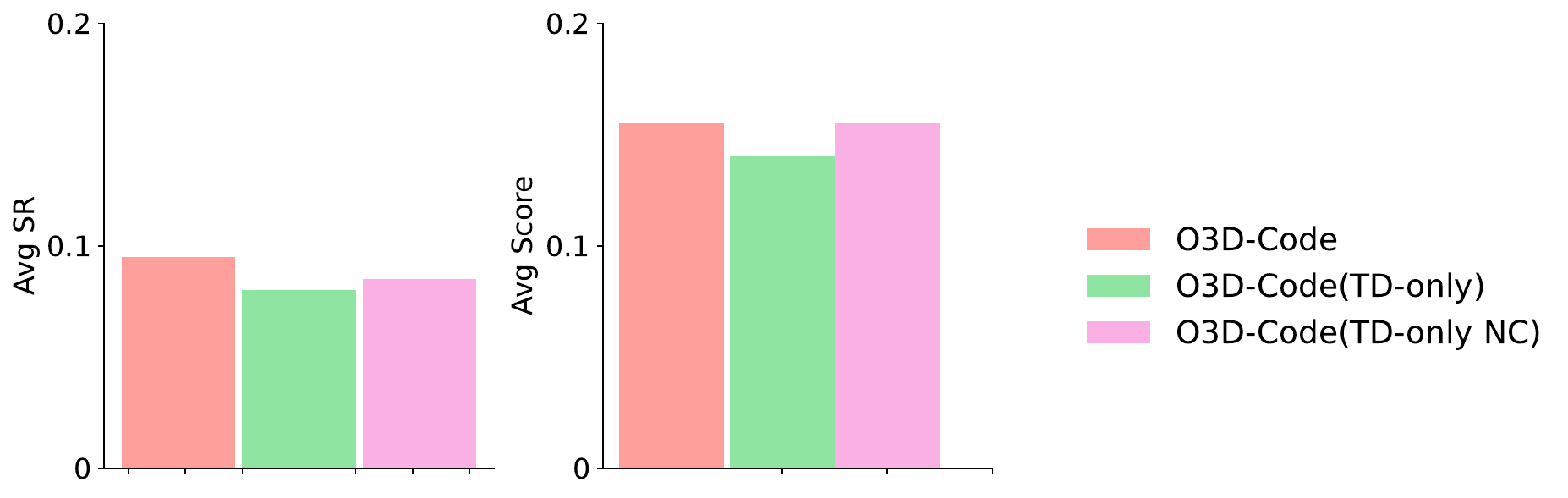}
    \caption{Results of \ourmodcode with the tips distilled via contrastive and non-contrastive ways against the original \ourmodcode in WebShop. We measure the averaged success rate and the averaged product matching score over GPT-4-0613 and GPT-3.5-16K-0613 as the performance of each method.}
    \label{app:code_contrast_webshop}
\end{figure}

Additionally, in WebShop, as the results shown in Fig~\ref{app:code_contrast_webshop}, the tips distilled for text-based policy is not helpful for the improvement of code-based policy.

\clearpage
\subsection{Detailed Interaction Logs: Examples and Comparisons}
\label{sec:logs}

\begin{figure}[htbp]
    \centering
    \includegraphics[scale=0.5]{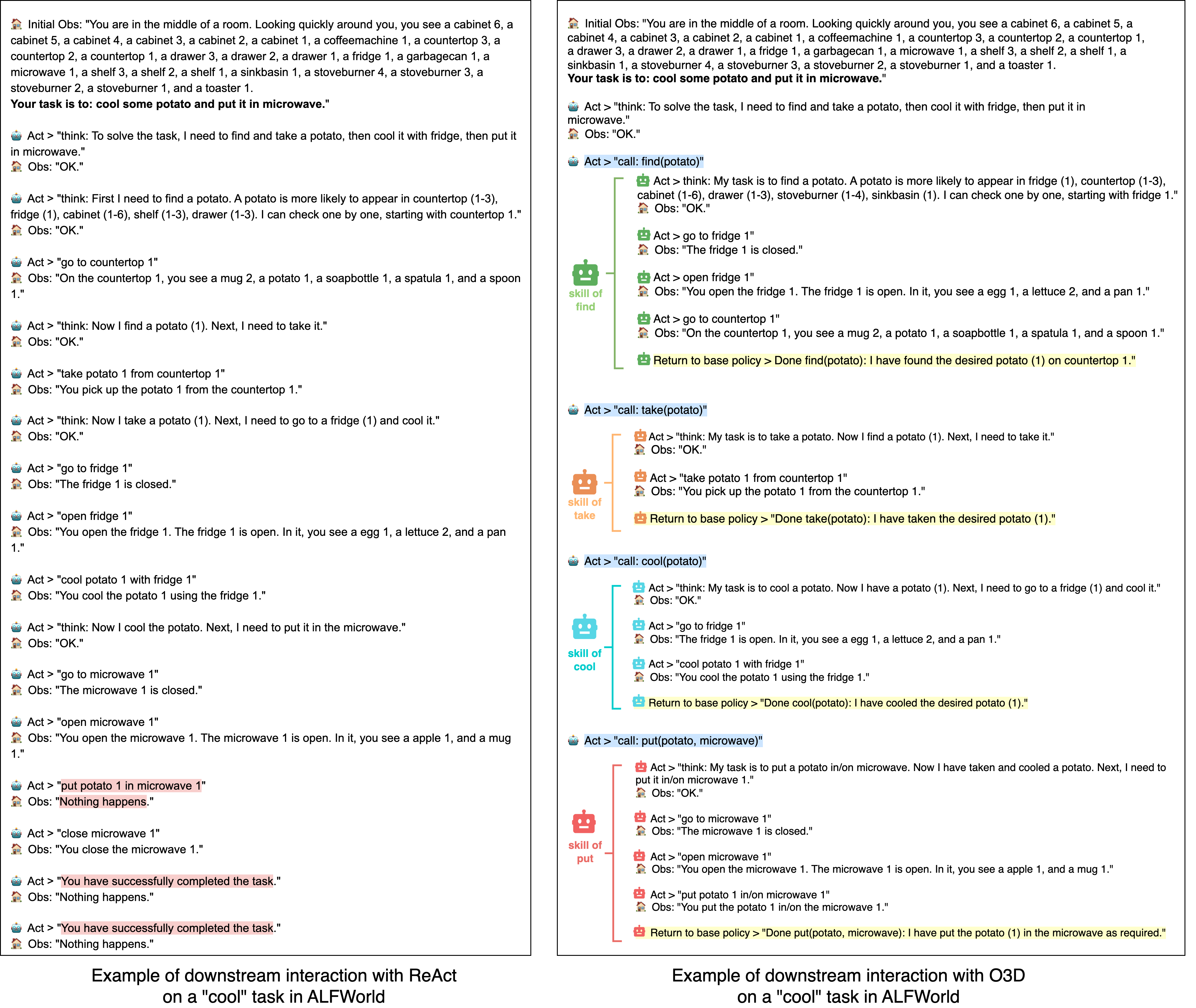}
    \caption{Interaction logs of ReAct (left) and \ourmod (right) in the same task with GPT-4.}
    \label{fig:example_log}
\end{figure}

In \Cref{fig:example_log}, we visualize the downstream interaction logs of both ReAct and our \ourmod on the same task from ALFWorld, to better illustrate why \ourmod has advanced performance. We can see that ReAct agent makes a syntax mistake when executing the action of putting a potato. As a result, the invalid action does not work in the environment, but the agent does not realize the error and believes it has successfully finished the task. In contrast, \ourmod executes the policy in a hierarchical way, and each skill-conditioned policy works as expected. Also, the "put" agent can successfully avoid the syntax error as suggested by the distilled knowledge.

\clearpage

\begin{figure}[htbp]
    \centering
    \includegraphics[scale=0.5]{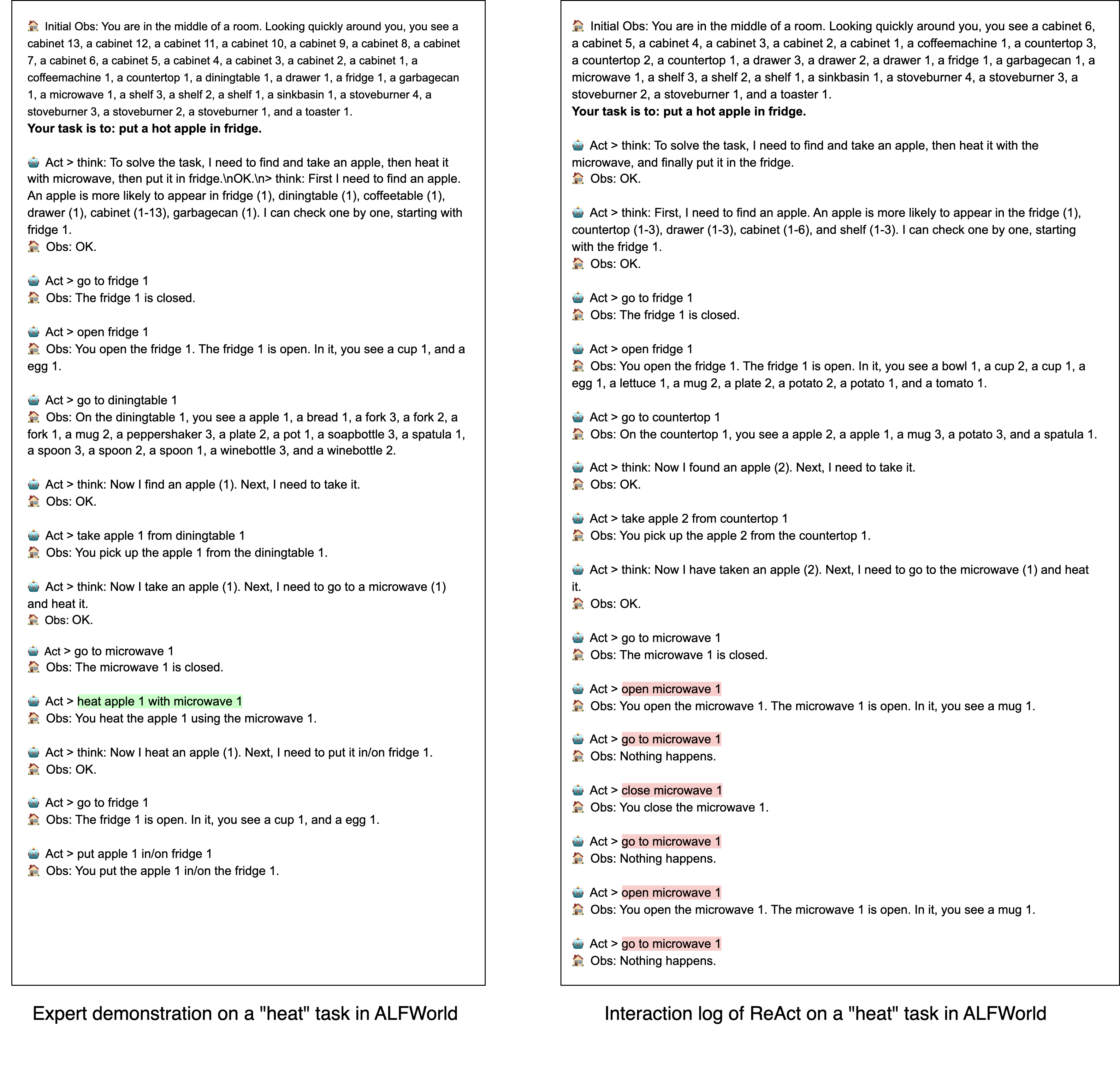}
    \caption{Expert demonstrations for a "heat" task in ALFWorld (left) and a failure trajectory of ReAct with GPT-3.5-0613 on a "heat" task given the expert demonstration in its prompt (right).}
    \label{fig:example_heat}
\end{figure}

In experiments, we observe that the decision-making performance can be significantly different in different GPT versions. For example, GPT-3.5-0613 struggles to solve some ALFWorld tasks (ReAct with GPT-3.5-0613 fails on all Heat and Cool tasks as shown in \Cref{tab:main-table}). 
To better understand this phenomenon, we examine the downstream interaction log of the agent in these tasks.
\Cref{fig:example_heat} (left) shows the expert demonstration in a task where the agent heats an apple and puts it in a fridge, which is the example provided by ReAct~\cite{shunyu2023react} in its base prompt. Given this example, however, we find that the ReAct agent with GPT-3.5-0613 model does not successfully solve the downstream task that is very similar to the expert example. When the agent finds that the microwave is closed, it does not follow the expert demonstration to heat the apple. Instead, it tries to open the microwave and retry going to the microwave. The agent finally gets stuck in a dead loop of going to the microwave and opening it. This is a common problem of GPT-3.5-0613 across many test tasks, showing the difficulty of letting LLM agent solve sequential decision-making problems. 

\clearpage
\section{Prompts and Learned Knowledge}
\label{app:prompts}



\subsection{Text-based Policy}
\subsubsection{Skill Discovery Prompt}
\input{prompts/ALFworld/task_decompo_prompt_v0}

\input{prompts/Webshop/task_decomp_prompt_v0}

\subsubsection{Primitive Discovery Prompt}
\input{prompts/ALFworld/all_subtask_primitive_prompt}

\input{prompts/Webshop/all_subtask_primitive_prompt_v1}

\subsubsection{Skill-conditioned Policy Improvement Tips Distillation Prompt}
\input{prompts/ALFworld/subtask_knowledge_prompot_contrast_v1}

\input{prompts/ALFworld/subtask_knowledge_prompot_v1}

\input{prompts/Webshop/subtask_knowledge_prompt}

\input{prompts/Webshop/subtask_knowledge_prompt_non_contrst}





\subsubsection{Prompts for Downstream Task Solving}
\input{prompts/ALFworld/hier_llm_llmdk_downstream}

\input{prompts/Webshop/hier_llm_llmdk_downstream}

\subsection{Code-based Policy}
\input{prompts/ALFworld/skill_discovery_prompt}
\input{prompts/Webshop/skill_discovery_prompt}

\subsection{Human-Generated Policy Improvement Tips and Primitives for Each Skill}
\label{app:human knowledge}
\subsubsection{ALFWorld}
\input{prompts/ALFworld/hier_llm_hdk}
\subsubsection{WebShop}
\input{prompts/Webshop/hier_llm_hdk}

\subsection{LLM-Distilled Policy Improvement Tips and Primitives for Each Skill}
\label{app:knowledge}
\label{app:skill and knowledge}
\subsubsection{ALFWorld}
\input{prompts/ALFworld/hier_llm_llmdk}

\subsubsection{WebShop}
\input{prompts/Webshop/hier_new_llmdk}

\subsection{Ablation Study and Discussion on Policy Improvement Tips}

\subsubsection{Tip Distillation Only (\texttt{TD-Only}) Prompts}
\input{prompts/ALFworld/knowledge_prompt_v1}

\input{prompts/ALFworld/knowledge_prompt_non_contrast_v1}

\input{prompts/Webshop/knowledge_prompt_v1}

\input{prompts/Webshop/knowledge_prompt_non_contrast}

\subsubsection{Results in ALFWorld}
\input{prompts/ALFworld/llm_dk_v2}

\input{prompts/ALFworld/llm_dk_noncontrast_v2}


\input{prompts/ALFworld/knowledge_exp_with_primitives}

\subsubsection{Results in WebShop}
\input{prompts/Webshop/llm_DK}

\input{prompts/Webshop/llm_DK_noncontrast}


\input{prompts/Webshop/knolwedge_with_primitives}

\subsection{Generated Code}
\subsubsection{ALFWorld}
Examples of generated code for each type of tasks under ALFWorld.
\input{prompts/ALFworld/skill_code}
\input{prompts/ALFworld/skill_composition}

\subsubsection{WebShop}
\label{app:code:webshop}

In the WebShop, we have observed that the GPT-4 model excels in skill composition, effectively generating responses based on instructions and context. In contrast, the GPT-3.5 model exhibits proficiency in task decomposition, breaking tasks into multiple sub-tasks; however, it struggles to seamlessly integrate these skill functions within the broader context. For instance, it cannot effectively pass the output of one function as input to subsequent functions.

\input{prompts/Webshop/skill_code}
\input{prompts/Webshop/skill_composition_GPT4}
\input{prompts/Webshop/skill_composition_GPT3_5}

%% file: tables_comp_human.tex
\begin{table*}[ht]
\centering
\caption{Comparison with \ourmodhuman in ALFWorld and WebShop.}
\vspace{-0.5em}
\label{tab:main-table}
\begin{subtable}{.62\linewidth}
\centering
\caption{Results in ALFWorld}
\label{comp_o3d_human_alfowrd}
\vspace{-0.5em}
\resizebox{\textwidth}{!}{
\begin{tabular}{c|c|cccccc|c}
\toprule
Model  & Method & Pick & Clean & Heat & Cool & Look & Pick2 & All \\
\midrule
\multirow{ 2}{*}{\shortstack{GPT-4\\(0613)}} & \ourmod   & \bf{92} & \bf{100} & \bf{96} & \bf{95} & \bf{100} & \bf{53} & \bf{91}\\
                            &\ourmodhuman & 83 & 100 & 87 & 95 & 100 & 53 & 88\\
\cmidrule{1-9}
\multirow{ 2}{*}{\shortstack{GPT-3.5\\(0613)}} & \ourmod       & \bf{71} & 35 & 4  & 67 & \bf{44} & \bf{24} & 41\\
                         & \ourmodhuman & 71 & \bf{68} & \bf{83} & \bf{71} & 44 & 24 & \bf{63}\\
\bottomrule
\end{tabular}
}
\end{subtable}
\hspace{1mm}
\vline
\begin{subtable}{.35\linewidth}
\centering
\caption{Results in WebShop}
\label{comp_o3d_human_webshop}
\vspace{-0.5em}
\resizebox{0.92\textwidth}{!}{
\begin{tabular}{c|c|c|c}
\toprule
Model  & Method & SR & Score \\
\midrule
\multirow{2}{*}{\shortstack{GPT-4\\(0613)}} & \ourmod& \bf{41} & 58\\
                            &\ourmodhuman & 41 & \bf{61}\\
\cmidrule{1-4}
\multirow{2}{*}{\shortstack{GPT-3.5\\(0613)}} & \ourmod & \bf{35} & \bf{61}\\
                            & \ourmodhuman           & 31      & 61 \\
\bottomrule
\end{tabular}
}
\end{subtable}

\end{table*}

%% file: a1_code.tex
This paper mainly focuses on \llmpolname which takes in textual observations from the environment and outputs textual actions to interact with the environment. But the proposed approach is also applicable to other scenarios of LLM-environment interaction. For example, it is well-known that LLMs can write code to solve tasks, where one asks LLMs to generate code and implement the policy function given proper prompts, i.e., 

\textit{$\bullet$ \codepolname.}
\setlength{\belowdisplayskip}{0pt} \setlength{\belowdisplayshortskip}{0pt}
\setlength{\abovedisplayskip}{0pt} \setlength{\abovedisplayshortskip}{0pt}
\begin{equation}
    \codepolicy(a|\tau):=Code(a|\tau)\leftarrow LLM(\prompara, \pretpara).
    \label{code-as-policy}
\end{equation}

In this case, the goal of learning is to optimize the produced function code $Code(a|\tau)$ without tuning the pre-trained weights $\theta$. The central idea of \ourmod naturally applies to \codepolname: the LLM discovers reusable skills (callable functions) and distills policy improvement tips from offline dataset; the obtained knowledge helps LLM refine its generated code policy iteratively; during execution stage, the base policy sequentially determines which skills/functions to call, and then the corresponding skill function is executed.

\input{algo_code}

\Cref{alg:main_code} illustrates the implementation of \codepolname. The major algorithm flow is similar to the text-based one in \Cref{alg:main}, with several distinctions below: \\
\textit{$\bullet$ Policy Initilization with Primitives.} \llmpolname directly provides the discovered primitives in the prompt of policy and advises the agent to follow the primitives, while \codepolname first lets the LLM write primitive functions and then calls these primitive functions in the code of skill-conditioned policies. \\
\textit{$\bullet$ Policy Improvement.} Since the distilled policy improvement tips are in natural language, we directly ask the LLM to merge the new suggestion into the prompt of \llmpolname. For \codepolname, we let the LLM consider the policy improvement tips and re-write the policy code.\\
\textit{$\bullet$ Policy Construction and Execution.} In Stage 3, we prompt the base policy to call the learned \llmpolname or \codepolname. Note that for \codepolname, it is possible that the generated skill-conditioned code has compilation errors, so that it requires human checking or validation on a small set of tasks to verify that the code is executable.

Using LLMs to directly interact with environments (\llmpolname) and using LLMs to write code to interact with environments (\codepolname) are usually discussed separately. Our study also reveals the different \textbf{pros and cons} of these two approaches that is not well-discussed in literature.\\
\textit{$\bullet$ Advantages of \codepolname. } \codepolname explicitly writes the acting policy in code, which is more interpretable and reliable, and can fully avoid hallucination or syntax errors in execution. Moreover, \codepolname is usually more cost efficient, as the generated code can be reused in new tasks without calling LLMs. Therefore, \codepolname can be more suitable for applications where reliability and efficiency are important.\\
\textit{$\bullet$ Advantages of \llmpolname. } \llmpolname is relatively easy to implement in practice with less human supervision. Also, in complicated environments such as WebShop where language understanding is important, \llmpolname can achieve much better performance than \codepolname, as it retains the commonsense, expressiveness and reasoning ability of pre-trained LLMs. Therefore, for language-oriented applications where reasoning and the ability of recovering from failure are crucial, \llmpolname, or a combination of the two approaches, can be a better choice.

\textit{$\bullet$ Advantages of \codepolname. }
\textit{$\bullet$ Reliability and robustness. } \codepolname understands the interaction logic and writes the acting policy in code, which is more interpretable and reliable, and can fully avoid hallucination or syntax errors in execution. \\
\textit{$\bullet$ Cost and efficiency. } 
\codepolname is usually more cost efficient, as the generated code can be reused in new tasks without calling LLMs. \\
\textit{$\bullet$ Flexibility and expressiveness. } In more complicated environments, \llmpolname may achieve much better performance than \codepolname, as it remains the commonsense and reasoning ability of pretrained LLMs, especially when language understanding is needed.\\
\textit{$\bullet$ Implementation difficulty and anticipated human supervision. } 
\llmpolname is relatively easy to implement in practice with less human supervision. For \codepolname, as discussed above, supervision or certain level of online validation may be required to ensure the code is executable. 

%% file: algo_code.tex
\begin{algorithm}[H]
\newcommand\mycommfont[1]{\footnotesize\ttfamily\textcolor{gray}{#1}}
\SetCommentSty{mycommfont}
\SetKwFunction{FInit}{InitSkillPolicy}
\SetKwFunction{FImp}{ImprovePolicy}
\SetKwFunction{FCst}{ConstructPolicy}
\SetKwProg{Fn}{Function}{:}{}
    \caption{Learning \codepolname with \ourmodfull (\ourmod)}
    \label{alg:main_code}
    \textbf{Input:} Pre-trained LLM, offline dataset $\data$, batch sizes $\bs_1, \bs_2$, max iteration steps $T_1, T_2$ 
    
    \textbf{Output:} \codepolname $\pi$ 
    
    \tcp{Stage 1: skill discovery}
    Initialize sets of skills $\skills=\emptyset$ 
    
    \For{$t=1,\dots,T_1$}{
        Sample $N_1$ trajectories $d \sim \data$ 
        
        Attempt to discover more skills $\skills \leftarrow LLM(\skills, d; \textcolor{blue}{\mathsf{DiscoverPrompt}})$
    }
    Segment trajectories in $\data$ based on skillset $\skills$, obtain $\data^{\skill_k}$ for each $\skill_k \in \skills$ 
            
    \tcp{Stage 2: knowledge distillation}
    \For{$\skill_k \in \skills$}{
        Initialize primitive set $\prims^{\skill_k}$ and the knowledge set $\tips^{\skill_k}$
        
        Initialize $\pi^{\skill_k} \leftarrow$ Let LLM write a function to reproduce $d \sim \data^{\skill}$ with primitive functions $\prims^{\skill}$
        
        \For{$t=1,\dots,T_2$}{
            Sample $N_2$ trajectories $d^{\skill_k} \sim \data^{\skill_k}$
            
            Attempt to discover more primitives $\prims^{\skill_k}\leftarrow LLM(\prims^{\skill_k}, d^{\skill_k}; \textcolor{blue}{\mathsf{DiscoverPrompt}})$ 
            
            Distill policy improvement tips 
            $\tips^{\skill_k} \leftarrow LLM(\tips^{\skill_k}, d^{\skill_k}; \textcolor{red}{\mathsf{DistillPrompt}})$
            
            Ask LLM to improve $\pi^{\skill}$ based on suggestions $\tips^{\skill}$
        }
    }
    \tcp{Stage 3: hierarchical policy execution}
    Sample examples $d \sim \data$ and segment them based on skills
    
    Provide the examples as demonstrations for $\llmpolicy$ to call skill-conditioned policies $\{\pi^\skill\}_{\skill\in\skills}$ given downstream tasks
\end{algorithm}

%% file: prompts/ALFworld/task_decompo_prompt_v0.tex
\begin{mdframed}[style=mystyle, frametitle={ALFWorld: Skill discovery prompt}]
\addcontentsline{toc}{paragraph}{ALFWorld: Skill discovery prompt}
\small\ttfamily
\begin{spverbatim}
You will be given the interaction histories between an agent and a household environment. Break the full interaction log into several sub-tasks, and summarize these sub-tasks in an abstract way. When there is already a list of environment-specific subtasks to start with, please keep them in your response or improve them if you figure out any mistakes, and you can also append new subtasks according to the given new trial. You should not add duplicate subtasks. Give your response after "New summarization: "

Here is an example:

Task 1:
Success trial:
You are in the middle of a room. Looking quickly around you, you see a cabinet 4, a cabinet 3, a cabinet 2, a cabinet 1, a coffeemachine 1, a countertop 1, a diningtable 3, a diningtable 2, a diningtable 1, a drawer 1, a fridge 1, a garbagecan 1, a microwave 1, a sidetable 1, a sinkbasin 1, a stoveburner 4, a stoveburner 3, a stoveburner 2, a stoveburner 1, and a toaster 1.
Your task is to: find some apple and put it in sidetable.
> go to fridge 1
The fridge 1 is closed.
> open fridge 1
You open the fridge 1. The fridge 1 is open. In it, you see a lettuce 2, a mug 2, a potato 2, and a tomato 1.
> close fridge 1
> go to garbagecan 1
On the garbagecan 1, you see a apple 3, and a egg 3.
> take apple 3 from garbagecan 1
You pick up the apple 3 from the garbagecan 1.
> go to sidetable 1
On the sidetable 1, you see a cup 1, a lettuce 1, a peppershaker 3, a potato 1, and a saltshaker 1.
> put apple 3 in/on sidetable 1
You put the apple 3 in/on the sidetable 1.

Here is no list of subtasks to start with.

New summarization: 

In the above interaction log, the agent completes three subtasks: find(apple) including "go to", "open" and "close" as primitive operations. take(apple) including "take" as primitive operations, and put(apple, sidetable) including "go to" and "put" as primitive operations. We thus can segment the entire log based on these three subtasks below.

You are in the middle of a room. Looking quickly around you, you see a cabinet 4, a cabinet 3, a cabinet 2, a cabinet 1, a coffeemachine 1, a countertop 1, a diningtable 3, a diningtable 2, a diningtable 1, a drawer 1, a fridge 1, a garbagecan 1, a microwave 1, a sidetable 1, a sinkbasin 1, a stoveburner 4, a stoveburner 3, a stoveburner 2, a stoveburner 1, and a toaster 1.
Your task is to: find some apple and put it in sidetable.
- find(apple):
    > go to fridge 1
    The fridge 1 is closed.
    > open fridge 1
    You open the fridge 1. The fridge 1 is open. In it, you see a lettuce 2, a mug 2, a potato 2, and a tomato 1.
    > close fridge 1
    You close fridge 1.
    > go to garbagecan 1
    On the garbagecan 1, you see a apple 3, and a egg 3.
- take(apple):
    > take apple 3 from garbagecan 1
    You pick up the apple 3 from the garbagecan 1.
- put(apple, sidetable):
    > go to sidetable 1
    On the sidetable 1, you see a cup 1, a lettuce 1, a peppershaker 3, a potato 1, and a saltshaker 1.
    > put apple 3 in/on sidetable 1
    You put the apple 3 in/on the sidetable 1.

As there is no a list of subtasks to start with, we list the Enviroment-specific subtasks below:

- find(object)
- take(object)
- put(object, receptable)
\end{spverbatim}
\end{mdframed}

%% file: prompts/Webshop/task_decomp_prompt_v0.tex
\begin{mdframed}[style=mystyle, frametitle={WebShop: Skill discovery prompt}]
\addcontentsline{toc}{paragraph}{WebShop: Skill discovery prompt}
\small\ttfamily
\begin{spverbatim}
You will be given the interaction histories between an agent and a webshop environment. Break the full interaction log into several sub-tasks, and summarize these sub-tasks in an abstrat way. When there is already a list of environment-specific subtasks to start with, please keep them in your response or improve them if you figure out any mistakes, and you can also append new subtasks according to the given new trial. You should not add duplicate subtasks. Give your response after "New summarization: "

Here is an example:

Task 1:
Success trial:
WebShop
Instruction:
i'm looking for a light pink long handle back loofah shower brush, and price lower than 40.00 dollars
[Search]

Action: search[light pink long handle back loofah shower brush]
Observation: 
[Back to Search] 
Page 1 (Total results: 50) 
[Next >] 
[B09N8YLMGJ] 
JulaJuyo Shower Back Scrubber, Long Handle Bath Body Brush Exfoliating Loofah Sponge on a Stick for Men Women, Colorful Nylon Mesh Showering Bathing Luffa Cleaning Lufa Washer Brush (1 Pack-Gray) 
\$7.99 
[B09J8KGN4P] 
JulaJuyo Back Scrubber for Shower, Long Handle Body Bath Brush Showering Loofah Sponge on a Stick for Men Women, Nylon Mesh Exfoliating Bathing Lufa Scrub Shower Cleaning Luffa Brush (1 Pack-White) 
\$7.99 
[B08Y6VZFT6] 
Prozklves Back Scrubber for Shower, Long Handle Back Loofah Shower Brush, Soft Nylon Mesh Back Cleaner Washer, Bath Brush for Women Men, Exfoliating Body Scrubber for Elderly (Light Pink) 
\$7.99 

Action: click[B08Y6VZFT6]
Observation: 
[Back to Search] 
[< Prev] 
scent [2 pack-beige][2 pack-green][2 pack-light pink][2 pcs=green+beige][2 pcs=green+pink][3 pack-beige][3 pack-light pink][3 pcs=colors][beige][green][light pink]
Prozklves Back Scrubber for Shower, Long Handle Back Loofah Shower Brush, Soft Nylon Mesh Back Cleaner Washer, Bath Brush for Women Men, Exfoliating Body Scrubber for Elderly (Light Pink) 
Price:\$7.99 
Rating: N.A. 
[Description] 
[Features] 
[Reviews] 
[Attributes] 
[Buy Now] 

Here is no list of subtasks to start with.

New summarization: 

In the above interaction log, the agent completes two subtasks: search_item() with the purpose of searching for the required item, including "search[QUERY]" as primitive operation, as well as select_item() with the purpose of selecting the item that matches the requirement, including "click[ITEM_TITLE]" as a primitive operation. We thus can segment the entire log based on these two subtasks below.

WebShop
Instruction:
i'm looking for a light pink long handle back loofah shower brush, and price lower than 40.00 dollars
[Search]

- search_item():
    Action: search[light pink long handle back loofah shower brush]
    Observation: 
    [Back to Search] 
    Page 1 (Total results: 50) 
    [Next >] 
    [B09N8YLMGJ] 
    JulaJuyo Shower Back Scrubber, Long Handle Bath Body Brush Exfoliating Loofah Sponge on a Stick for Men Women, Colorful Nylon Mesh Showering Bathing Luffa Cleaning Lufa Washer Brush (1 Pack-Gray) 
    \$7.99 
    [B09J8KGN4P] 
    JulaJuyo Back Scrubber for Shower, Long Handle Body Bath Brush Showering Loofah Sponge on a Stick for Men Women, Nylon Mesh Exfoliating Bathing Lufa Scrub Shower Cleaning Luffa Brush (1 Pack-White) 
    \$7.99 
    [B08Y6VZFT6] 
    Prozklves Back Scrubber for Shower, Long Handle Back Loofah Shower Brush, Soft Nylon Mesh Back Cleaner Washer, Bath Brush for Women Men, Exfoliating Body Scrubber for Elderly (Light Pink) 
    \$7.99 
- select_item():
    Action: click[B08Y6VZFT6]
    Observation: 
    [Back to Search] 
    [< Prev] 
    scent [2 pack-beige][2 pack-green][2 pack-light pink][2 pcs=green+beige][2 pcs=green+pink][3 pack-beige][3 pack-light pink][3 pcs=colors][beige][green][light pink]
    Prozklves Back Scrubber for Shower, Long Handle Back Loofah Shower Brush, Soft Nylon Mesh Back Cleaner Washer, Bath Brush for Women Men, Exfoliating Body Scrubber for Elderly (Light Pink) 
    Price: \$7.99 
    Rating: N.A. 
    [Description] 
    [Features] 
    [Reviews] 
    [Attributes] 
    [Buy Now] 

As there is no a list of subtasks to start with, we list the Enviroment-specific subtasks below:

- search_item()
- select_item()
\end{spverbatim}
\end{mdframed}

%% file: prompts/ALFworld/all_subtask_primitive_prompt.tex
\begin{mdframed}[style=mystyle, frametitle={ALFWorld: Skill-conditioned primitive discovery prompt}]
\addcontentsline{toc}{paragraph}{ALFWorld: Skill-conditioned primitive discovery prompt}
\small\ttfamily
\begin{spverbatim}
You will be given success histories in which you were placed an environment and given a task to complete. Please extract the valid environment-specific actions with the correct syntax from the trial, and list them under the corresponding subtask catergory. When there is already a list of environment-specific actions to start with, please keep them in your response or improve them if any syntax error detected, and you can also append new actions according to the given new trial. You should not add duplicate actions regardless of the examples. For existing actions in the list, do not add more examples. Give your response after "New summarization: "

Here is one example:

Task 1:
Success trial:
Plan for Completing the Task:
1. find(apple)
2. take(apple)

Segmented Interaction History:
> > find(apple): find some apple
> go to garbagecan 1
On the garbagecan 1, you see a apple 3, and a egg 3.
> > take(apple): take the apple
> take apple 3 from garbagecan 1
You pick up the apple 3 from the garbagecan 1.

New summarization: 

> > find(object)
Environment-specific actions:
    - go to [receptacle] [id]
    Example: go to garbagecan 1

> > take(object)
Environment-specific actions:
    - take [object] [id] from [receptacle] [id]
    Example: take apple 3 from garbagecan 1

> > put(object, receptacle)
Environment-specific actions:

> > cool(object)
Environment-specific actions:

> > use(object)
Environment-specific actions:

> > clean(object)
Environment-specific actions:

> > heat(object)
Environment-specific actions:
\end{spverbatim}
\end{mdframed}

%% file: prompts/Webshop/all_subtask_primitive_prompt_v1.tex
\begin{mdframed}[style=mystyle, frametitle={WebShop: Skill-conditioned primitive discovery prompt}]
\addcontentsline{toc}{paragraph}{WebShop: Skill-conditioned primitive discovery prompt}
\small\ttfamily
\begin{spverbatim}
You will be given success histories in which you were placed an environment and given a task to complete. Please extract the valid environment-specific actions with the correct syntax from the trial, and list them under the corresponding subtask catergory. When there is already a list of environment-specific actions to start with, please keep them in your response or improve them if any syntax error detected, and you can also append new actions according to the given new trial. You should not add duplicate actions regardless of the examples. For existing actions in the list, do not add more examples. Give your response after "New summarization: "

Here is one example:

Webshop 
Instruction:  
i would like a 3 ounce bottle of bright citrus deodorant for sensitive skin, and price lower than 50.00 dollars 
[Search]  

> > search_item(): search with a query regarding the instruction
Action: search[3 ounce bright citrus deodorant sensitive skin]
Observation: 
[Back to Search] 
Page 1 (Total results: 50) 
[Next >] 
[B078GWRC1J] 
Bright Citrus Deodorant by Earth Mama | Natural and Safe for Sensitive Skin, Pregnancy and Breastfeeding, Contains Organic Calendula 3-Ounce 
\$10.99 
[B078GTKVXY] 
Ginger Fresh Deodorant by Earth Mama | Natural and Safe for Sensitive Skin, Pregnancy and Breastfeeding, Contains Organic Calendula 3-Ounce 
\$10.99 
[B08KBVJ4XN] 
Barrel and Oak - Aluminum-Free Deodorant, Deodorant for Men, Essential Oil-Based Scent, 24-Hour Odor Protection, Cedar & Patchouli Blend, Gentle on Sensitive Skin (Mountain Sage, 2.7 oz, 2-Pack) 
\$15.95  

Action: click[Back to Search]
Observation:
Webshop 
Instruction:  
i would like a 3 ounce bottle of bright citrus deodorant for sensitive skin, and price lower than 50.00 dollars 
[Search]

New summarization: 

> > search_item()
Environment-specific actions:
    search[Query]
    # Example search[3 ounce bright citrus deodorant sensitive skin]
    click[Back to Search]
    # Example click[Back to Search]

> > select_item()
Environment-specific actions:

> > select_item_attributes()
Environment-specific actions:

> > purchase_item()
Environment-specific actions:
\end{spverbatim}
\end{mdframed}

%% file: prompts/ALFworld/subtask_knowledge_prompot_contrast_v1.tex
\begin{mdframed}[style=mystyle, frametitle={ALFWorld: Policy improvement tips contrastive distillation prompt}]
\addcontentsline{toc}{paragraph}{ALFWorld: Policy improvement tips contrastive distillation prompt}
\small\ttfamily
\begin{spverbatim}
You will be given both failure and success histories in which you were placed in an environment and given a task to complete. You should generate tips according to your task description rather than the historical interactions. By comparing the two histories, please find the mistakes in the failure trial, and summarize tips to fix the mistakes. The tips can also be the rules in this particular environment. You will need these tips later to solve a similar task. When there is already a list of tips to start with, please keep them in your response or improve them, and you can also append new tips according to the given new trials. If the new mistake you detect can be solved by any existing tips, you do not need to generate new tips. You do not have to add new tips if there is no one. Give your response after "New summarization: "

Here is one example:

Task 1:
Failure trial:
You are in the middle of a room. Looking quickly around you, you see a cabinet 10, a cabinet 9, a cabinet 8, a cabinet 7, a cabinet 6, a cabinet 5, a cabinet 4, a cabinet 3, a cabinet 2, a cabinet 1, a coffeemachine 1, a countertop 2, a countertop 1, a diningtable 2, a diningtable 1, a drawer 3, a drawer 2, a drawer 1, a fridge 1, a garbagecan 1, a microwave 1, a sinkbasin 1, a stoveburner 4, a stoveburner 3, a stoveburner 2, a stoveburner 1, and a toaster 1.
Your historical interactions:
> go to countertop 1
On the countertop 1, you see a apple 2, a butterknife 1, a cup 1, a knife 2, a knife 1, a pan 1, a potato 2, a spatula 2, a spatula 1, a spoon 2, a spoon 1, a tomato 2, a winebottle 3, and a winebottle 1.
> go to countertop 2
On the countertop 2, you see a butterknife 1, a cup 1, a fork 2, and a mug 1.
> take mug 1 from countertop 2
You pick up the mug 1 from the countertop 2.
Your task is to: heat the mug
> go to stove burner 1
On the stoveburner 1, you see a pan 2.
> heat mug 1 with stoveburner 1
Nothing happens.
> heat mug 1 with stoveburner 1
Nothing happens.

Success trial:
You are in the middle of a room. Looking quickly around you, you see a cabinet 10, a cabinet 9, a cabinet 8, a cabinet 7, a cabinet 6, a cabinet 5, a cabinet 4, a cabinet 3, a cabinet 2, a cabinet 1, a coffeemachine 1, a countertop 2, a countertop 1, a diningtable 2, a diningtable 1, a drawer 3, a drawer 2, a drawer 1, a fridge 1, a garbagecan 1, a microwave 1, a sinkbasin 1, a stoveburner 4, a stoveburner 3, a stoveburner 2, a stoveburner 1, and a toaster 1.
Your historical interactions:
> go to countertop 1
On the countertop 1, you see a apple 2, a butterknife 1, a cup 1, a knife 2, a knife 1, a pan 1, a potato 2, a spatula 2, a spatula 1, a spoon 2, a spoon 1, a tomato 2, a winebottle 3, and a winebottle 1.
> go to countertop 2
On the countertop 2, you see a butterknife 1, a cup 1, a fork 2, and a mug 1.
> take mug 1 from countertop 2
You pick up the mug 1 from the countertop 2.
Your task is to: heat the mug
> go to microwave 1
You arrive at loc 22. The microwave 1 is closed.
> heat mug 1 with microwave 1
You heat the mug 1 using the microwave 1.

New summarization: In task 1, my task is to heat the mug 1. I made a mistake in heating the mug with a stoveburner, and I also got stuck in a loop where I continually did this. As shown in the success trial, the correct way is to heat a mug with a microwave. Thus, I should try different actions when getting stuck in a loop; I should use a microwave to heat a mug, and I can directly heat an object with a microwave without opening the microwave and putting the object in it. 

Tips:
- Try to execute a different action when you get stuck in a loop of taking identical actions.
- Mug should be heated by a microwave rather than a stoveburner.
- You can directly heat an object with a microwave by taking the corresponding action.

Note, you should not directly include the above example tips in your response. Give your response after "New summarization: "
\end{spverbatim}
\end{mdframed}

%% file: prompts/ALFworld/subtask_knowledge_prompot_v1.tex
\begin{mdframed}[style=mystyle, frametitle={ALFWorld: Policy improvement tips non-contrastive distillation prompt}]
\addcontentsline{toc}{paragraph}{ALFWorld: Policy improvement tips non-contrastive distillation prompt}
\small\ttfamily
\begin{spverbatim}
You will be given success histories in which you were placed in an environment and given a task to complete. Please summarize tips according to your task description rather than the historical interactions.The tips can be the rules in this particular environment. You will need these tips later to solve a similar task. When there is already a list of tips to start with, please keep them in your response or improve them, and you can also append new tips according to the given new trials. You do not have to add new tips if there is no one. Give your response after "New summarization: "

Here is one example:

Task 1:
Success trial:
You are in the middle of a room. Looking quickly around you, you see a cabinet 10, a cabinet 9, a cabinet 8, a cabinet 7, a cabinet 6, a cabinet 5, a cabinet 4, a cabinet 3, a cabinet 2, a cabinet 1, a coffeemachine 1, a countertop 2, a countertop 1, a diningtable 2, a diningtable 1, a drawer 3, a drawer 2, a drawer 1, a fridge 1, a garbagecan 1, a microwave 1, a sinkbasin 1, a stoveburner 4, a stoveburner 3, a stoveburner 2, a stoveburner 1, and a toaster 1.
Your historical interactions:
> go to countertop 1
On the countertop 1, you see a apple 2, a butterknife 1, a cup 1, a knife 2, a knife 1, a pan 1, a potato 2, a spatula 2, a spatula 1, a spoon 2, a spoon 1, a tomato 2, a winebottle 3, and a winebottle 1.
> go to countertop 2
On the countertop 2, you see a butterknife 1, a cup 1, a fork 2, and a mug 1.
> take mug 1 from countertop 2
You pick up the mug 1 from the countertop 2.
Your task is to: heat the mug
> go to microwave 1
You arrive at loc 22. The microwave 1 is closed.
> heat mug 1 with microwave 1
You heat the mug 1 using the microwave 1.

New summarization: In task 1, my task is to heat the mug 1. As shown in the success trial, the correct way is to heat a mug with a microwave. Thus, I should use a microwave to heat a mug, and I can directly heat an object with a microwave without opening the microwave and putting the object in it. 

Tips:
- Mug should be heated by a microwave.
- You can directly heat an object with a microwave by taking the corresponding action.

Note, you should not directly include the above example tips in your response. Give your response after "New summarization: "
\end{spverbatim}
\end{mdframed}

%% file: prompts/Webshop/subtask_knowledge_prompt.tex
\begin{mdframed}[style=mystyle, frametitle={WebShop: Policy improvement tips contrastive distillation prompt}]
\addcontentsline{toc}{paragraph}{WebShop: Policy improvement tips contrastive distillation prompt}
\small\ttfamily
\begin{spverbatim}
You will be given both failure and success histories in which you were placed in an environment and given a task to complete. You should generate tips according to your task description rather than the historical interactions. By comparing the two histories, please find the mistakes in the failure trial, and summarize general tips to fix the mistakes. The tips can also be about the rules in this particular environment. You will need these tips later to solve a similar task. When there is already a list of tips to start with, please keep them in your response or improve them, and you can also append new tips according to the given new trials. If the new mistake you detect can be solved by any existing tips, you do not need to generate new tips. You do not have to add new tips if there is no one. Give your response after "New summarization: "

Here is one example:

Task1:
Failure Trial:
Webshop 
Instruction:  
i would like a 3 ounce bottle of bright citrus deodorant for sensitive skin, and price lower than 50.00 dollars 
[Search]
Your historical interactions:
Action: search[3 ounce bright citrus deodorant sensitive skin]
Observation: 
[Back to Search] 
Page 1 (Total results: 50) 
[Next >] 
[B078GWRC1J] 
Bright Citrus Deodorant by Earth Mama | Natural and Safe for Sensitive Skin, Pregnancy and Breastfeeding, Contains Organic Calendula 3-Ounce 
\$10.99 
[B078GTKVXY] 
Ginger Fresh Deodorant by Earth Mama | Natural and Safe for Sensitive Skin, Pregnancy and Breastfeeding, Contains Organic Calendula 3-Ounce 
\$10.99 
[B08KBVJ4XN] 
Barrel and Oak - Aluminum-Free Deodorant, Deodorant for Men, Essential Oil-Based Scent, 24-Hour Odor Protection, Cedar & Patchouli Blend, Gentle on Sensitive Skin (Mountain Sage, 2.7 oz, 2-Pack) 
\$15.95  
Action: click[B078GWRC1J]
Observation: 
[Back to Search] 
[< Prev] 
scent [assorted scents][bright citrus][calming lavender][ginger fresh][simply non-scents]
size [travel set (4-pack)][3 ounce (pack of 1)][3-ounce (2-pack)]
Bright Citrus Deodorant by Earth Mama | Natural and Safe for Sensitive Skin, Pregnancy and Breastfeeding, Contains Organic Calendula 3-Ounce 
Price: \$10.99 
Rating: N.A. 
[Description] 
[Features] 
[Reviews] 
[Buy Now]  
Your task is: select_item_attributes()
Action: click[BRIGHT CITRUS]
Observation: Invalid action!

Success Trial:
Webshop 
Instruction:  
i would like a 3 ounce bottle of bright citrus deodorant for sensitive skin, and price lower than 50.00 dollars 
[Search]  
Your historical interactions:
Action: search[3 ounce bright citrus deodorant sensitive skin]
Observation: 
[Back to Search] 
Page 1 (Total results: 50) 
[Next >] 
[B078GWRC1J] 
Bright Citrus Deodorant by Earth Mama | Natural and Safe for Sensitive Skin, Pregnancy and Breastfeeding, Contains Organic Calendula 3-Ounce 
\$10.99 
[B078GTKVXY] 
Ginger Fresh Deodorant by Earth Mama | Natural and Safe for Sensitive Skin, Pregnancy and Breastfeeding, Contains Organic Calendula 3-Ounce 
\$10.99 
[B08KBVJ4XN] 
Barrel and Oak - Aluminum-Free Deodorant, Deodorant for Men, Essential Oil-Based Scent, 24-Hour Odor Protection, Cedar & Patchouli Blend, Gentle on Sensitive Skin (Mountain Sage, 2.7 oz, 2-Pack) 
\$15.95  
Action: click[B078GWRC1J]
Observation: 
[Back to Search] 
[< Prev] 
scent [assorted scents][bright citrus][calming lavender][ginger fresh][simply non-scents]
size [travel set (4-pack)][3 ounce (pack of 1)][3-ounce (2-pack)]
Bright Citrus Deodorant by Earth Mama | Natural and Safe for Sensitive Skin, Pregnancy and Breastfeeding, Contains Organic Calendula 3-Ounce 
Price: \$10.99 
Rating: N.A. 
[Description] 
[Features] 
[Reviews] 
[Buy Now]  
Your task is: select_item_attributes()
Action: click[bright citrus]
Observation: You have clicked bright citrus. 
Action: click[3 ounce (pack of 1)]
Observation: You have clicked 3 ounce (pack of 1).

New summarization: In task 1, my task is select_item_attributes(), I made a mistake on capitalizing the name of a scent option [bright citrus], so that click[BRIGHT CITRUS] become invalid. As shown in the success trial, the correct action should be click[bright citrus]. Thus, I should strictly use the exact words of each option without making any changes.  

Tips:
- You should strictly use the exact words of each available option without making any changes.

Note, do not directly include the above example tips in your response. Give your response after "New summarization: "
\end{spverbatim}
\end{mdframed}

%% file: prompts/Webshop/subtask_knowledge_prompt_non_contrst.tex
\begin{mdframed}[style=mystyle, frametitle={WebShop: Policy improvement tips non-contrastive distillation prompt}]
\addcontentsline{toc}{paragraph}{WebShop: Policy improvement tips non-contrastive distillation prompt}
\small\ttfamily
\begin{spverbatim}
You will be given success histories in which you were placed in an environment and given a task to complete. Please summarize tips according to your task description rather than the historical interactions. The tips can be the rules in this particular environment. You will need these tips later to solve a similar task. When there is already a list of tips to start with, please keep them in your response or improve them, and you can also append new tips according to the given new trials. You do not have to add new tips if there is no one. Give your response after "New summarization: "

Here is one example:

Task1:
Success Trial:
Webshop 
Instruction:  
i would like a 3 ounce bottle of bright citrus deodorant for sensitive skin, and price lower than 50.00 dollars 
[Search]  
Your historical interactions:
Action: search[3 ounce bright citrus deodorant sensitive skin]
Observation: 
[Back to Search] 
Page 1 (Total results: 50) 
[Next >] 
[B078GWRC1J] 
Bright Citrus Deodorant by Earth Mama | Natural and Safe for Sensitive Skin, Pregnancy and Breastfeeding, Contains Organic Calendula 3-Ounce 
\$10.99 
[B078GTKVXY] 
Ginger Fresh Deodorant by Earth Mama | Natural and Safe for Sensitive Skin, Pregnancy and Breastfeeding, Contains Organic Calendula 3-Ounce 
\$10.99 
[B08KBVJ4XN] 
Barrel and Oak - Aluminum-Free Deodorant, Deodorant for Men, Essential Oil-Based Scent, 24-Hour Odor Protection, Cedar & Patchouli Blend, Gentle on Sensitive Skin (Mountain Sage, 2.7 oz, 2-Pack) 
\$15.95  
Action: click[B078GWRC1J]
Observation: 
[Back to Search] 
[< Prev] 
scent [assorted scents][bright citrus][calming lavender][ginger fresh][simply non-scents]
size [travel set (4-pack)][3 ounce (pack of 1)][3-ounce (2-pack)]
Bright Citrus Deodorant by Earth Mama | Natural and Safe for Sensitive Skin, Pregnancy and Breastfeeding, Contains Organic Calendula 3-Ounce 
Price: \$10.99 
Rating: N.A. 
[Description] 
[Features] 
[Reviews] 
[Buy Now]  
Your task is: select_item_attributes()
Action: click[bright citrus]
Observation: You have clicked bright citrus. 
Action: click[3 ounce (pack of 1)]
Observation: You have clicked 3 ounce (pack of 1). 

New summarization: In task 1, my task is select_item_attributes(). As shown in the success trial, I selected attributes according to the instruction. 

Tips:
- You should selected attributes according to instruction.

Note, do not directly include the tips in the above example in your response. Give your response after "New summarization: "
\end{spverbatim}
\end{mdframed}

%% file: prompts/ALFworld/hier_llm_llmdk_downstream.tex
\begin{mdframed}[style=mystyle, frametitle={ALFWorld: Base policy prompt}]
\addcontentsline{toc}{paragraph}{ALFWorld: Base policy prompt}
\small\ttfamily
\begin{spverbatim}
Interact with a household to complete a given task. 
You can call subprocedures find(object), take(object), use(object), clean(object), heat(object), cool(object) and put(object, location). Here are some examples.

You are in the middle of a room. Looking quickly around you, you see a cabinet 4, a cabinet 3, a cabinet 2, a cabinet 1, a countertop 1, a garbagecan 1, a handtowelholder 2, a handtowelholder 1, a sinkbasin 2, a sinkbasin 1, a toilet 1, a toiletpaperhanger 1, and a towelholder 1.
Your task is to: put some spraybottle on toilet.
> think: To solve the task, I need to find and take a sparybottle, then put it on toilet.
OK.
> call: find(spraybottle)
Subprocedure finished.
> call: take(spraybottle)
Subprocedure finished.
> call: put(spraybottle, toilet)
Subprocedure finished.

You are in the middle of a room. Looking quickly around you, you see a cabinet 4, a cabinet 3, a cabinet 2, a cabinet 1, a coffeemachine 1, a countertop 1, a diningtable 3, a diningtable 2, a diningtable 1, a drawer 1, a fridge 1, a garbagecan 1, a microwave 1, a sidetable 1, a sinkbasin 1, a stoveburner 4, a stoveburner 3, a stoveburner 2, a stoveburner 1, and a toaster 1.
Your task is to: find some apple and put it in sidetable.
> think: To solve the task, I need to find and take an apple, then put it in sidetable.
OK.
> call: find(apple)
Subprocedure finished.
> call: take(apple)
Subprocedure finished.
> call: put(apple, sidetable)
Subprocedure finished.

You are in the middle of a room. Looking quickly around you, you see a cabinet 13, a cabinet 12, a cabinet 11, a cabinet 10, a cabinet 9, a cabinet 8, a cabinet 7, a cabinet 6, a cabinet 5, a cabinet 4, a cabinet 3, a cabinet 2, a cabinet 1, a coffeemachine 1, a countertop 1, a diningtable 1, a drawer 1, a fridge 1, a garbagecan 1, a microwave 1, a shelf 3, a shelf 2, a shelf 1, a sinkbasin 1, a stoveburner 4, a stoveburner 3, a stoveburner 2, a stoveburner 1, and a toaster 1.
Your task is to: put a clean lettuce in diningtable.
> think: To solve the task, I need to find and take a lettuce, then clean it with sinkbasin, then put it in diningtable. 
OK.
> call: find(lettuce)
Subprocedure finished.
> call: take(lettuce)
Subprocedure finished.
> call: clean(lettuce)
Subprocedure finished.
> call: put(lettuce, diningtable)
Subprocedure finished.

You are in the middle of a room. Looking quickly around you, you see a cabinet 10, a cabinet 9, a cabinet 8, a cabinet 7, a cabinet 6, a cabinet 5, a cabinet 4, a cabinet 3, a cabinet 2, a cabinet 1, a coffeemachine 1, a countertop 3, a countertop 2, a countertop 1, a diningtable 1, a drawer 6, a drawer 5, a drawer 4, a drawer 3, a drawer 2, a drawer 1, a fridge 1, a garbagecan 1, a microwave 1, a sinkbasin 1, a stoveburner 4, a stoveburner 3, a stoveburner 2, a stoveburner 1, and a toaster 1.
Your task is to: heat some egg and put it in diningtable.
> think: To solve the task, I need to find and take an egg, then heat it with microwave, then put it in diningtable.
OK.
> call: find(egg)
Subprocedure finished.
> call: take(egg)
Subprocedure finished.
> call: heat(egg)
Subprocedure finished.
> call: put(egg, diningtable)
Subprocedure finished.

You are in the middle of a room. Looking quickly around you, you see a cabinet 13, a cabinet 12, a cabinet 11, a cabinet 10, a cabinet 9, a cabinet 8, a cabinet 7, a cabinet 6, a cabinet 5, a cabinet 4, a cabinet 3, a cabinet 2, a cabinet 1, a coffeemachine 1, a countertop 1, a diningtable 1, a drawer 1, a fridge 1, a garbagecan 1, a microwave 1, a shelf 3, a shelf 2, a shelf 1, a sinkbasin 1, a stoveburner 4, a stoveburner 3, a stoveburner 2, a stoveburner 1, and a toaster 1.
Your task is to: put a hot apple in fridge.
> think: To solve the task, I need to find and take an apple, then heat it with microwave, then put it in fridge.
OK.
> call: find(apple)
Subprocedure finished.
> call: take(apple)
Subprocedure finished.
> call: heat(apple)
Subprocedure finished.
> call: put(apple, fridge)
Subprocedure finished.

You are in the middle of a room. Looking quickly around you, you see a cabinet 16, a cabinet 15, a cabinet 14, a cabinet 13, a cabinet 12, a cabinet 11, a cabinet 10, a cabinet 9, a cabinet 8, a cabinet 7, a cabinet 6, a cabinet 5, a cabinet 4, a cabinet 3, a cabinet 2, a cabinet 1, a coffeemachine 1, a countertop 2, a countertop 1, a diningtable 1, a drawer 5, a drawer 4, a drawer 3, a drawer 2, a drawer 1, a fridge 1, a garbagecan 1, a microwave 1, a safe 1, a sinkbasin 1, a stoveburner 4, a stoveburner 3, a stoveburner 2, a stoveburner 1, and a toaster 1.
Your task is to: cool some pan and put it in stoveburner.
> think: To solve the task, I need to find and take a pan, then cool it with fridge, then put it in stoveburner.
OK.
> call: find(pan)
Subprocedure finished.
> call: take(pan)
Subprocedure finished.
> call: cool(pan)
Subprocedure finished.
> call: put(pan, stoveburner)
Subprocedure finished.

You are in the middle of a room. Looking quickly around you, you see a coffeetable 1, a diningtable 1, a drawer 4, a drawer 3, a drawer 2, a drawer 1, a dresser 1, a garbagecan 1, a sidetable 2, a sidetable 1, and a sofa 1.
Your task is to: put two cellphone in sofa.
> think: To solve the task, I need to find and take the first cellphone, then put it in sofa, then find and take the second cellphone, then put it in sofa.
OK.
> call: find(cellphone1)
Subprocedure finished.
> call: take(cellphone1)
Subprocedure finished.
> call: put(cellphone1, sofa)
Subprocedure finished.
> call: find(cellphone2)
Subprocedure finished.
> call: take(cellphone2)
Subprocedure finished.
> call: put(cellphone2, sofa)
Subprocedure finished.

You are in the middle of a room. Looking quickly around you, you see a bed 1, a desk 1, a drawer 3, a drawer 2, a drawer 1, a garbagecan 1, a safe 1, a shelf 5, a shelf 4, a shelf 3, a shelf 2, a shelf 1, a sidetable 2, and a sidetable 1.
Your task is to: look at bowl under the desklamp.
> think: To solve the task, I need to find and take a bowl, then find and use a desklamp. 
OK. 
> call: find(bowl)
Subprocedure finished.
> call: take(bowl)
Subprocedure finished.
> call: find(desklamp)
Subprocedure finished.
> call: use(desklamp)
Subprocedure finished.

You are in the middle of a room. Looking quickly around you, you see a bed 1, a drawer 10, a drawer 9, a drawer 8, a drawer 7, a drawer 6, a drawer 5, a drawer 4, a drawer 3, a drawer 2, a drawer 1, a dresser 1, a garbagecan 1, a shelf 9, a shelf 8, a shelf 7, a shelf 6, a shelf 5, a shelf 4, a shelf 3, a shelf 2, and a shelf 1.
Your task is to: examine the pen with the desklamp.
> think: To solve the task, I need to find and take a pen, then find and use a desklamp. First I need to find a pen. A pen is more likely to appear in drawer (1-10), shelf (1-9), bed (1), garbagecan (1). I can check one by one, starting with drawer 1.
OK.
> call: find(pen)
Subprocedure finished.
> call: take(pen)
Subprocedure finished.
> call: find(desklamp)
Subprocedure finished.
> call: use(desklamp)
Subprocedure finished.
\end{spverbatim}
\end{mdframed}

There are 7 skills in ALFWorld, which are find(object), take(object), use(object), clean(object), heat(object), cool(object) and put(object, location). Each of the skill induces a skill-conditioned policy. Here we first give a general prompt template for these policies.

\begin{mdframed}[style=mystyle, frametitle={ALFWorld: skill-conditioned policy prompt template}]
\addcontentsline{toc}{paragraph}{ALFWorld: subtask skill-conditioned policy prompt template}
\small\ttfamily
\begin{spverbatim}
Interact with a household to complete a given task. Please consider the following tips to solve new tasks:

```
{policy improvement tips}
{primitives}
```

Here are some examples.

```
{examples}
```
\end{spverbatim}
\end{mdframed}

In the above template, the policy improvement tips and primitives for each skill are provided in \Cref{app:knowledge}. For the examples, we use the same examples provided by ReAct. Note that ReAct uses different examples to prompt different types of ALFWorld tasks, while we reorganize the examples such that a single set of prompts is used across all types of tasks. This is because we discovered skills that are shared across tasks, allowing more generalizable knowledge.

Below we provide the full prompt of a skill-conditioned policy for skill "heat(object)". Prompts of other skills are available in our codebased, which will be released soon.

\begin{mdframed}[style=mystyle, frametitle={ALFWorld: skill-conditioned policy prompt for heat(object)}]
\addcontentsline{toc}{paragraph}{ALFWorld: subtask skill-conditioned policy prompt for heat(object)}
\small\ttfamily
\begin{spverbatim}
Interact with a household to complete a given task. Please consider the following rules to solve new tasks:
1. Try to execute a different action when you get stuck in a loop of taking identical actions.
2. Ensure you have the correct object in your possession before attempting to perform actions on it.
3. Check all possible locations if you can't find an object in the expected places.
4. You can directly heat an object with a microwave by taking the corresponding action.
5. If an action does not work, try a different approach or command.
6. Focus on relevant locations to complete the task efficiently.
7. A cup is not the same as a mug. Make sure to find and use the correct object for the task.
8. You can think about your plan in the first step, and after that you can only select from the following actions in your response (please strictly follow the syntax):
    go to [receptacle] [id]
    # Example: 
        go to countertop 1
    heat [object] [id] with [receptacle] [id]
    # Example: 
        heat egg 1 with microwave 1 

Here are some examples.

You are in the middle of a room. Looking quickly around you, you see a cabinet 10, a cabinet 9, a cabinet 8, a cabinet 7, a cabinet 6, a cabinet 5, a cabinet 4, a cabinet 3, a cabinet 2, a cabinet 1, a coffeemachine 1, a countertop 3, a countertop 2, a countertop 1, a diningtable 1, a drawer 6, a drawer 5, a drawer 4, a drawer 3, a drawer 2, a drawer 1, a fridge 1, a garbagecan 1, a microwave 1, a sinkbasin 1, a stoveburner 4, a stoveburner 3, a stoveburner 2, a stoveburner 1, and a toaster 1.
Your historical interactions: 
> open fridge 1
You open the fridge 1. The fridge 1 is open. In it, you see a lettuce 2, a mug 2, and a potato 3.
> go to countertop 1
On the countertop 1, you see a bread 1, a fork 1, and a saltshaker 1.
> go to countertop 2
On the countertop 2, you see nothing.
> go to countertop 3
On the countertop 3, you see a bowl 1, a butterknife 1, a egg 2, a kettle 2, a plate 1, a sink 1, and a spatula 2.
> take egg 2 from countertop 3
You pick up the egg 2 from the countertop 3. 
Your task is: heat(egg)
> think: My task is to heat an egg. Now I take an egg (2). Next, I need go to a microwave (1) and heat it.
OK.
> go to microwave 1
The microwave 1 is closed.
> heat egg 2 with microwave 1
You heat the egg 2 using the microwave 1.
> Done heat(egg): I have heated the desired egg (2).

You are in the middle of a room. Looking quickly around you, you see a cabinet 13, a cabinet 12, a cabinet 11, a cabinet 10, a cabinet 9, a cabinet 8, a cabinet 7, a cabinet 6, a cabinet 5, a cabinet 4, a cabinet 3, a cabinet 2, a cabinet 1, a coffeemachine 1, a countertop 1, a diningtable 1, a drawer 1, a fridge 1, a garbagecan 1, a microwave 1, a shelf 3, a shelf 2, a shelf 1, a sinkbasin 1, a stoveburner 4, a stoveburner 3, a stoveburner 2, a stoveburner 1, and a toaster 1.
Your historical interactions: 
> go to fridge 1
The fridge 1 is closed.
> open fridge 1
You open the fridge 1. The fridge 1 is open. In it, you see a cup 1, and a egg 1.
> go to diningtable 1
On the diningtable 1, you see a apple 1, a bread 1, a fork 3, a fork 2, a fork 1, a mug 2, a peppershaker 3, a plate 2, a pot 1, a soapbottle 3, a spatula 1, a spoon 3, a spoon 2, a spoon 1, a winebottle 3, and a winebottle 2.
> take apple 1 from diningtable 1
You pick up the apple 1 from the diningtable 1.
Your task is: heat(apple)
> think: My task is to heat an apple. Now I take an apple (1). Next, I need to go to a microwave (1) and heat it.
OK.
> go to microwave 1
The microwave 1 is closed.
> heat apple 1 with microwave 1
You heat the apple 1 using the microwave 1.
> Done heat(apple): I have heated the desired apple (1).
```
\end{spverbatim}
\end{mdframed}

%% file: prompts/Webshop/hier_llm_llmdk_downstream.tex
We note that in WebShop, whether to call a skill depends on the environment feedback. Therefore, the base policy needs to act based on environment observation at the end of each skill. For example, if the current item has no available attribute to select, then the base policy will not call the \textit{select\_item\_attributes()} skill. Otherwise it will first select attributes then purchase the item. Below is the prompt for the base policy.

\begin{mdframed}[style=mystyle, frametitle={WebShop: Base policy prompt}]
\addcontentsline{toc}{paragraph}{WebShop: Base policy prompt}
\small\ttfamily
\begin{spverbatim}
Interact with a webshop to solve a task. 
You can call subprocedures search_item(), select_item(), select_item_attributes(), purchase_item(). Here are some examples.

Webshop 
Instruction:  
i would like a 3 ounce bottle of bright citrus deodorant for sensitive skin, and price lower than 50.00 dollars 
[Search]  

Action: think[I need to first search for available items, then select the item that matches the instruction. If there are additional options for certain attributes, I also need to select them before purchase. Otherwise I can directly purchase the item.]
Observation: OK.

Action: search_item()
Observation: 
Subprocedure Done[I have found a list of items.]
Current Page:
[Back to Search] 
Page 1 (Total results: 50) 
[Next >] 
[B078GWRC1J] 
Bright Citrus Deodorant by Earth Mama | Natural and Safe for Sensitive Skin, Pregnancy and Breastfeeding, Contains Organic Calendula 3-Ounce 
\$10.99 
[B078GTKVXY] 
Ginger Fresh Deodorant by Earth Mama | Natural and Safe for Sensitive Skin, Pregnancy and Breastfeeding, Contains Organic Calendula 3-Ounce 
\$10.99 
[B08KBVJ4XN] 
Barrel and Oak - Aluminum-Free Deodorant, Deodorant for Men, Essential Oil-Based Scent, 24-Hour Odor Protection, Cedar & Patchouli Blend, Gentle on Sensitive Skin (Mountain Sage, 2.7 oz, 2-Pack) 
\$15.95  

Action: select_item()
Observation:
Subprocedure Done[I have selected the required item and it matches the instruction.]
Current Page:
[Back to Search] 
[< Prev] 
scent [assorted scents][bright citrus][calming lavender][ginger fresh][simply non-scents]
size [travel set (4-pack)][3 ounce (pack of 1)][3-ounce (2-pack)]
Bright Citrus Deodorant by Earth Mama | Natural and Safe for Sensitive Skin, Pregnancy and Breastfeeding, Contains Organic Calendula 3-Ounce 
Price: \$10.99 
Rating: N.A. 
[Description] 
[Features] 
[Reviews] 
[Buy Now]  

Action: select_item_attributes()
Observation:
Subprocedure Done[I have selected the additional attributes.]
Current Page:
You have clicked 3 ounce (pack of 1). 

Action: purchase_item()

WebShop 
Instruction:  
i need some living room furniture, and price lower than 100.00 dollars 

Action: think[I need to first search for available items, then select the item that matches the instruction. If there are additional options for certain attributes, I also need to select them before purchase. Otherwise I can directly purchase the item.]
Observation: OK.

Action: search_item()
Observation:
Subprocedure Done[I have found a list of items.]
Current Page:
Observation: 
[Back to Search] 
Page 1 (Total results: 50) 
[Next >] 
[B09G71YGSV] 
Patterned Square Pouf in Black and Gray Boho Decor Bean Bag Bedroom Furniture Storage Ottoman Accent Chair Bean Bag Chair Papasan Chair Living Room Rugs Floor Pillow Bedroom Chair Foot Stool 
\$100.0 
[B00UNYEOVE] 
TOV Furniture The Gavin Collection Modern Tufted Linen Fabric Upholstered Living Room Sofa, Beige, 85.25" x 34.5" x 27.75" 
\$100.0 
[B07194KS6G] 
Hekman Furniture Round Coffee Table 
\$100.0 

Action: select_item()
Observation:
Subprocedure Done[I have selected the required item and it matches the instruction.]
Current Page:
[Back to Search] 
[< Prev] 
TOV Furniture The Gavin Collection Modern Tufted Linen Fabric Upholstered Living Room Sofa, Beige, 85.25" x 34.5" x 27.75" 
Price: \$100.0 
Rating: N.A. 
[Description] 
[Features] 
[Reviews] 
[Attributes] 
[Buy Now] 

Action: purchase_item()
\end{spverbatim}
\end{mdframed}

Below is a template of the prompt for skill-conditioned policies. Then we provide an example with skill \textit{select\_item\_attributes()}.

\begin{mdframed}[style=mystyle, frametitle={WebShop: skill-conditioned policy prompt template}]
\addcontentsline{toc}{paragraph}{WebShop: skill-conditioned policy prompt template}
\small\ttfamily
\begin{spverbatim}
Interact with a webshop to solve a task. Please consider the following tips to solve new tasks:

```
{policy improvement tips}
{primitives}
```

Here are some examples.

```
{examples}
```
\end{spverbatim}
\end{mdframed}

\begin{mdframed}[style=mystyle, frametitle={WebShop: skill-conditioned policy prompt for \textit{select\_item\_attributes()}}]
\addcontentsline{toc}{paragraph}{WebShop: skill-conditioned policy prompt for \textit{select\_item\_attributes()}}
\small\ttfamily
\begin{spverbatim}
Interact with a webshop to solve a task. Please consider the following tips to solve new tasks:
1. You should strictly use the exact words of each clickable option without making any changes.
2. Ensure to select all the necessary attributes as per the instruction.
3. Pay attention to the product details to ensure they match the instruction.
4. Be aware that color and size options may not always be straightforward and may include additional information or codes.
5. Be careful with the case and spacing of the words when selecting item attributes.
6. Make sure to select a product that has all the required attributes as per the instruction.
7. Use quotation marks around specific phrases when searching for items to get more accurate results.
8. Please only generate following actions in your response:
    click[Attribute]
    # Example click[38 m eu]
    click[Color]
    # Example click[brown]
    click[Style]
    # Example click[powerlite 1785w]
    click[Size]
    # Example click[8.5]
    click[Flavor]
    # Example click[plantain chips]
    click[Features]
    # Example click[Features]
    click[Digital Storage Capacity]
    # Example click[32 gb]
    click[Offer Type]
    # Example click[lockscreen ad-supported]
    click[Description]
    # Example click[Description]

Here are some examples.

Webshop 
Instruction:  
i would like a 3 ounce bottle of bright citrus deodorant for sensitive skin, and price lower than 50.00 dollars 
[Search]  

Your historical interactions: 
Action: search[3 ounce bright citrus deodorant sensitive skin]
Observation: 
[Back to Search] 
Page 1 (Total results: 50) 
[Next >] 
[B078GWRC1J] 
Bright Citrus Deodorant by Earth Mama | Natural and Safe for Sensitive Skin, Pregnancy and Breastfeeding, Contains Organic Calendula 3-Ounce 
$10.99 
[B078GTKVXY] 
Ginger Fresh Deodorant by Earth Mama | Natural and Safe for Sensitive Skin, Pregnancy and Breastfeeding, Contains Organic Calendula 3-Ounce 
$10.99 
[B08KBVJ4XN] 
Barrel and Oak - Aluminum-Free Deodorant, Deodorant for Men, Essential Oil-Based Scent, 24-Hour Odor Protection, Cedar & Patchouli Blend, Gentle on Sensitive Skin (Mountain Sage, 2.7 oz, 2-Pack) 
$15.95  
Action: click[B078GWRC1J]
Observation: 
[Back to Search] 
[< Prev] 
scent [assorted scents][bright citrus][calming lavender][ginger fresh][simply non-scents]
size [travel set (4-pack)][3 ounce (pack of 1)][3-ounce (2-pack)]
Bright Citrus Deodorant by Earth Mama | Natural and Safe for Sensitive Skin, Pregnancy and Breastfeeding, Contains Organic Calendula 3-Ounce 
Price: $10.99 
Rating: N.A. 
[Description] 
[Features] 
[Reviews] 
[Buy Now]  

Your task is: select_attributes()

Action: think[For 3 ounce bottle of bright citrus deodorant for sensitive skin, the item has options 'bright citrus' and '3 ounce (pack of 1)' and seems good to buy.]
Observation: OK.

Action: click[bright citrus]
Observation: You have clicked bright citrus. 

Action: click[3 ounce (pack of 1)]
Observation: You have clicked 3 ounce (pack of 1). 

Action: Done[I have selected the additional attributes.]
\end{spverbatim}
\end{mdframed}

%% file: prompts/ALFworld/skill_discovery_prompt.tex
\begin{mdframed}[style=mystyle, frametitle={ALFWorld: Skill discovery}]
\addcontentsline{toc}{paragraph}{ALFWorld: Skill discovery}
\small\ttfamily
\begin{spverbatim}
You are a code generator. 
You based on the history of agents interacting with the environment to generate a general code solution to solve the task.

************************************************************************
You have access to the following primitive functions:
# example: go to table 1 => go('table 1')
def go(location: str):
    return 'go to ' + location

# example: open drawer 1 => open('drawer 1')
def open(object: str):
    return 'open ' + object
    
# example: The cabinet 5 is closed => isclose('The cabinet 5 is closed') return True
def isclose(observation: str):
    if "is closed" in observation:
        return True
    else:
        return False

# example: input_str = "cabinet 3, coffeemachine 1,cabinet 5,", object_name ="coffeemachine", return "coffeemachine 1"
def extract_pattern(input_str, object_name):
    # Using RegEx pattern to find the object_name followed by a space and a number
    pattern = r'{} \d+'.format(object_name)
    match = re.search(pattern, input_str)

    # If a match is found
    if match:
        return match.group()  # This returns the first match found
    else:
        return "No match found."

def interact_with_environment(action):
    env_last_obs, _, _, info = env.step([action])
    print(action)
    print(env_last_obs)
    return env_last_obs[0],info['won'][0]

def is_in(object,env_last_obs):
    # check whether an object is in the envrionment observation
    if object in env_last_obs:
        return True
    return False

# example: use desklamp 1 => use('desklamp 1')
def use(desklamp: str):
    return 'use ' + desklamp

# example: close drawer 1 => close('drawer 1')
def close(object: str):
    return 'close ' + object

# example: put spraybottle 1 in/on toilet 1 => put('spraybottle 1', 'toilet 1')
def put(object: str, location: str):
    return 'put ' + object + ' in/on ' + location

# example: clean cloth 1 with sinkbasin 1  => clean('cloth 1','sinkbasin 1')
def clean(object_1: str, sinkbasin: str):
    return 'clean ' + object_1 + ' with ' + sinkbasin

# example: take keychain 2 from sofa 1 => take('keychain 2', 'sofa 1')
def take(object: str, location: str):
    return 'take ' + object + ' from ' + location

# example: heat tomato 2 with microwave 1
def heat(object: str, microwave: str):
    return 'heat ' + object + ' with ' + microwave

# example: cool pot 1 with fridge 1
def cool(object: str, fridge: str):
    return 'cool ' + object + ' with ' + fridge

************************************************************************
Tips:
1. The basic interaction can be in the following format:
action = go(location)
env_observation = interact_with_environment(action)

2. You can extract specific object names by: 
specific_object = extract_pattern(env_observation, object_to_find)

I will start showing you the interaction history of task find. 
You can summarize new tips based on the new interaction and generate the function based on the tips and primitives.
The function should be in the format of

def clean(object_to_heat):
    # You can assume the object is already with you.
    # Your implementation here
\end{spverbatim}
\end{mdframed}

%% file: prompts/Webshop/skill_discovery_prompt.tex
\begin{mdframed}[style=mystyle, frametitle={WebShop: Skill discovery}]
\addcontentsline{toc}{paragraph}{WebShop: Skill discovery}
\small\ttfamily
\begin{spverbatim}
You are a code generator. 
You based on the history of agent interacting with the environment to generate a general code solution to solve the task.

************************************************************************
You have access to the following primitive functions:

# Example select(B0054KM7IY)
def select(item_number):
    return f'click[{item_number}]'

# Example buy()
def buy():
    return f'click[Buy Now]'

# Example:
# click(brown)
# click(Flavor)
# click(rectangular)
# click(Description)
# click(32 gb)
def click(feature):
    return f'click[{feature}]'

def interact_with_environment(action):  
    global env_id
    env_obs,reward,done = env.step(env_id,action)
    print(action)
    print(env_obs)
    return env_obs, reward

def search(instruction):
    #Usage: search item based on the instruction
    #Example #1:
    #Instruction:  im looking for a light pink long handle back loofah shower brush, and price lower than 40.00 dollars 
    #Return: search[light pink long handle back loofah shower brush]

def extract_requirement(instruction):   
    #Usage: based on instruction, extract requirements
    #Instruction:  im looking for a light pink long handle back loofah shower brush, and price lower than 40.00 dollars 
    #Return: a list of requirement ['easy apply','pine tar scented','price lower than 50.00 dollars']

def requirement_match_check(item_discription,requirement):
    #Usage: Check whether an item discription (not item attributes) satisfy certain requirement
    #item_discription = 'JulaJuyo Back Scrubber for Shower, Long Handle Body Bath Brush Showering Loofah Sponge on a Stick for Men Women, Nylon Mesh Exfoliating Bathing Lufa Scrub Shower Cleaning Luffa Brush (1 Pack-White) \$7.99 '
    #requirement = 'light pink'
    #return  Ture/False

def parse_items_observation_into_items_dict(items_observation):
    #Usage: parse a string of item into a dictionary.
    #items_observation="[B09GLCXKP6] ApexDesk MK Series Children's Height Adjustable Chair with Study Desk w/Integrated Shelf & Drawer (Pink Desk & MK Chair Bundle) \$419.99 
    #[B0184JRW1I] ApexDesk Little Soleil DX 43" Children's Height Adjustable Study Desk w/ Integrated Shelf & Drawer (Desk+Chair Bundle – Blue) \$359.99 "
    # return {'B09GLCXKP6':'ApexDesk MK...','B0184JRW1I': 'ApexDesk Little...'}

def select_most_relevant_attibutes(item_description,requirements):
    #Usage: return a list of most relevant attributes you can directly click based on item_description
    #item_description: string
    #requirements: [requirements#1,requirements#2,requirements#3...]
    return [attribute#1,attributes#2..]

************************************************************************
Tips:
1. The basic interaction can be in the following format:
env_observation,reward = interact_with_environment(action)
2. You should focus on to generate the code for the interaction after "Your task is:select_item()", i.e why the Action is executed (after "Your task is:select_item()")? 
3. You need to based on the interaction history before "Your task is:select_item()" to infer why the Action is executed
4. It is a web shopping interface, you are browsing web pages. 
5. You are allowed to use numpy library: import numpy as np
6. You code should be able to generate action based on the observation and interact with env with interact_with_environment(action)

I will start show you the interaction history of task find. 
You can summarize new tips based on the new interaction and generate the function based on the tips and primitives.
The function should be in the format of

def select_item(items,instruction):
    # items: dict {{'item number': discription}}
    # instruction: str
    
    # Hints: 
    1. You may maintain a dictionary where key is the item number and the values containing number of satified requirements
    2. Never select or click [<Prev] and [Next>]. Do not evenwrite [<Prev] and [Next>] in your code.

    # return the best match item number among all items. no need to match all requirements in the instruction
\end{spverbatim}
\end{mdframed}

We encapsulate the LLM within the specified functions, exposing only the function signature and usage in the aforementioned prompt to the LLM.

\begin{mdframed}[style=mystyle, frametitle={Webshop: Skill discovery}]
\addcontentsline{toc}{paragraph}{Webshop: Skill discovery}
\small\ttfamily
\begin{spverbatim}

def extract_requirement(instruction):
    user_content = f"""
    Given an instruction, you need decompose it and find the requirments:
    For example:
    
    Example #1:
    Instruction:  
    i'm looking for a light pink long handle back loofah shower brush, and price lower than 40.00 dollars 
    
    You return:
    requirement_1 = 'light pink'
    requirement_2 = 'long handle back loofah'
    requirement_3 =  'price lower than 40.00 dollars'
    
    Example #2:
    Instruction:  
    i need easy apply pine tar scented mustache wax stick for men, and price lower than 50.00 dollars
    
    You return:
    {{'requirement_1' : 'easy apply'
    'requirement_2': 'pine tar scented'
    'requirement_3':'price lower than 50.00 dollars'}}
    
    Now given an new instruction:
    {instruction}
    You return: 
    """
    res, _ = query_openai_api(user_content=user_content)
    return res 

def requirement_match_check(item,requirement):
    requirement_embedding = get_embedding(requirement)
    item_embedding = get_embedding(item)
    score = cosine_similarity(requirement_embedding,item_embedding)
    print("similarity score",score)
    if score>0.8:
        return True
    else:
        return False 

def select_most_relevant_attibutes(item_description,requirement):
    user_content = """
    For example:
    
    Example #1:
    Given item_description:
    scent [citrus & spice][essential 7][extra firm hold][lime & sage][mountain fresh][original blend][pine tar][unscented]
    Mountaineer Brand Stache Stick - All-Natural Convenient 1.5 Oz Mustache Wax Stick for Men (Extra Firm Hold) 
    Price: \$16.99 
    Rating: N.A. 
    [Description] 
    [Features] 
    [Reviews] 
    [Attributes] 
    [Buy Now] 
    
    And requirements:
    requirement_1 = 'light pink'
    requirement_2 = 'long handle back loofah'
    requirement_3 =  'price lower than 40.00 dollars'

    You return:
    [[pine tar]]
    
    Example #2:
    Given item_description:
    scent [2 pack-beige][2 pack-green][2 pack-light pink][2 pcs=green+beige][2 pcs=green+pink][3 pack-beige][3 pack-light pink][3 pcs=colors][beige][green][light pink]
    Prozklves Back Scrubber for Shower, Long Handle Back Loofah Shower Brush, Soft Nylon Mesh Back Cleaner Washer, Bath Brush for Women Men, Exfoliating Body Scrubber for Elderly (Light Pink) 
    Price: \$7.99 
    Rating: N.A. 
    [Description] 
    [Features] 
    [Reviews] 
    [Attributes] 
    [Buy Now] 
    
    And requirements:
    requirement_1 = 'light pink'
    requirement_2 = 'long handle back loofah'
    requirement_3 =  'price lower than 40.00 dollars'
    
    You return:
    [[light pink]]
    """
    \end{spverbatim}
\end{mdframed}

%% file: prompts/ALFworld/hier_llm_hdk.tex
\begin{mdframed}[style=mystyle, frametitle={ALFWorld: Policy improvement tips and primitives for skill clean(object)}]
\addcontentsline{toc}{paragraph}{ALFWorld: Policy improvement tips and primitives for skill clean(object)}
\small\ttfamily
\begin{spverbatim}
1. Do not try to turn on the sinkbasin; it does not work.
2. You can directly clean an object with a sinkbasin by taking the corresponding action.
3. Avoid going to irrelevant places when you are not sure about the next action; it wastes time and does not help in accomplishing the task.
4. Focus on the task at hand and avoid unnecessary actions. 
5. Check the contents of a location before moving to another one.
6. Attempt to perform the task as soon as you have the necessary item and are in the correct location.
7. You can think about your plan in the first step, and after that you can only select from the following actions in your response (please strictly follow the syntax):
    go to [receptacle] [id]
    # Example: 
        go to countertop 1
    clean [object] [id] with [receptacle] [id]
    # Example: 
        clean apple 1 with sinkbasin 1
\end{spverbatim}
\end{mdframed}

\begin{mdframed}[style=mystyle, frametitle={ALFWorld: Policy improvement tips and primitives for skill cool(object)}]
\addcontentsline{toc}{paragraph}{ALFWorld: Policy improvement tips and primitives for skill cool(object)}
\small\ttfamily
\begin{spverbatim}
1. Please complete the task with as few steps as possible.
2. You can think about your plan in the first step, and after that you can only select from the following actions in your response (please strictly follow the syntax):
    go to [receptacle] [id]
    # Example: 
        go to countertop 1
    cool [object] [id] with [receptacle] [id]
    # Example: 
        cool egg 1 with fridge 1
3. You can take action cool [object] [id] with [receptacle] [id] even when the receptacle is closed.
\end{spverbatim}
\end{mdframed}

\begin{mdframed}[style=mystyle, frametitle={ALFWorld: Policy improvement tips and primitives for skill find(object)}]
\addcontentsline{toc}{paragraph}{ALFWorld: Policy improvement tips and primitives for skill find(object)}
\small\ttfamily
\begin{spverbatim}
1. Try to execute a different action when you stuck in a loop of taking identical actions.
2. Try to search all other locations if the object could not be found in initial likely locations.
3. Remember the location of each object you have seen in order to quickly find it later. 
4. Remember the number of objects you have seen in each location, so you can directly go back the location to find another one.
5. Do not mismatch target obejct with similar object, for example, a mug is different with a cup. 
6. Please complete the task with as few steps as possible.
7. You can think about your plan in the first step, and after that you can only select from the following actions in your response (please strictly follow the syntax):
    go to [receptacle] [id]
    # Example: 
        go to countertop 1
    open [receptacle] [id]
    # Example: 
        open drawer 2 
\end{spverbatim}
\end{mdframed}

\begin{mdframed}[style=mystyle, frametitle={ALFWorld: Policy improvement tips and primitives for skill heat(object)}]
\addcontentsline{toc}{paragraph}{ALFWorld: Policy improvement tips and primitives for skill heat(object)}
\small\ttfamily
\begin{spverbatim}
1. You need to first go to a microwave to heat an object.
2. You can heat any object with microwave.
3. Do not open the microwave, directly take the heat operation.
5. Only select from the following actions in your response (please strictly follow the syntax):
go to [receptacle] [id]
# Example: 
    go to countertop 1
heat [object] [id] with [receptacle] [id]
# Example: 
    heat egg 1 with microwave 1  
\end{spverbatim}
\end{mdframed}

\begin{mdframed}[style=mystyle, frametitle={ALFWorld: Policy improvement tips and primitives for skill put(object, receptacle)}]
\addcontentsline{toc}{paragraph}{ALFWorld: Policy improvement tips and primitives for skill put(object, receptacle)}
\small\ttfamily
\begin{spverbatim}
1. Try to execute a different action when you stuck in a loop of taking identical actions.
2. You can directly put an object in/on an occupied receptacle.
3. Please complete the task with as few steps as possible.
4. You can think about your plan in the first step, and after that you can only select from the following actions in your response (please strictly follow the syntax):
    go to [receptacle] [id]
    # Example: 
        go to countertop 1
    open [receptacle] [id]
    # Example: 
        open drawer 2 
    put [object] [id] in/on [receptacle] [id]
    # Example: 
        put lettuce 1 in/on countertop 1
\end{spverbatim}
\end{mdframed}

\begin{mdframed}[style=mystyle, frametitle={ALFWorld: Policy improvement tips and primitives for skill take(object)}]
\addcontentsline{toc}{paragraph}{ALFWorld: Policy improvement tips and primitives for skill take(object)}
\small\ttfamily
\begin{spverbatim}
1. Try to execute a different action when you stuck in a loop of taking identical actions.
2. You can think about your plan in the first step, and after that you can only select from the following actions in your response (please strictly follow the syntax):
    take [object] [id] from [receptacle] [id]
    # Example: 
        take lettuce 1 from countertop 1    
\end{spverbatim}
\end{mdframed}

\begin{mdframed}[style=mystyle, frametitle={ALFWorld: Policy improvement tips and primitives for skill use(object)}]
\addcontentsline{toc}{paragraph}{ALFWorld: Policy improvement tips and primitives for skill use(object)}
\small\ttfamily
\begin{spverbatim}
1. Try to execute a different action when you stuck in a loop of taking identical actions.
2. Please complete the task with as few steps as possible.
3. You can think about your plan in the first step, and after that you can only select from the following actions in your response (please strictly follow the syntax):
    use [receptacle] [id]
    # Example: 
        use desklamp 1
\end{spverbatim}
\end{mdframed}

%% file: prompts/Webshop/hier_llm_hdk.tex
\begin{mdframed}[style=mystyle, frametitle={WebShop: Policy improvement tips and primitives for skill search\_item()}]
\addcontentsline{toc}{paragraph}{WebShop: Policy improvement tips and primitives for skill search\_item()}
\small\ttfamily
\begin{spverbatim}
1. Please pay attention to instruction description to find the product that exactly matches with the requirement in the instruction;   
2. You should pay attention to the current observation to check clickable buttons;
3. When the current selected product does not match the requirement in the instruction well, you should navigate to the results page to check other likely product candidates; 
4. When there is no product that matches the requirement in the instruction well, you should search again with a new query; 
5. You should click[Back to Search] before re-searching with a new query;
6. Please only generate following actions in your response:
    search[Query]
    # Example search[3 ounce bright citrus deodorant sensitive skin]
    click[Back to Search]
    # Example click[Back to Search]
\end{spverbatim}
\end{mdframed}

\begin{mdframed}[style=mystyle, frametitle={WebShop: Policy improvement tips and primitives for skill select\_item()}]
\addcontentsline{toc}{paragraph}{WebShop: Policy improvement tips and primitives for skill select\_item()}
\small\ttfamily
\begin{spverbatim}
1. Please pay attention to instruction description to find the product that exactly matches with the requirement in the instruction;   
2. You can click[Description] or click[Features] to find the detailed description of a product; 
3. You should pay attention to the current observation to check clickable buttons;
4. When the current selected product does not match the requirement in the instruction well, you should navigate to the results page to check other likely product candidates; 
5. Please only generate following actions in your response:
    click[Item ID]
    # Example click[B0054KM7IY]
    click[< Prev]
    # Example click[< Prev]
\end{spverbatim}
\end{mdframed}

\begin{mdframed}[style=mystyle, frametitle={WebShop: Policy improvement tips and primitives for skill select\_item\_attributes()}]
\addcontentsline{toc}{paragraph}{WebShop: Policy improvement tips and primitives for skill select\_item\_attributes()}
\small\ttfamily
\begin{spverbatim}
1. Please pay attention to instruction description to find the product that exactly matches with the requirement in the instruction;   
2. You can click[Description] to find the detailed description of a product; 
3. Remember to select all required attributes using click[Attributes] before click[Buy Now]; 
4. You should click[Description] or click[Features] to double-check attributes when there is no clickable attributes; 
5. You should pay attention to the current observation to check clickable buttons;
6. Please only generate following actions in your response:
    click[Attributes]
    # Example 1: click[16.9 fl oz (pack of 1)] 
    # Example 2: click[blue]
    click[< Prev]
    # Example: click[< Prev]
    click[Description]
    # Example: click[Description]
    click[Features]
    # Example: click[Features]
\end{spverbatim}
\end{mdframed}

\begin{mdframed}[style=mystyle, frametitle={WebShop: Policy improvement tips and primitives for skill purchase\_item()}]
\addcontentsline{toc}{paragraph}{WebShop: Policy improvement tips and primitives for skill purchase\_item()}
\small\ttfamily
\begin{spverbatim}
1. Please only generate following actions in your response:
    click[Buy Now]
    # Example click[Buy Now]
\end{spverbatim}
\end{mdframed}

%% file: prompts/ALFworld/hier_llm_llmdk.tex
\begin{mdframed}[style=mystyle, frametitle={ALFWorld: Policy improvement tips and primitives for skill clean(object)}]
\addcontentsline{toc}{paragraph}{ALFWorld: Policy improvement tips and primitives for skill clean(object)}
\small\ttfamily
\begin{spverbatim}
1. Do not try to turn on the sinkbasin; it does not work.
2. You can directly clean an object with a sinkbasin by taking the corresponding action.
3. Avoid going to irrelevant places when you are not sure about the next action; it wastes time and does not help in accomplishing the task.
4. Focus on the task at hand and avoid unnecessary actions. 
5. Check the contents of a location before moving to another one.
6. Attempt to perform the task as soon as you have the necessary item and are in the correct location.
7. You can think about your plan in the first step, and after that you can only select from the following actions in your response (please strictly follow the syntax):
    go to [receptacle] [id]
    # Example: 
        go to countertop 1
    clean [object] [id] with [receptacle] [id]
    # Example: 
        clean apple 1 with sinkbasin 1
\end{spverbatim}
\end{mdframed}

\begin{mdframed}[style=mystyle, frametitle={ALFWorld: Policy improvement tips and primitives for skill cool(object)}]
\addcontentsline{toc}{paragraph}{ALFWorld: Policy improvement tips and primitives for skill cool(object)}
\small\ttfamily
\begin{spverbatim}
1. Always remember to locate and pick up the object first before trying to perform an action on it.
2. An object can be cooled using a fridge.
3. You can directly cool an object with a fridge by taking the corresponding action.
4. Avoid getting stuck in a loop of actions that do not contribute to the completion of the task.
5. Explore the environment thoroughly to locate the object needed to complete the task.
6. If an object is not in its usual place, check other possible locations.
7. If you can't find an object in one location, it might be in another. In this case, the lettuce was on the countertop, not the fridge.
8. Avoid attempting actions that are not possible in the given environment.
9. You can think about your plan in the first step, and after that you can only select from the following actions in your response (please strictly follow the syntax):
    go to [receptacle] [id]
    # Example: 
        go to countertop 1
    cool [object] [id] with [receptacle] [id]
    # Example: 
        cool egg 1 with fridge 1
\end{spverbatim}
\end{mdframed}

\begin{mdframed}[style=mystyle, frametitle={ALFWorld: Policy improvement tips and primitives for skill find(object)}]
\addcontentsline{toc}{paragraph}{ALFWorld: Policy improvement tips and primitives for skill find(object)}
\small\ttfamily
\begin{spverbatim}
1. Thoroughly check each location before moving on to the next.
2. Do not pick up unnecessary items.
3. Do not waste time checking locations where the item is unlikely to be found.
4. Check all possible locations when looking for an item.
5. Don't assume an item won't be in a certain location based on where similar items were found.
6. If you can't find an item, try to revisit the locations you have already checked. The item might be hidden or covered by other items.
7. Do not exclude any locations when searching for an item, even if they seem unlikely.
8. Expand your search area when you can't find an item in the expected locations.
9. Remember that items can be found in unexpected places.
10. Avoid trying to take an item from a location where it has not been found.
11. Revisit locations where similar items were found as they might contain the item you are looking for.
12. Always check all possible locations, even if they seem unlikely to contain the item.
13. Remember to check the dining table when looking for items like a remote control.
14. You can think about your plan in the first step, and after that you can only select from the following actions in your response (please strictly follow the syntax):
    go to [receptacle] [id]
    # Example: 
        go to countertop 1
    open [receptacle] [id]
    # Example: 
        open drawer 2 
\end{spverbatim}
\end{mdframed}

\begin{mdframed}[style=mystyle, frametitle={ALFWorld: Policy improvement tips and primitives for skill heat(object)}]
\addcontentsline{toc}{paragraph}{ALFWorld: Policy improvement tips and primitives for skill heat(object)}
\small\ttfamily
\begin{spverbatim}
1. Try to execute a different action when you get stuck in a loop of taking identical actions.
2. Ensure you have the correct object in your possession before attempting to perform actions on it.
3. Check all possible locations if you can't find an object in the expected places.
4. You can directly heat an object with a microwave by taking the corresponding action.
5. If an action does not work, try a different approach or command.
6. Focus on relevant locations to complete the task efficiently.
7. A cup is not the same as a mug. Make sure to find and use the correct object for the task.
8. You can think about your plan in the first step, and after that you can only select from the following actions in your response (please strictly follow the syntax):
    go to [receptacle] [id]
    # Example: 
        go to countertop 1
    heat [object] [id] with [receptacle] [id]
    # Example: 
        heat egg 1 with microwave 1 
\end{spverbatim}
\end{mdframed}

\begin{mdframed}[style=mystyle, frametitle={ALFWorld: Policy improvement tips and primitives for skill put(object, receptacle)}]
\addcontentsline{toc}{paragraph}{ALFWorld: Policy improvement tips and primitives for skill put(object, receptacle)}
\small\ttfamily
\begin{spverbatim}
1. When looking for an object, make sure to check all possible locations.
2. If an object is not in the expected location, it may be somewhere else in the room.
3. Avoid getting stuck in a loop of unsuccessful actions. If an action doesn't work, try a different approach.
4. Not all surfaces may be suitable for placing objects. If one doesn't work, try another.
5. Focus on the task at hand and avoid unnecessary actions, such as searching through irrelevant locations or performing unnecessary actions.
6. Use the correct preposition when placing an object on a surface. The correct action is to put the object "in/on" the surface, not just "on" the surface.
7. Do not attempt to modify the object if it is not necessary for the task.
8. When placing an object on a surface, make sure the surface is not already full. If it is, try a different surface.
9. After picking up an object, perform the necessary action immediately instead of moving around unnecessarily.
10. If you already have the object needed for the task, do not try to find more of the same object.
11. Place the object in the first available and suitable location instead of unnecessarily checking other locations.
12. Use the correct command to put an object in a cabinet. The correct command is "put [object] in/on [surface]".
13. After cleaning an object, if the task is to put it in a drawer, do it immediately instead of checking other drawers.
14. If you put an object in a device (like a microwave), remember to take it out before trying to put it in another location.
15. Do not attempt to put an object on top of the fridge. The correct action is to put the object "in/on" the fridge.
16. Do not attempt to put an object on top of other objects on a surface. The correct action is to put the object directly "in/on" the surface.
17. You can think about your plan in the first step, and after that you can only select from the following actions in your response (please strictly follow the syntax):
    go to [receptacle] [id]
    # Example: 
        go to countertop 1
    open [receptacle] [id]
    # Example: 
        open drawer 2 
    put [object] [id] in/on [receptacle] [id]
    # Example: 
        put lettuce 1 in/on countertop 1
\end{spverbatim}
\end{mdframed}

\begin{mdframed}[style=mystyle, frametitle={ALFWorld: Policy improvement tips and primitives for skill take(object)}]
\addcontentsline{toc}{paragraph}{ALFWorld: Policy improvement tips and primitives for skill take(object)}
\small\ttfamily
\begin{spverbatim}
1. Ensure the object you want to take is actually in the location you are trying to take it from.
2. Do not get stuck in a loop of trying to take an object from a location where it is not present.
3. Avoid unnecessary actions that do not contribute to the completion of the task.
4. Be aware of your current location before attempting to take an object.
5. Explore different locations if you cannot find the object in the initial locations you check.
6. If you can't find an object, try to look for it in other possible locations.
7. Don't waste time checking empty or irrelevant locations. Focus on the locations where the object is likely to be found.
8. Pay attention to the task description and take the correct object.
9. Don't confuse similar objects. Make sure to take the correct object as per the task description.
10. Check all possible locations before attempting to take an object.
11. Avoid unnecessary actions that are not related to the task, such as trying to put an object on various shelves when the task is to take an object.
12. Laptops can also be found on sofas, not just on tables or in drawers.
13. You can think about your plan in the first step, and after that you can only select from the following actions in your response (please strictly follow the syntax):
    take [object] [id] from [receptacle] [id]
    # Example: 
        take lettuce 1 from countertop 1  
\end{spverbatim}
\end{mdframed}

\begin{mdframed}[style=mystyle, frametitle={ALFWorld: Policy improvement tips and primitives for skill use(object)}]
\addcontentsline{toc}{paragraph}{ALFWorld: Policy improvement tips and primitives for skill use(object)}
\small\ttfamily
\begin{spverbatim}
1. Stay focused on the task and avoid unnecessary actions that are not related to the task.
2. Some objects can be used directly without needing to be moved or interacted with other objects.
3. If an action doesn't work, it's likely not necessary for the task. Try a different approach.
4. Avoid getting stuck in a loop of unproductive actions. If an action doesn't produce a result, it's unlikely to do so with repeated attempts.
5. Don't assume that a task requires complex actions or interactions between objects. Sometimes, the solution is straightforward.
6. You can think about your plan in the first step, and after that you can only select from the following actions in your response (please strictly follow the syntax):
    use [receptacle] [id]
    # Example: 
        use desklamp 1
\end{spverbatim}
\end{mdframed}

%% file: prompts/Webshop/hier_new_llmdk.tex
\begin{mdframed}[style=mystyle, frametitle={WebShop: Policy improvement tips and primitives for skill purchase\_item()}]
\addcontentsline{toc}{paragraph}{WebShop: Policy improvement tips and primitives for skill purchase\_item()}
\small\ttfamily
\begin{spverbatim}
1. You should strictly use the exact words of each clickable option without making any changes.
2. Always remember to select the attributes of an item before purchasing it.
3. Ensure to select the correct color, size, and style options as per the task instruction before purchasing an item.
4. Do not rush to purchase an item without selecting its attributes.
5. Carefully read and understand the task instruction before making a selection.
6. Always ensure that the options you want to select are available on the page before attempting to click on them.
7. Always select all the necessary attributes of an item as per the task instruction before purchasing it.
8. Always consider the price limit in the instruction when selecting a product.
9. Use simplified and more general terms when searching for items to get more accurate results.
10. Please only generate following actions in your response:
    click[Buy Now]
    # Example click[Buy Now]
\end{spverbatim}
\end{mdframed}

\begin{mdframed}[style=mystyle, frametitle={WebShop: Policy improvement tips and primitives for skill search\_item()}]
\addcontentsline{toc}{paragraph}{WebShop: Policy improvement tips and primitives for skill search\_item()}
\small\ttfamily
\begin{spverbatim}
1. Use specific search terms that include all the essential details of the item you are looking for.
2. Use the exact terminology when performing a search to get the most accurate results.
3. Always include all the necessary details in the search query to get the most accurate results.
4. Use quotation marks around specific search terms to get the most accurate search results.
5. Avoid being overly specific in your search terms as it may limit the search results.
6. Please only generate following actions in your response:
    search[Query]
    # Example search[3 ounce bright citrus deodorant sensitive skin]
    click[Back to Search]
    # Example click[Back to Search]
\end{spverbatim}
\end{mdframed}

\begin{mdframed}[style=mystyle, frametitle={WebShop: Policy improvement tips and primitives for skill select\_item\_attributes()}]
\addcontentsline{toc}{paragraph}{WebShop: Policy improvement tips and primitives for skill select\_item\_attributes()}
\small\ttfamily
\begin{spverbatim}
1. You should strictly use the exact words of each clickable option without making any changes.
2. Ensure to select all the necessary attributes as per the instruction.
3. Pay attention to the product details to ensure they match the instruction.
4. Be aware that color and size options may not always be straightforward and may include additional information or codes.
5. Be careful with the case and spacing of the words when selecting item attributes.
6. Make sure to select a product that has all the required attributes as per the instruction.
7. Use quotation marks around specific phrases when searching for items to get more accurate results.
8. Please only generate following actions in your response:
    click[Attribute]
    # Example click[38 m eu]
    click[Color]
    # Example click[brown]
    click[Style]
    # Example click[powerlite 1785w]
    click[Size]
    # Example click[8.5]
    click[Flavor]
    # Example click[plantain chips]
    click[Features]
    # Example click[Features]
    click[Digital Storage Capacity]
    # Example click[32 gb]
    click[Offer Type]
    # Example click[lockscreen ad-supported]
    click[Description]
    # Example click[Description]
\end{spverbatim}
\end{mdframed}

\begin{mdframed}[style=mystyle, frametitle={WebShop: Policy improvement tips and primitives for skill select\_item()}]
\addcontentsline{toc}{paragraph}{WebShop: Policy improvement tips and primitives for skill select\_item()}
\small\ttfamily
\begin{spverbatim}
1. You should strictly use the exact words of each clickable option without making any changes.
2. When searching for items with specific characteristics, use quotation marks around the specific characteristic to narrow down the search results.
3. Review multiple items before making a selection to ensure the chosen item best fits the given criteria.
4. After finding the desired item, make sure to complete the task by selecting the item.
5. Avoid over-specifying the search terms to get more relevant results.
6. Always ensure to complete the task as per the instructions given.
7. Pay attention to the item details in the search results to ensure they match the given criteria.
8. Ensure the selected item strictly meets the given criteria.
9. Include all the criteria in the search query to narrow down the search results and find the most suitable item more efficiently.
10. Confirm the specific characteristics of the item by checking the item details.
11. Always consider the price limit when selecting an item.
12. After clicking on an item, make sure to select it to complete the task.
13. Please only generate following actions in your response:
    click[Item ID]
    # Example click[B0054KM7IY]
    click[< Prev]
    # Example click[< Prev]
\end{spverbatim}
\end{mdframed}

%% file: prompts/ALFworld/knowledge_prompt_v1.tex
\begin{mdframed}[style=mystyle, frametitle={ALFWorld: Contrastive tip distillation prompt}]
\addcontentsline{toc}{paragraph}{ALFWorld: Contrastive tip distillation prompt}
\small\ttfamily
\begin{spverbatim}
You will be given both failure and success histories in which you were placed in an environment and given a task to complete. By comparing the two histories, please find the mistakes in the failure trial, and summarize general tips to fix the mistakes. The tips can also be about the rules in this particular environment. You will need these tips later to solve a similar task. When there is already a list of tips to start with, please keep them in your response or improve them, and you can also append new tips according to the given new trials. You should not add duplicate tips. Give your response after "New summarization: "

Here is one example:

Task 1:
Failure trial:
You are in the middle of a room. Looking quickly around you, you see a cabinet 10, a cabinet 9, a cabinet 8, a cabinet 7, a cabinet 6, a cabinet 5, a cabinet 4, a cabinet 3, a cabinet 2, a cabinet 1, a coffeemachine 1, a countertop 2, a countertop 1, a diningtable 2, a diningtable 1, a drawer 3, a drawer 2, a drawer 1, a fridge 1, a garbagecan 1, a microwave 1, a sinkbasin 1, a stoveburner 4, a stoveburner 3, a stoveburner 2, a stoveburner 1, and a toaster 1.
Your task is to: put a hot mug on a countertop.
> go to countertop 1
On the countertop 1, you see a butterknife 1, a cup 1, a fork 2, and a mug 1.
> take mug 1 from countertop 1
You pick up the mug 1 from the countertop 1.
> go to stove burner 1
On the stoveburner 1, you see a pan 2.
> heat mug 1 with stoveburner 1
Nothing happens.
> heat mug 1 with stoveburner 1
Nothing happens.

Success trial:
You are in the middle of a room. Looking quickly around you, you see a cabinet 10, a cabinet 9, a cabinet 8, a cabinet 7, a cabinet 6, a cabinet 5, a cabinet 4, a cabinet 3, a cabinet 2, a cabinet 1, a coffeemachine 1, a countertop 2, a countertop 1, a diningtable 2, a diningtable 1, a drawer 3, a drawer 2, a drawer 1, a fridge 1, a garbagecan 1, a microwave 1, a sinkbasin 1, a stoveburner 4, a stoveburner 3, a stoveburner 2, a stoveburner 1, and a toaster 1.
Your task is to: put a hot mug on a countertop.
> go to countertop 1
On the countertop 1, you see a butterknife 1, a cup 1, a fork 2, and a mug 1.
> take mug 1 from countertop 1
You pick up the mug 1 from the countertop 1.
> go to microwave 1
You arrive at loc 22. The microwave 1 is closed.
> heat mug 1 with microwave 1
You heat the mug 1 using the microwave 1.
> go to countertop 1
On the countertop 1, you see a butterknife 1, a cup 1, a fork 2.
> put mug 1 in/on countertop 1
You put the mug 1 in/on the countertop 1

New summarization: In task 1, I made a mistake in heating a mug with a stoveburner, and I got stuck in a loop where I continually did this. As shown in the success trial, the correct way is to heat a mug with a microwave. Thus, I should try different actions when getting stuck in a loop; I should use a microwave to heat a mug, and I can directly heat an object with a microwave without opening the microwave and putting the object in it.

Tips:
- Try to execute a different action when you get stuck in a loop of taking identical actions.
- Mug should be heated by a microwave rather than a stoveburner.
- You can directly heat an object with a microwave by taking the corresponding action.
\end{spverbatim}
\end{mdframed}

%% file: prompts/ALFworld/knowledge_prompt_non_contrast_v1.tex
\begin{mdframed}[style=mystyle, frametitle={ALFWorld: Non-contrastive tip distillation prompt}]
\addcontentsline{toc}{paragraph}{ALFWorld: Non-contrastive tip distillation prompt}
\small\ttfamily
\begin{spverbatim}
You will be given success histories in which you were placed in an environment and given a task to complete. Please summarize general tips according to the success histories. The tips can be the rules in this particular environment. You will need these tips later to solve a similar task. When there is already a list of tips to start with, please keep them in your response or improve them, and you can also append new tips according to the given new trials. You do not have to add new tips if there is no one. You should not add duplicate tips. Give your response after "New summarization: "

Here is one example:

Task 1:
Success trial:
You are in the middle of a room. Looking quickly around you, you see a cabinet 10, a cabinet 9, a cabinet 8, a cabinet 7, a cabinet 6, a cabinet 5, a cabinet 4, a cabinet 3, a cabinet 2, a cabinet 1, a coffeemachine 1, a countertop 2, a countertop 1, a diningtable 2, a diningtable 1, a drawer 3, a drawer 2, a drawer 1, a fridge 1, a garbagecan 1, a microwave 1, a sinkbasin 1, a stoveburner 4, a stoveburner 3, a stoveburner 2, a stoveburner 1, and a toaster 1.
Your task is to: put a hot mug on a countertop.
> go to countertop 1
On the countertop 1, you see a butterknife 1, a cup 1, a fork 2, and a mug 1.
> take mug 1 from countertop 1
You pick up the mug 1 from the countertop 1.
> go to microwave 1
You arrive at loc 22. The microwave 1 is closed.
> heat mug 1 with microwave 1
You heat the mug 1 using the microwave 1.
> go to countertop 1
On the countertop 1, you see a butterknife 1, a cup 1, a fork 2.
> put mug 1 in/on countertop 1
You put the mug 1 in/on the countertop 1

New summarization: In task 1, as shown in the success trial, the correct way is to heat a mug with a microwave. Thus, I should use a microwave to heat a mug, and I can directly heat an object with a microwave without opening the microwave and putting the object in it. 

Tips:
- Mug should be heated by a microwave rather than a stoveburner.
- You can directly heat an object with a microwave by taking the corresponding action.
\end{spverbatim}
\end{mdframed}

%% file: prompts/Webshop/knowledge_prompt_v1.tex
\begin{mdframed}[style=mystyle, frametitle={WebShop: Contrastive tip distillation prompt}]
\addcontentsline{toc}{paragraph}{WebShop: Contrastive tip distillation prompt}
\small\ttfamily
\begin{spverbatim}
You will be given both failure and success histories in which you were placed in an environment and given a task to complete. By comparing the two histories, please find the mistakes in the failure trial, and summarize general tips to fix the mistakes. The tips can also be about the rules in this particular environment. You will need these tips later to solve a similar task. When there is already a list of tips to start with, please keep them in your response or improve them, and you can also append new tips according to the given new trials. You should not add duplicate tips. You do not have to add new tips if there is no one. Give your response after "New summarization: "

Here is one example:
Task1:
Failure Trial:
Webshop 
Instruction:  
i would like a 3 ounce bottle of bright citrus deodorant for sensitive skin, and price lower than 50.00 dollars 
[Search]  

Action: search[3 ounce bright citrus deodorant sensitive skin]
Observation: 
[Back to Search] 
Page 1 (Total results: 50) 
[Next >] 
[B078GWRC1J] 
Bright Citrus Deodorant by Earth Mama | Natural and Safe for Sensitive Skin, Pregnancy and Breastfeeding, Contains Organic Calendula 3-Ounce 
\$10.99 
[B078GTKVXY] 
Ginger Fresh Deodorant by Earth Mama | Natural and Safe for Sensitive Skin, Pregnancy and Breastfeeding, Contains Organic Calendula 3-Ounce 
\$10.99 
[B08KBVJ4XN] 
Barrel and Oak - Aluminum-Free Deodorant, Deodorant for Men, Essential Oil-Based Scent, 24-Hour Odor Protection, Cedar & Patchouli Blend, Gentle on Sensitive Skin (Mountain Sage, 2.7 oz, 2-Pack) 
\$15.95  

Action: click[B078GWRC1J]
Observation: 
[Back to Search] 
[< Prev] 
scent [assorted scents][bright citrus][calming lavender][ginger fresh][simply non-scents]
size [travel set (4-pack)][3 ounce (pack of 1)][3-ounce (2-pack)]
Bright Citrus Deodorant by Earth Mama | Natural and Safe for Sensitive Skin, Pregnancy and Breastfeeding, Contains Organic Calendula 3-Ounce 
Price: \$10.99 
Rating: N.A. 
[Description] 
[Features] 
[Reviews] 
[Buy Now]  

Action: click[BRIGHT CITRUS]
Observation: Invalid action!

Action: click[BRIGHT CITRUS]
Observation: Invalid action!

Action: click[BRIGHT CITRUS]
Observation: Invalid action!

Success Trial:
Webshop 
Instruction:  
i would like a 3 ounce bottle of bright citrus deodorant for sensitive skin, and price lower than 50.00 dollars 
[Search]  

Action: search[3 ounce bright citrus deodorant sensitive skin]
Observation: 
[Back to Search] 
Page 1 (Total results: 50) 
[Next >] 
[B078GWRC1J] 
Bright Citrus Deodorant by Earth Mama | Natural and Safe for Sensitive Skin, Pregnancy and Breastfeeding, Contains Organic Calendula 3-Ounce 
\$10.99 
[B078GTKVXY] 
Ginger Fresh Deodorant by Earth Mama | Natural and Safe for Sensitive Skin, Pregnancy and Breastfeeding, Contains Organic Calendula 3-Ounce 
\$10.99 
[B08KBVJ4XN] 
Barrel and Oak - Aluminum-Free Deodorant, Deodorant for Men, Essential Oil-Based Scent, 24-Hour Odor Protection, Cedar & Patchouli Blend, Gentle on Sensitive Skin (Mountain Sage, 2.7 oz, 2-Pack) 
\$15.95  

Action: click[B078GWRC1J]
Observation: 
[Back to Search] 
[< Prev] 
scent [assorted scents][bright citrus][calming lavender][ginger fresh][simply non-scents]
size [travel set (4-pack)][3 ounce (pack of 1)][3-ounce (2-pack)]
Bright Citrus Deodorant by Earth Mama | Natural and Safe for Sensitive Skin, Pregnancy and Breastfeeding, Contains Organic Calendula 3-Ounce 
Price: \$10.99 
Rating: N.A. 
[Description] 
[Features] 
[Reviews] 
[Buy Now]  

Action: click[bright citrus]
Observation: You have clicked bright citrus. 

Action: click[3 ounce (pack of 1)]
Observation: You have clicked 3 ounce (pack of 1). 

Action: click[Buy Now]

New summarization: In task 1, I get stuck in a loop where I continued taking an invalid action click[BRIGHT CITRUS]. Thus, I should try different actions when getting stuck in a loop; I also made a mistake on capitalizing the name of a scent option [bright citrus], so that click[BRIGHT CITRUS] become invalid. As shown in the success trial, the correct action should be click[bright citrus]. Thus, I should strictly use the exact words of each option without making any changes.

Tips:
- Try to execute a different action when you get stuck in a loop of taking identical actions.
- You should strictly use the exact words of each clickable option without making any changes.
\end{spverbatim}
\end{mdframed}

%% file: prompts/Webshop/knowledge_prompt_non_contrast.tex
\begin{mdframed}[style=mystyle, frametitle={WebShop: Non-contrastive tip distillation prompt}]
\addcontentsline{toc}{paragraph}{WebShop: Non-contrastive tip distillation prompt}
\small\ttfamily
\begin{spverbatim}
You will be given success histories in which you were placed in an environment and given a task to complete. Please summarize general tips according to the success histories. The tips can also be about the rules in this particular environment. You will need these tips later to solve a similar task. When there is already a list of tips to start with, please keep them in your response or improve them, and you can also append new tips according to the given new trials. You should not add duplicate tips. You do not have to add new tips if there is no one. Give your response after "New summarization: "

Here is one example:
Task1:
Success Trial:
Webshop 
Instruction:  
i would like a 3 ounce bottle of bright citrus deodorant for sensitive skin, and price lower than 50.00 dollars 
[Search]  

Action: search[3 ounce bright citrus deodorant sensitive skin]
Observation: 
[Back to Search] 
Page 1 (Total results: 50) 
[Next >] 
[B078GWRC1J] 
Bright Citrus Deodorant by Earth Mama | Natural and Safe for Sensitive Skin, Pregnancy and Breastfeeding, Contains Organic Calendula 3-Ounce 
\$10.99 
[B078GTKVXY] 
Ginger Fresh Deodorant by Earth Mama | Natural and Safe for Sensitive Skin, Pregnancy and Breastfeeding, Contains Organic Calendula 3-Ounce 
\$10.99 
[B08KBVJ4XN] 
Barrel and Oak - Aluminum-Free Deodorant, Deodorant for Men, Essential Oil-Based Scent, 24-Hour Odor Protection, Cedar & Patchouli Blend, Gentle on Sensitive Skin (Mountain Sage, 2.7 oz, 2-Pack) 
\$15.95  

Action: click[B078GWRC1J]
Observation: 
[Back to Search] 
[< Prev] 
scent [assorted scents][bright citrus][calming lavender][ginger fresh][simply non-scents]
size [travel set (4-pack)][3 ounce (pack of 1)][3-ounce (2-pack)]
Bright Citrus Deodorant by Earth Mama | Natural and Safe for Sensitive Skin, Pregnancy and Breastfeeding, Contains Organic Calendula 3-Ounce 
Price: \$10.99 
Rating: N.A. 
[Description] 
[Features] 
[Reviews] 
[Buy Now]  

Action: click[bright citrus]
Observation: You have clicked bright citrus. 

Action: click[3 ounce (pack of 1)]
Observation: You have clicked 3 ounce (pack of 1). 

Action: click[Buy Now]

New summarization: In task 1, as shown in the success trial, I searched by several keywords, selected a relevant item and clicked required attributes before placing the order.

Tips:
- You should searched with keywords, select the relevant item to check and click attributes according to the instruction. 
\end{spverbatim}
\end{mdframed}

%% file: prompts/ALFworld/llm_dk_v2.tex
\begin{mdframed}[style=mystyle, frametitle={ALFWorld: Distilled policy improvement tips with contrastive prompt}]
\addcontentsline{toc}{paragraph}{ALFWorld: Distilled policy improvement tips with contrastive prompt}
\small\ttfamily
\begin{spverbatim}
Please consider the following tips to solve new tasks:

1. Try to execute a different action when you get stuck in a loop of taking identical actions.
2. When stuck in a loop, try a different action or look for the item in a different location.
3. Mug should be heated by a microwave rather than a stoveburner.
4. You can directly heat an object with a microwave by taking the corresponding action.
5. Remember to heat items in the microwave when required.
6. Use the correct command to place items on the dining table.
7. Use the correct commands to interact with items.
8. Search in all possible locations for the required item.
9. Check all possible locations for the required item, not just the most obvious ones.
10. Items can be found in unexpected places like the garbage can or microwave.
11. Remember to cool items in the fridge when required.
12. Remember to clean items when required before placing them in their final location.
13. You can clean items directly with the sinkbasin without needing to turn it on.
14. You don't need to move items to use them together.
15. Sometimes the simplest action is the correct one.
16. Perform tasks in the correct order.
\end{spverbatim}
\end{mdframed}

%% file: prompts/ALFworld/llm_dk_noncontrast_v2.tex
\begin{mdframed}[style=mystyle, frametitle={ALFWorld: Distilled policy improvement tips with non-contrastive prompt}]
\addcontentsline{toc}{paragraph}{ALFWorld: Distilled policy improvement tips with non-contrastive prompt}
\small\ttfamily
\begin{spverbatim}
Please consider the following tips to solve new tasks:

1. Mug should be heated by a microwave rather than a stoveburner. You can directly heat an object with a microwave by taking the corresponding action. Bread can also be heated using a microwave.
2. After heating an object, it can be placed in the fridge for cooling. You can directly cool an object with a fridge by taking the corresponding action. After cooling a cup in the fridge, it can be placed back in the microwave.
3. Cellphones can be found on sidetables and beds. After finding a cellphone, place it in the dresser.
4. Toiletpaper can be found in drawers, not necessarily on toiletpaperhangers.
5. Cups can be found in microwaves, not necessarily on countertops or in cabinets.
6. Kettles can be found on stoveburners. After finding the kettles, place them in the cabinet.
7. Keychains can be found on shelves. After finding the keychains, place them in the sidetable.
8. Watches can be found in drawers and on dining tables. After finding the watches, place them on the desired shelf.
9. To clean a dish sponge, a plate, or a knife, use the sink basin. After cleaning the dish sponge, the plate, or the knife, place it on the shelf, countertop, or drawer respectively.
10. To cool a bowl or a mug, use the fridge. After cooling the bowl or the mug, place it in the cabinet.
11. Vases can be found on shelves. After finding the vase, take it to the desk where the desklamp is located and turn on the desklamp to examine the vase.
12. Eggs can be found in garbage cans. After finding the egg, heat it using the microwave before placing it back in the garbage can.
13. Apples can be found on countertops or in garbage cans. After finding the apple, cool it using the fridge before placing it in the microwave or heat it using the microwave before placing it back in the garbage can.
14. Forks can be found on countertops. After finding the fork, place it in the drawer.
15. Spoons can be found on countertops. After finding the spoon, clean it using the sink basin before placing it on the sidetable.
16. Lettuce can be found in the sink basin. After finding the lettuce, cool it using the fridge before placing it in the garbage can.
17. Bowls can be found in cabinets. After finding the bowl, cool it using the fridge before placing it in another cabinet.
18. Potatoes can be found on dining tables or in microwaves. After finding the potato, heat it using the microwave before placing it in the fridge or clean it using the sink basin before placing it on the sidetable.
19. Pillows can be found on beds. After finding the pillow, take it to the desk where the desklamp is located and turn on the desklamp to examine the pillow.
20. CDs can be found on dining tables. After finding the CD, place it on the sidetable.
21. Butterknives can be found on countertops or in drawers. After finding the butterknife, clean it using the sink basin before placing it on the countertop or dining table.
22. Credit cards can be found on sidetables. After finding the credit cards, place them on the dining table.
23. You can directly use a desklamp to examine an object without turning it on first.
24. Pots can be found in sink basins. After finding the pot, cool it using the fridge before placing it in the cabinet.
\end{spverbatim}
\end{mdframed}

%% file: prompts/ALFworld/knowledge_exp_with_primitives.tex
\begin{mdframed}[style=mystyle, frametitle={ALFWorld: Human distilled tips and primitives}]
\addcontentsline{toc}{paragraph}{ALFWorld: Human distilled tips and primitives}
\small\ttfamily
\begin{spverbatim}
Please consider the following tips to solve new tasks:

1. Try to execute a different action when you stuck in a loop of taking identical actions. 
2. Try to search all other locations if the object could not be found in initial likely locations.
3. Remember the location of each object you have seen in order to quickly find it later.
4. Remember the number of objects you have seen in each location, so you can directly go back the location to find another one.
5. Carefully check the outcome of an action before taking next action. 
6. You can directly put an object in/on an occupied receptacle.
7. You can only take actions to the object in hand or the objects in the current scene.
8. Do not mismatch target obejct with similar object, for example, a mug is different with a cup. 
9. Do not express sorry/apology in your response.
10. Please only generate following actions in your response and strictly follow the syntax:

    go to [receptacle] [id]
    # Example: 
        go to countertop 1
    open [receptacle] [id]
    # Example: 
        open drawer 2 
    clean [object] [id] with [receptacle] [id]
    # Example: 
        clean lettuce 1 with sinkbasin 1 
    take [object] [id] from [receptacle] [id]
    # Example: 
        take lettuce 1 from countertop 1  
    close [receptacle] [id]
    # Example: 
        close drawer 2 
    heat [object] [id] with [receptacle] [id]
    # Example: 
        heat egg 1 with microwave 1 
    put [object] [id] in/on [receptacle] [id]
    # Example: 
        put lettuce 1 in/on countertop 1
    use [object] [id]
    # Example: 
        use desklamp 1 
    cool [object] [id] with [receptacle] [id]
    # Example: 
        cool egg 1 with fridge 1
\end{spverbatim}
\end{mdframed}

%% file: prompts/Webshop/llm_DK.tex
\begin{mdframed}[style=mystyle, frametitle={WebShop: Distilled policy improvement tips with contrastive prompt}]
\addcontentsline{toc}{paragraph}{WebShop: Distilled policy improvement tips with contrastive prompt}
\small\ttfamily
\begin{spverbatim}
Please consider the following tips to solve new tasks:

1. You should strictly use the exact words of each clickable option without making any changes.
2. Always ensure to select all the necessary options before making a purchase.
3. Pay attention to the specific requirements in the instruction, such as color and size, and make sure to select them correctly.
4. Use the correct search terms to find the desired product. Avoid using special characters like slashes (/) in the search term.
5. Make sure the product selected meets all the requirements in the instruction.
6. Avoid making assumptions about the available options. Always select from the options provided.
7. Always select the color or any other specific attribute mentioned in the instruction before making a purchase.
8. Ensure to select the correct style or model of the product when multiple options are available.
9. Check the product's features to ensure it meets all the requirements in the instruction before making a purchase.
10. If the search results are not satisfactory, adjust the search terms instead of repeating the same search.
11. Do not include the price limit in the search term. Instead, check the price of each item in the search results.
12. If the first product selected does not meet all the requirements, go back to the search results and select a different product.
13. Pay attention to the use of quotation marks in the search term to ensure accurate results.
14. Only select the options that are specifically mentioned in the instruction. Avoid selecting unnecessary options.
\end{spverbatim}
\end{mdframed}

%% file: prompts/Webshop/llm_DK_noncontrast.tex
\begin{mdframed}[style=mystyle, frametitle={WebShop: Distilled policy improvement tips with non-contrastive prompt}]
\addcontentsline{toc}{paragraph}{WebShop: Distilled policy improvement tips with non-contrastive prompt}
\small\ttfamily
\begin{spverbatim}
Please consider the following tips to solve new tasks:

1. You should search with specific keywords, including the color or any specific feature if required, select the relevant item to check and click required attributes before placing the order. 
2. Make sure to select the correct color and feature before purchasing the item.
3. Always check the price of the item to ensure it is within the specified budget.
4. If the first selected item does not meet the requirements, go back to the search results and select another item. Repeat this process until a suitable item is found.
5. Ensure to select the correct quantity or count of the item before purchasing.
6. Make sure to select the correct size before purchasing the item.
7. When searching for a specific model of a product, include the model name in the search keywords.
8. Ensure to select the correct flavor or variety when purchasing food items.
9. Check the material of the item if it is specified in the task.
10. If the item does not meet the requirements after checking its features, go back and select another item.
11. When searching for a wig, include the color and material in the search keywords.
12. Check the features of the product to ensure it meets the requirements before purchasing.
13. Check the description of the product to ensure it meets the requirements before purchasing.
14. Always check the product's features to ensure it meets the specific requirements of the task.
15. After selecting an item, check its features to ensure it meets the task requirements before purchasing.
\end{spverbatim}
\end{mdframed}

%% file: prompts/Webshop/knolwedge_with_primitives.tex
\begin{mdframed}[style=mystyle, frametitle={WebShop: Human distilled tips and primitives}]
\addcontentsline{toc}{paragraph}{WebShop: Human distilled tips and primitives}
\small\ttfamily
\begin{spverbatim}
Please consider the following tips to solve new tasks:

1. Please pay attention to instruction description to find the product that exactly matches with the requirement in the instruction;   
2. You can click[Description] to find the detailed description of a product; 
3. Remember to select all required attributes using click[Attributes] before click[Buy Now]; 
4. You should click[Description] or click[Features] to double-check attributes when there is no clickable attributes; 
5. You should pay attention to the current observation to check clickable buttons;
6. When the current selected product does not match the requirement in the instruction well, you should navigate to the results page to check other likely product candidates; 
7. When there is no product that matches the requirement in the instruction well, you should search again with a new query; 
8. You should click[Back to Search] before re-searching with a new query;
9. Please only generate following actions in your response:

    search[Query]
    # Example: search[noise cancelling cosycost usb microphone] 
    click[Attributes]
    # Example 1: click[16.9 fl oz (pack of 1)] 
    # Example 2: click[blue]
    click[Product Title]
    # Example: click[B0972Q1T8T] 
    click[Back to Search]
    # Example: click[Back to Search]
    click[< Prev]
    # Example: click[< Prev]
    click[Description]
    # Example: click[Description]
    click[Features]
    # Example: click[Features]
    click[Buy Now]
    # Example: click[Buy Now]

\end{spverbatim}
\end{mdframed}

%% file: prompts/ALFworld/skill_code.tex
\begin{mdframed}[style=mystyle, frametitle={ALFworld: Generated skill code}]
\addcontentsline{toc}{paragraph}{ALFworld: Generated skill code}
\small\ttfamily
\begin{spverbatim}
def clean_object(specific_object_to_clean):
    # Go to the sinkbasin
    action = go('sinkbasin 1')
    env_observation = interact_with_environment(action)

    # Clean the object with the sinkbasin
    action = clean(specific_object_to_clean, 'sinkbasin 1')
    env_observation = interact_with_environment(action)

def cool_object(specific_object_to_cool):
    # Go to the fridge
    action = go('fridge 1')
    env_observation = interact_with_environment(action)

    # Cool the object with the fridge
    action = cool(specific_object_to_cool, 'fridge 1')
    env_observation = interact_with_environment(action)

    return env_observation

def find_object(object_to_find, locations_to_search):
    for location in locations_to_search:
        action = go(location)
        env_observation, _ = interact_with_environment(action)
        if is_in(object_to_find, env_observation):
            specific_object_name = extract_pattern(env_observation,
                                                   object_to_find)
            return specific_object_name, location
        elif isclose(env_observation):
            action = open(location)
            env_observation, _ = interact_with_environment(action)
            if is_in(object_to_find, env_observation):
                specific_object_name = extract_pattern(env_observation,
                                                       object_to_find)
                return specific_object_name, location
    return None, None

def heat_object(specific_object_to_heat):
    # Go to the microwave
    action = go('microwave 1')
    env_observation = interact_with_environment(action)

    # Heat the specific object with the microwave
    action = heat(specific_object_to_heat, 'microwave 1')
    env_observation = interact_with_environment(action)

def put_object(specific_object_to_put, specific_target_location):
    # Go to the target location
    action = go(specific_target_location)
    env_observation, won = interact_with_environment(action)

    # If the target location is closed, open it
    if isclose(env_observation):
        action = open(specific_target_location)
        env_observation, won = interact_with_environment(action)

    # Put the object on the target location
    action = put(specific_object_to_put, specific_target_location)
    env_observation, won = interact_with_environment(action)

    # return whether env return won or not
    return won

def take_object(specific_object_to_take, pick_location):
    action = take(specific_object_to_take, pick_location)
    env_observation = interact_with_environment(action)
    return env_observation

def use_object(specific_object_to_use):
    action = use(specific_object_to_use)
    env_observation, won = interact_with_environment(action)
    return won
\end{spverbatim}
\end{mdframed}

%% file: prompts/ALFworld/skill_composition.tex
\begin{mdframed}[style=mystyle, frametitle={ALFworld: Skill Composition}]
\addcontentsline{toc}{paragraph}{ALFworld: Skill Composition}
\small\ttfamily
\begin{spverbatim}
# Based on the historical interaction and the new task description, the sub-task sequence should be:
# find(bowl)
# take(bowl)
# clean(bowl)
# put(bowl, cabinet 1)

# Find the bowl
specific_object_name, location = find_object('bowl', ["cabinet 6", "cabinet 5", "cabinet 4", "cabinet 3", "cabinet 2", "cabinet 1", "coffeemachine 1", "countertop 3", "countertop 2", "countertop 1", "drawer 3", "drawer 2", "drawer 1", "fridge 1", "garbagecan 1", "microwave 1", "shelf 3", "shelf 2", "shelf 1", "sinkbasin 1", "stoveburner 4", "stoveburner 3", "stoveburner 2", "stoveburner 1", "toaster 1"])
# Take the bowl
take_object(specific_object_name, location)
# Clean the bowl
clean_object(specific_object_name)
# Put the bowl in cabinet 1
won = put_object(specific_object_name, 'cabinet 1')
\end{spverbatim}
\end{mdframed}

%% file: prompts/Webshop/skill_code.tex
\begin{mdframed}[style=mystyle, frametitle={Webshop: Generated skill code}]
\addcontentsline{toc}{paragraph}{Webshop: Generated skill code}
\small\ttfamily
\begin{spverbatim}
def select_item(items, instruction):
    # items: dict {'item number': discription}
    # instruction: str

    # Extract requirements from instruction
    requirements = extract_requirement(instruction)

    # Initialize a dictionary to store the number of satisfied requirements for each item
    satisfied_requirements = {item: 0 for item in items.keys()}

    # Check each item against each requirement
    for item, description in items.items():
        for requirement in requirements:
            if requirement_match_check(description, requirement):
                satisfied_requirements[item] += 1
            
    # Select the item with the most satisfied requirements
    selected_item = max(satisfied_requirements, key=satisfied_requirements.get)
    return selected_item

def select_item_attributes(item_description, requirements):
    # item_description: string 
    # requirements: [requirements#1,requirements#2,requirements#3,...]
    
    # Extract the most relevant attributes from the item description based on the requirements
    relevant_attributes = select_most_relevant_attibutes(item_description, requirements)
    
    # Return the list of relevant attributes
    return relevant_attributes
\end{spverbatim}
\end{mdframed}

%% file: prompts/Webshop/skill_composition_GPT4.tex
\begin{mdframed}[style=mystyle, frametitle={Webshop: Skill Composition (GPT-4)}]
\addcontentsline{toc}{paragraph}{Webshop: Skill Composition (GPT-4)}
\small\ttfamily
\begin{spverbatim}
# Sub-task sequence based on the historical interaction and new task description:
# 1. search_item: search for a space-saving ottoman bench to match blue living room
# 2. select_item: select the item that's 100x45x45cm and price lower than 410.00 dollars
# 3. select_item_attributes: select the attributes that match the requirements
# 4. purchase_item: buy the selected item

# Start the task
# Sub-task 1: search_item
instruction = "i'm looking for a space-saving ottoman bench to match my blue living room. pick that one that's 100x45x45cm, and price lower than 410.00 dollars"
action = search(instruction)
env_obs, reward = interact_with_environment(action)

# Sub-task 2: select_item
items = parse_items_observation_into_items_dict(env_obs)
selected_item = select_item(items, instruction)
action = select(selected_item)
env_obs, reward = interact_with_environment(action)

# Sub-task 3: select_item_attributes
requirements = extract_requirement(instruction)
relevant_attributes = select_item_attributes(env_obs, requirements)
for attribute in relevant_attributes:
    action = click(attribute)
    env_obs, reward = interact_with_environment(action)

# Sub-task 4: purchase_item
action = buy()
env_obs, reward = interact_with_environment(action)

# Print the final reward
print(f"Final reward: {reward}")
\end{spverbatim}
\end{mdframed}

%% file: prompts/Webshop/skill_composition_GPT3_5.tex
\begin{mdframed}[style=mystyle, frametitle={Webshop: Skill Composition (GPT-3.5)}]
\addcontentsline{toc}{paragraph}{Webshop: Skill Composition (GPT-3.5)}
\small\ttfamily
\begin{spverbatim}
# Sub-task sequence:
# 1. search(instruction)
# 2. select_item()
# 3. select_item_attributes()
# 4. purchase_item()

# Code:
search('ultra hd motion detection surveillance dome camera color black size :6mp wdr 2.8 mm, and price lower than 190.00 dollars')

select_item()

select_item_attributes(item_description, requirements)

purchase_item()
select_item()
\end{spverbatim}
\end{mdframed}

%% file: 0_main.bbl
\begin{thebibliography}{27}
\providecommand{\natexlab}[1]{#1}
\providecommand{\url}[1]{\texttt{#1}}
\expandafter\ifx\csname urlstyle\endcsname\relax
  \providecommand{\doi}[1]{doi: #1}\else
  \providecommand{\doi}{doi: \begingroup \urlstyle{rm}\Url}\fi

\bibitem[Driess et~al.(2023)Driess, Xia, Sajjadi, Lynch, Chowdhery, Ichter,
  Wahid, Tompson, Vuong, Yu, et~al.]{driess2023palm}
Driess, D., Xia, F., Sajjadi, M.~S., Lynch, C., Chowdhery, A., Ichter, B.,
  Wahid, A., Tompson, J., Vuong, Q., Yu, T., et~al.
\newblock Palm-e: An embodied multimodal language model.
\newblock \emph{arXiv preprint arXiv:2303.03378}, 2023.

\bibitem[Fu et~al.(2022)Fu, Peng, Sabharwal, Clark, and Khot]{fu2022complexity}
Fu, Y., Peng, H., Sabharwal, A., Clark, P., and Khot, T.
\newblock Complexity-based prompting for multi-step reasoning.
\newblock \emph{arXiv preprint arXiv:2210.00720}, 2022.

\bibitem[Liang et~al.(2022)Liang, Huang, Xia, Xu, Hausman, Ichter, Florence,
  and Zeng]{codeaspolicies2022}
Liang, J., Huang, W., Xia, F., Xu, P., Hausman, K., Ichter, B., Florence, P.,
  and Zeng, A.
\newblock Code as policies: Language model programs for embodied control.
\newblock In \emph{arXiv preprint arXiv:2209.07753}, 2022.

\bibitem[Lightman et~al.(2023)Lightman, Kosaraju, Burda, Edwards, Baker, Lee,
  Leike, Schulman, Sutskever, and Cobbe]{lightman2023let}
Lightman, H., Kosaraju, V., Burda, Y., Edwards, H., Baker, B., Lee, T., Leike,
  J., Schulman, J., Sutskever, I., and Cobbe, K.
\newblock Let's verify step by step.
\newblock \emph{arXiv preprint arXiv:2305.20050}, 2023.

\bibitem[Ling et~al.(2023)Ling, Fang, Li, Huang, Lee, Memisevic, and
  Su]{ling2023deductive}
Ling, Z., Fang, Y., Li, X., Huang, Z., Lee, M., Memisevic, R., and Su, H.
\newblock Deductive verification of chain-of-thought reasoning.
\newblock \emph{arXiv preprint arXiv:2306.03872}, 2023.

\bibitem[Liu et~al.(2023{\natexlab{a}})Liu, Jiang, Zhang, Liu, Zhang, Biswas,
  and Stone]{liu2023llm+}
Liu, B., Jiang, Y., Zhang, X., Liu, Q., Zhang, S., Biswas, J., and Stone, P.
\newblock Llm+ p: Empowering large language models with optimal planning
  proficiency.
\newblock \emph{arXiv preprint arXiv:2304.11477}, 2023{\natexlab{a}}.

\bibitem[Liu et~al.(2023{\natexlab{b}})Liu, Yu, Zhang, Xu, Lei, Lai, Gu, Ding,
  Men, Yang, et~al.]{liu2023agentbench}
Liu, X., Yu, H., Zhang, H., Xu, Y., Lei, X., Lai, H., Gu, Y., Ding, H., Men,
  K., Yang, K., et~al.
\newblock Agentbench: Evaluating llms as agents.
\newblock \emph{arXiv preprint arXiv:2308.03688}, 2023{\natexlab{b}}.

\bibitem[Liu et~al.(2023{\natexlab{c}})Liu, Bahety, and Song]{liu2023reflect}
Liu, Z., Bahety, A., and Song, S.
\newblock Reflect: Summarizing robot experiences for failure explanation and
  correction.
\newblock \emph{arXiv preprint arXiv:2306.15724}, 2023{\natexlab{c}}.

\bibitem[Shinn et~al.(2023)Shinn, Labash, and Gopinath]{shinn2023reflexion}
Shinn, N., Labash, B., and Gopinath, A.
\newblock Reflexion: an autonomous agent with dynamic memory and
  self-reflection.
\newblock \emph{arXiv preprint arXiv:2303.11366}, 2023.

\bibitem[Shridhar et~al.(2020)Shridhar, Thomason, Gordon, Bisk, Han, Mottaghi,
  Zettlemoyer, and Fox]{ALFRED20}
Shridhar, M., Thomason, J., Gordon, D., Bisk, Y., Han, W., Mottaghi, R.,
  Zettlemoyer, L., and Fox, D.
\newblock {ALFRED: A Benchmark for Interpreting Grounded Instructions for
  Everyday Tasks}.
\newblock In \emph{The IEEE Conference on Computer Vision and Pattern
  Recognition (CVPR)}, 2020.
\newblock URL \url{https://arxiv.org/abs/1912.01734}.

\bibitem[Shridhar et~al.(2021)Shridhar, Yuan, C\^ot\'e, Bisk, Trischler, and
  Hausknecht]{ALFWorld20}
Shridhar, M., Yuan, X., C\^ot\'e, M.-A., Bisk, Y., Trischler, A., and
  Hausknecht, M.
\newblock {ALFWorld: Aligning Text and Embodied Environments for Interactive
  Learning}.
\newblock In \emph{Proceedings of the International Conference on Learning
  Representations (ICLR)}, 2021.
\newblock URL \url{https://arxiv.org/abs/2010.03768}.

\bibitem[Silver et~al.(2023)Silver, Dan, Srinivas, Tenenbaum, Kaelbling, and
  Katz]{silver2023generalized}
Silver, T., Dan, S., Srinivas, K., Tenenbaum, J.~B., Kaelbling, L.~P., and
  Katz, M.
\newblock Generalized planning in pddl domains with pretrained large language
  models.
\newblock \emph{arXiv preprint arXiv:2305.11014}, 2023.

\bibitem[Sun et~al.(2023)Sun, Zhuang, Kong, Dai, and Zhang]{sun2023adaplanner}
Sun, H., Zhuang, Y., Kong, L., Dai, B., and Zhang, C.
\newblock Adaplanner: Adaptive planning from feedback with language models.
\newblock \emph{arXiv preprint arXiv:2305.16653}, 2023.

\bibitem[Sutton \& Barto(2018)Sutton and Barto]{sutton2018reinforcement}
Sutton, R.~S. and Barto, A.~G.
\newblock \emph{Reinforcement learning: An introduction}.
\newblock MIT press, 2018.

\bibitem[Sutton et~al.(1999)Sutton, Precup, and Singh]{sutton1999between}
Sutton, R.~S., Precup, D., and Singh, S.
\newblock Between mdps and semi-mdps: A framework for temporal abstraction in
  reinforcement learning.
\newblock \emph{Artificial intelligence}, 112\penalty0 (1-2):\penalty0
  181--211, 1999.

\bibitem[Torabi et~al.(2018)Torabi, Warnell, and Stone]{torabi2018behavioral}
Torabi, F., Warnell, G., and Stone, P.
\newblock Behavioral cloning from observation.
\newblock In \emph{Proceedings of the 27th International Joint Conference on
  Artificial Intelligence}, IJCAI'18, pp.\  4950–4957. AAAI Press, 2018.
\newblock ISBN 9780999241127.

\bibitem[Wang et~al.(2023{\natexlab{a}})Wang, Xie, Jiang, Mandlekar, Xiao, Zhu,
  Fan, and Anandkumar]{wang2023voyager}
Wang, G., Xie, Y., Jiang, Y., Mandlekar, A., Xiao, C., Zhu, Y., Fan, L., and
  Anandkumar, A.
\newblock Voyager: An open-ended embodied agent with large language models.
\newblock \emph{arXiv preprint arXiv: Arxiv-2305.16291}, 2023{\natexlab{a}}.

\bibitem[Wang et~al.(2023{\natexlab{b}})Wang, Gonzalez-Pumariega, Sharma, and
  Choudhury]{wang2023demo2code}
Wang, H., Gonzalez-Pumariega, G., Sharma, Y., and Choudhury, S.
\newblock Demo2code: From summarizing demonstrations to synthesizing code via
  extended chain-of-thought.
\newblock \emph{arXiv preprint arXiv:2305.16744}, 2023{\natexlab{b}}.

\bibitem[Wang et~al.(2022)Wang, Wei, Schuurmans, Le, Chi, Narang, Chowdhery,
  and Zhou]{wang2022self}
Wang, X., Wei, J., Schuurmans, D., Le, Q., Chi, E., Narang, S., Chowdhery, A.,
  and Zhou, D.
\newblock Self-consistency improves chain of thought reasoning in language
  models.
\newblock \emph{arXiv preprint arXiv:2203.11171}, 2022.

\bibitem[Wang et~al.(2023{\natexlab{c}})Wang, Cai, Liu, Ma, and
  Liang]{wang2023describe}
Wang, Z., Cai, S., Liu, A., Ma, X., and Liang, Y.
\newblock Describe, explain, plan and select: Interactive planning with large
  language models enables open-world multi-task agents.
\newblock \emph{arXiv preprint arXiv:2302.01560}, 2023{\natexlab{c}}.

\bibitem[Wei et~al.(2022)Wei, Wang, Schuurmans, Bosma, Xia, Chi, Le, Zhou,
  et~al.]{wei2022chain}
Wei, J., Wang, X., Schuurmans, D., Bosma, M., Xia, F., Chi, E., Le, Q.~V.,
  Zhou, D., et~al.
\newblock Chain-of-thought prompting elicits reasoning in large language
  models.
\newblock \emph{Advances in Neural Information Processing Systems},
  35:\penalty0 24824--24837, 2022.

\bibitem[Xu et~al.(2023)Xu, Xu, Chi, Veloso, and Song]{xu2023xskill}
Xu, M., Xu, Z., Chi, C., Veloso, M., and Song, S.
\newblock {XS}kill: Cross embodiment skill discovery.
\newblock In \emph{7th Annual Conference on Robot Learning}, 2023.
\newblock URL \url{https://openreview.net/forum?id=8L6pHd9aS6w}.

\bibitem[Yang et~al.(2023)Yang, Yue, and He]{yang2023auto}
Yang, H., Yue, S., and He, Y.
\newblock Auto-gpt for online decision making: Benchmarks and additional
  opinions.
\newblock \emph{arXiv preprint arXiv:2306.02224}, 2023.

\bibitem[Yao et~al.(2022)Yao, Chen, Yang, and Narasimhan]{yao2022webshop}
Yao, S., Chen, H., Yang, J., and Narasimhan, K.
\newblock Webshop: Towards scalable real-world web interaction with grounded
  language agents.
\newblock In \emph{ArXiv}, 2022.

\bibitem[Yao et~al.(2023{\natexlab{a}})Yao, Yu, Zhao, Shafran, Griffiths, Cao,
  and Narasimhan]{yao2023tree}
Yao, S., Yu, D., Zhao, J., Shafran, I., Griffiths, T.~L., Cao, Y., and
  Narasimhan, K.
\newblock Tree of thoughts: Deliberate problem solving with large language
  models.
\newblock \emph{arXiv preprint arXiv:2305.10601}, 2023{\natexlab{a}}.

\bibitem[Yao et~al.(2023{\natexlab{b}})Yao, Zhao, Yu, Du, Shafran, Karthik, and
  Cao]{shunyu2023react}
Yao, S., Zhao, J., Yu, D., Du, N., Shafran, I., Karthik, N., and Cao, Y.
\newblock React: Synergizing reasoning and acting in language models.
\newblock \emph{https://arxiv.org/pdf/2210.03629.pdf}, 2023{\natexlab{b}}.

\bibitem[Zheng et~al.(2023)Zheng, Liu, Xie, Li, and Li]{zheng2023progressive}
Zheng, C., Liu, Z., Xie, E., Li, Z., and Li, Y.
\newblock Progressive-hint prompting improves reasoning in large language
  models.
\newblock \emph{arXiv preprint arXiv:2304.09797}, 2023.

\end{thebibliography}
